\documentclass{article}

\usepackage{main}
\usepackage{subfigure}
\usepackage[utf8]{inputenc} 
\usepackage[T1]{fontenc}    
\usepackage{hyperref}       
\usepackage{url}            
\usepackage{booktabs}       
\usepackage{amsfonts}       
\usepackage{nicefrac}       
\usepackage{microtype}      
\usepackage{lipsum}
\usepackage{fancyhdr}       
\usepackage{graphicx}       
\graphicspath{{media/}}     
\usepackage{float}
\usepackage{graphicx}
\usepackage{longtable}
\usepackage{multirow}
\usepackage{ragged2e}
\usepackage{makecell}
\title{Swin-Transformer-YOLOv5 for Real-Time Wine Grape Bunch Detection
\thanks{\textit{\underline{Citation}}: 
\textbf{Lu \textit{et al.} (2022). Swin-transformer-YOLOv5 for real-time wine grape bunch detection. https://arxiv.org/...
}} 
}

\author{
  Shenglian Lu (ORCID: 0000-0002-4957-9418), Xiaoyu Liu \\
  Guangxi Key Lab of Multisource Information Mining $\&$ Security \\
  College of Computer Science $\&$ Engineering \\
  Guangxi Normal University
  Guilin 541004, China\\
   \\
   \And
  Zixuan He \\
  Center for Precision and Automated Agricultural Systems \\
  Department of Biological Systems Engineering\\
  Washington State University \\
  Prosser, WA 99350, USA\\
   \\
   \And
  Xin Zhang (ORCID: 0000-0001-9654-3859), Wenbo Liu \\
  Department of Agricultural and Biological Engineering\\
  Mississippi State University \\
  Mississippi State, MS 39762, USA\\
    \And
   Manoj Karkee \\
  Center for Precision and Automated Agricultural Systems \\
  Department of Biological Systems Engineering\\
  Washington State University \\
  Prosser, WA 99350, USA\\
   \\
  \texttt{\{Corresponding author: Xin Zhang (\href{mailto:xzhang@abe.msstate.edu}{xzhang@abe.msstate.edu)}\}}
}

\begin{document}
\maketitle

\begin{abstract}
Precise canopy management is critical in vineyards for premium wine production because maximum crop load does not guarantee the best economic return for wine producers. Therefore, the wine grape growers need to keep tracking the number of grape bunches during the entire growth season for the optimal crop load per vine. Manual counting grape bunches can be highly labor-intensive, inefficient, and error prone. In this research, an integrated detection model, Swin-transformer-YOLOv5 or Swin-T-YOLOv5, was proposed for real-time wine grape bunch detection to inherit the advantages from both YOLOv5 and Swin-transformer. The research was conducted on two different grape varieties of Chardonnay (always white berry skin) and Merlot (white or white-red mix berry skin when immature; red when matured) from July to September in 2019. To verify the superiority of Swin-T-YOLOv5, its performance was compared against several commonly used object detectors, including Faster R-CNN, YOLOv3, YOLOv4, and YOLOv5. All models were assessed under different test conditions, including two different weather conditions (sunny and cloudy), two different berry maturity stages (immature and mature), and three different sunlight directions/intensities (morning, noon, and afternoon) for a comprehensive comparison. Additionally, the predicted number of grape bunches by Swin-T-YOLOv5 was further compared with ground truth values, including both in-field manual counting and manual labeling during the annotation process. Results showed that the proposed Swin-T-YOLOv5 outperformed all other studied models for grape bunch detection, with up to 97$\%$ of mean Average Precision (mAP) and 0.89 of F1-score when the weather was cloudy. This mAP was approximately 44$\%$, 18$\%$, 14$\%$, and 4$\%$greater than Faster R-CNN, YOLOv3, YOLOv4, and YOLOv5, respectively. Swin-T-YOLOv5 achieved its lowest mAP (90$\%$) and F1-score (0.82) when detecting immature berries, where the mAP was approximately 40$\%$, 5$\%$, 3$\%$, and 1$\%$ greater than the same. Furthermore, Swin-T-YOLOv5 performed better on Chardonnay variety with achieved up to 0.91 of R2 and 2.36 root mean square error (RMSE) when comparing the predictions with ground truth. However, it underperformed on Merlot variety with achieved only up to 0.70 of R2 and 3.30 of RMSE. 
\end{abstract}


\section{Introduction}
The overall grape production in the United States has reached 6.04 million tons in 2020, in which approximately 3.59 million tons (~59$\%$) were from wine grape production in California and Washington citation USDA. To maintain the premium quality of wine, wine vineyard needs to be elaborately managed so that the quantity and quality of the grapes can be well balanced for maximum vineyard profitability. Such vineyard management can be difficult, because the number of the berry bunches should be closely monitored by labors throughout the entire growth season to avoid a high volume of bunches overburdening the plant and thus the berry composition may not be optimal \cite{bellvert2021optimizing}. This presents significant challenges for wine vineyard owners and managers due to agricultural workforce shrinking and cost increasing. Potentially, this issue might be mitigated by leveraging the superiority of state-of-the-art computer vision technologies and data-driven artificial intelligence (AI) techniques \cite{zou2019object}.

Object detection is one of the fundamental tasks in computer vision, which is used for detecting instances of one or more classes of objects in digital images. Several common challenges, that prevent a target object from being successfully detected, include but are not limited to variable outdoor light conditions, scale changes of the objects, small objects, and partially occluded objects. In recent years, numerous deep learning-driven object detectors have been developed for various real-world tasks, such as fully connected networks (FCNs), convolutional neural networks (CNNs), and vision Transformer. Among these, CNN-based object detectors have demonstrated promising results \cite{gu2018liu}, \cite{jiaor}. Generally, CNN-based object detectors can be divided into two types, including one-stage detectors and two-stage detectors. Taking a few examples, one-stage detectors include Single Shot multibox Detector (SSD) \cite{liu2016ssd}, RetinaNet \cite{lin2017focal}, Fully Convolutional One-Stage (FCOS) \cite{tian2019fcos}, DEtection TRansformer (DETR) \cite{carion2020end}, EfficientDet \cite{tan2020efficientdet}, You Only Look Once (YOLO) family \cite{bochkovskiy2020yolov4}, \cite{redmon2018yolov3}, \cite{glenn20227002879}, \cite{wang2021scaled}, while two-stage detectors include Region-based CNN (R-CNN) \cite{girshick2014rich}, Fast/Faster R-CNN \cite{girshick2015fast}, \cite{ren2015faster}, Spatial Pyramid Pooling Networks (SPPNet)\cite{he2015spatial}, Feature Pyramid Network (FPN) \cite{lin2017feature}, and CenterNet2 \cite{duan2019centernet}.

As agriculture is being digitalized, both one-stage and two-stage object detectors have been widely applied on various orchard and vineyard scenarios, such as fruit detection and localization, with promising results achieved. Some of the major reasons, which made object detection challenging in agricultural environments, including severe occlusions from non-target objects (e.g., leaves, branches, trellis-wires, and densely clustered fruits) to target objects (e.g., fruit) \cite{gao2020multi}. Thus, in some cases, the two-stage detectors were preferred by the researchers due to their greater accuracy and robustness. Shuqin et al.\cite{tu2020passion} developed an improved model based on multi-scale Fast R-CNN that used both RGB (i.e., red, green, and blue) and depth images to detect passion fruit. Results indicated that the accuracy of the proposed model was improved from 0.850 to 0.931 (by ~10$\%$). Gao et al.\cite{gao2020multi} proposed a Faster R-CNN-based multi-class apple detection model for dense fruit-wall trees. It could detect apples under different canopy conditions, including non-occluded, leaf-occluded, branch/trellis-wire occluded, and fruit-occluded apple fruits, with an average detection accuracy of 0.879 across the four occlusion conditions. Additionally, the model processed each image in 241ms on average. Although two-stage detectors have shown robustness and promising detection results in agricultural applications, there are still two major concerns, the corresponding slow inference speed and high requirement of computational resources, to further implement them in the field. Therefore, it has become more popular nowadays to utilize one-stage detectors in identifying objects in orchards and vineyards, particularly using YOLO family models with their feature of real-time detection. 

Huang et al. \cite{huang2021immature} proposed an improved YOLOv3 model for detecting immature apples in orchards, using Cross Stage Partial (CSP)-Darknet53 as the backbone network of the model to improve the detection accuracy. Results showed that the F1-Score and mean Average Precision (mAP) were 0.652 and 0.675, respectively, for those severely occluded fruits. Furthermore, Chen et al. \cite{chen2021improved} also improved YOLOv3 model for cherry tomato detection, which adopted a dual-path network \cite{chen2017dual} to extract features. The model established four feature layers at different scales for multi-scale detection, achieving an overall detection accuracy of 94.29$\%$, Recall of 94.07$\%$, F1-Score of 94.18$\%$, and inference speed of 58 ms per image. Lu et al. \cite{lu2022canopy} introduced a convolutional block attention module (CBAM) \cite{woo2018cbam} and embedded a larger-scale feature map to the original YOLOv4 to enhance the detection performance on canopy apples in different growth stages. In general, YOLO family models tend to have a fast inference speed when testing on images. However, the detection accuracy in YOLO detectors might be influenced by the occlusion, which could cause loss of information during detection, in the canopy. Some strategies should be taken to compensate for this shortcoming of YOLO models. During the past two years, vision Transformer has demonstrated outstanding performances in numerous computer vision tasks \cite{khan2021transformers} and, therefore, are worth being further investigated to be employed together with YOLO models in addressing the challenges. 

A typical vision Transformer architecture is based on a self-attention mechanism that can learn the relationships between components of a sequence \cite{khan2021transformers}, \cite{li2022exploring}. Among all types, Swin-transformer is a novel backbone network of hierarchical vision Transformer, using a multi-head self-attention mechanism that can focus on a sequence of image patches to encode global, local, and contextual cues with certain flexibilities \cite{liu2021swin}. Swin-transformer has already shown its compelling records in various computer vision tasks, including region-level object detection \cite{liu2021swinnet}, pixel-level semantic segmentation \cite{hatamizadeh2022swin}, and image-level image classification \cite{jannat2022improving}. Particularly, it exhibited strong robustness to severe occlusions from foreground objects, random patch locations, and non-salient background regions. However, using Swin-transformer alone in object detections requires large computing resources as encoding-decoding structure of the Swin-transformer is different from the conventional CNNs. For example, each encoder of Swin-transformer contains two sublayers. The first sublayer is a multi-head attention layer, and the second sublayer is a fully connected layer, where the residual connections are used between the two sublayers. It can explore the potential of feature representation through a self-attention mechanism \cite{srinivas2021bottleneck}, \cite{zhao2020exploring}. Previous studies on public datasets, e.g., COCO \cite{lin2014microsoft}, have demonstrated that Swin-transformer outperformed other models on severely occluded objects \cite{naseer2021intriguing}. Recently, Swin-transformer has also been applied in agricultural field. For example, Wang et al. \cite{wang2021swingd} proposed “SwinGD” for grape bunch detection using Swin-transformer and Detection Transformer (DETR) models. Results showed that SwinGD achieved 94$\%$ of mAP, which was more accurate and robust in overexposed, darkened, and occluded field conditions. Zheng et al. \cite{zheng2022swin} researched a method for the recognition of strawberry appearance quality based on Swin-transformer and Multilayer Perceptron (MLP), or “Swin-MLP”, in which Swin-transformer was used to extract strawberry features and MLP was used to identify strawberry according to the imported features. Wang et al. \cite{wang2022practical} improved the backbone of Swin-transformer and then applied it to identify cucumber leaf diseases using an augmented dataset. The improved model had a strong ability to recognize the diseases with a 98.97$\%$ accuracy.

Although many models for fruit detection have been studied in orchards and vineyards \cite{lu2022canopy},\cite{fu2020application}, \cite{koirala2019deep}, and \cite{tian2019apple}, the critical challenges in grape detection in the field environment (e.g., multi-variety, multi-stage of growth, multi-condition of light source) have not yet been fully studied using a combined model of YOLOv5 and Swin-transformer. In this research, to achieve better accuracy and efficiency of grape bunch detection under dense-foliage and occlusion conditions in vineyards, we architecturally combined the state-of-the-art, one-stage detector of YOLOv5 and Swin-transformer (i.e., Swin-Transformer-YOLOv5 or Swin-T-YOLOv5), so that the proposed new network structure had the potential to inherently preserve the advantages from both models. The overarching goal of this research was to detect wine grape bunches accurately and efficiently under complex vineyard environment using the developed Swin-T-YOLOv5. The specific research objectives were to: 
\begin{enumerate}
            \item Combine the architectures of YOLOv5 and Swin-transformer to propose a novel network (i.e., Swin-T-YOLOv5) for wine grape bunch detection;
            \item Compare the performance of the developed Swin-T-YOLOv5 with other competitive object detectors, including Faster R-CNN, generic YOLOv3, YOLOv4, and YOLOv5;
            \item Investigate the results under different scenarios, including different wine grape varieties (i.e., white variety of Chardonnay and red variety of Merlot), weather conditions (i.e., sunny and cloudy), berry maturities or growth stages (i.e., immature and mature), and sunlight directions/intensities (i.e., morning, noon, and afternoon) in vineyards. 

        \end{enumerate}

\section{Materials and Methods}
\subsection{Data acquisition and preprocessing}
\subsubsection{Wine grape dataset}
\begin{figure}[H]
  \centering
  \subfigure[]{
  \includegraphics[width = 9cm,height = 5cm]{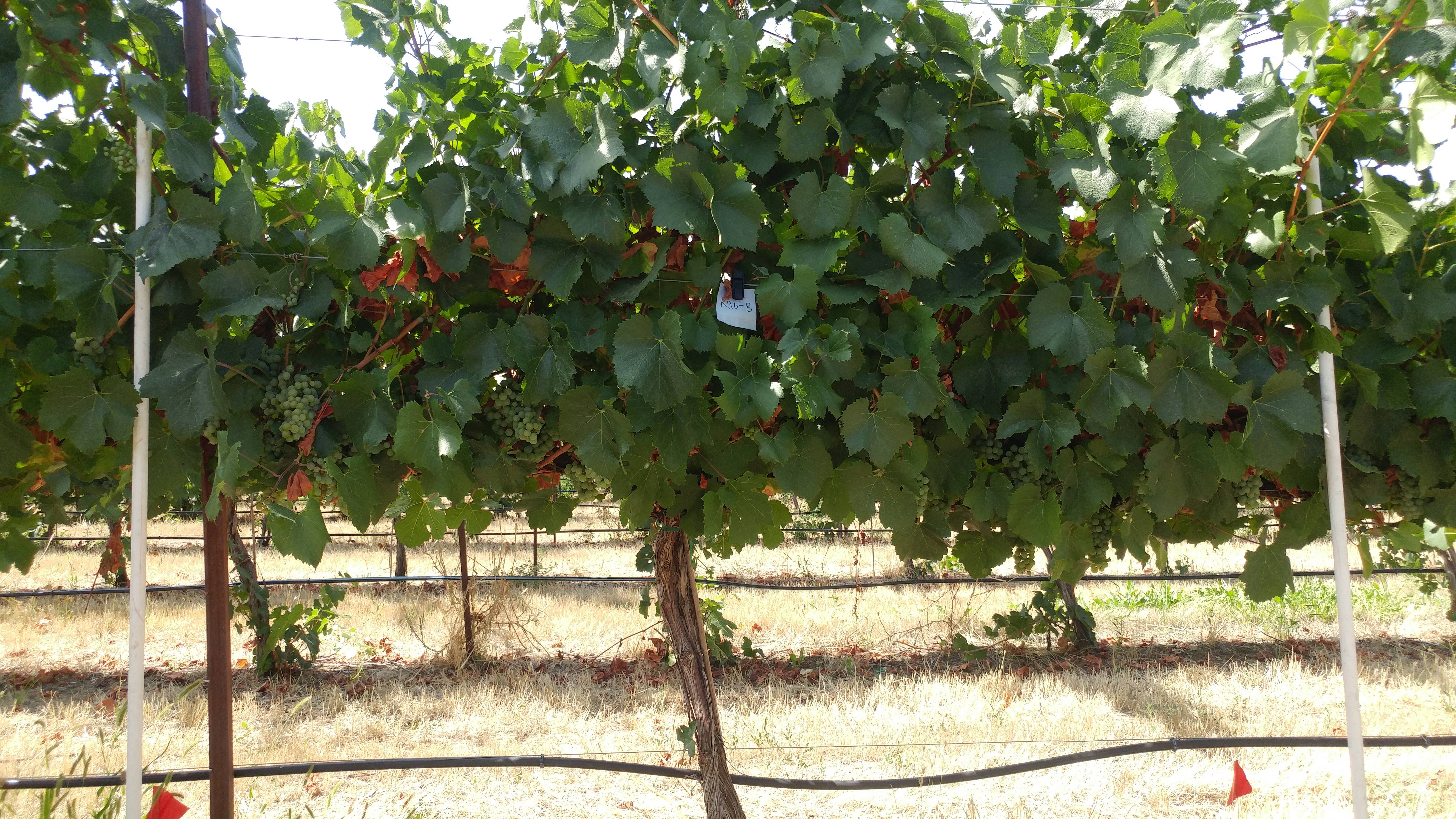} \label{fig.1a}
  }
  \subfigure[]{
  \includegraphics[width = 3cm,height = 5cm]{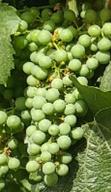} \label{fig.1b}
  }
  \subfigure[]{
  \includegraphics[width = 3cm,height = 5cm]{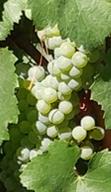} \label{fig.1c}
  }
    \subfigure[]{
  \includegraphics[width = 9cm,height = 5cm]{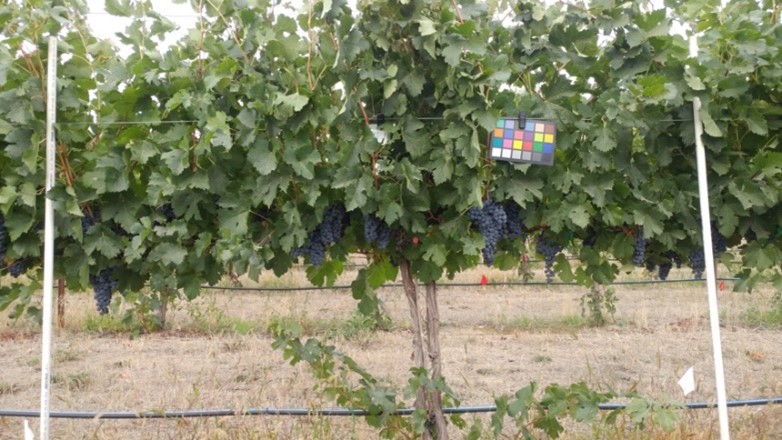} \label{fig.1d}
  }
  \subfigure[]{
  \includegraphics[width = 3cm,height = 5cm]{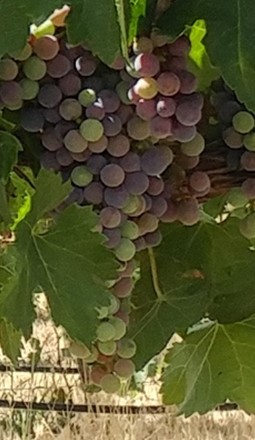} \label{fig.1e}
  }
  \subfigure[]{
  \includegraphics[width = 3cm,height = 5cm]{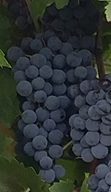} \label{fig.1f}
  }
  \caption{Grape dataset acquisition on (a) Chardonnay (white color of berry skin when matured; 10–33 grape bunches per plant) with the close-up views of grape bunch during (b) immature and (c) mature stages, and (d) Merlot (red color of berry skin when matured; 12–32 grape bunches per plant) with the close-up views of grape bunch during (e) immature and (f) mature stages.}
  \label{fig:fig1}
\end{figure}

The data acquisition and research activities in this study were carried out in a wine vineyard located in Washington State University (WSU) Roza Experimental Orchards, Prosser, WA. Two different wine grape varieties were selected as the target due to their distinct color of berry skin when matured, including Chardonnay (white berries; Figure \ref{fig.1a}) and Merlot (red berries; Figure\ref{fig.1d}). The color of berry skin for Chardonnay was consistently white throughout the growth season (Figure \ref{fig.1b}-\ref{fig.1c}), while the color for Merlot changed from white to red (Figure \ref{fig.1e}-\ref{fig.1f}) during the season. There were approximately 10-33 and 12-32 grape bunches per vine for the experimental Chardonnay and Merlot plants in this study. The wine vineyard was maintained by a professional manager for optimal productivity. The row and inter-plant spaces were about 2.5 m and 1.8 m for both varieties. 

The imagery data collection was completed using a Samsung Galaxy S6 smartphone (Samsung Electronics Co., Ltd., Suwon, South Korea) at the distances of 1-2 m, while the camera was facing perpendicularly to the canopy. The data collection was carried out during the entire growth season (i.e., from the berries were developed to matured) at a periodical frequency of one day per week and three times per day from 7/4/2019 to 9/30/2019. More specific details were given in Table \ref{t1} that the images of the canopies were captured under two weather conditions (i.e., sunny and cloudy), two berry maturity conditions (i.e., immature and mature), and three sunlight direction/intensity conditions (i.e., morning at 8am-9am, noon at 11am-12pm, and afternoon at 4pm-5pm, Pacific Daylight Time). All these various outdoor conditions largely represented the diversity of the imagery dataset. Note that all images were acquired from the consistent side of the canopy. As a result, 459 raw images were collected in total for Chardonnay (234 images) and Merlot (225 images) grape varieties in the original resolution of 5,312 x 2,988 pixels (Table \ref{t2}). Specific number of raw images under individual conditions can also be found in Table \ref{t1}.

\subsubsection{Dataset annotation and augmentation}
\begin{figure}[!h]
  \centering
  \subfigure[]{
    \includegraphics[width = 3cm,height = 6.5cm]{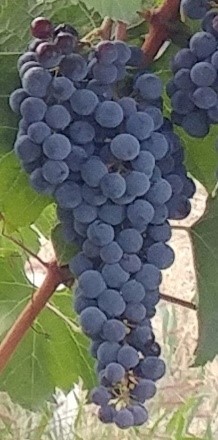} \label{fig.2a}
  }
  \subfigure[]{
  \includegraphics[width = 3cm,height = 6.5cm]{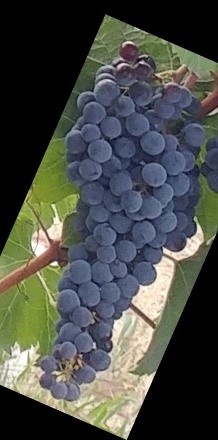} \label{fig.2b}
  }
  \subfigure[]{
  \includegraphics[width = 3cm,height = 6.5cm]{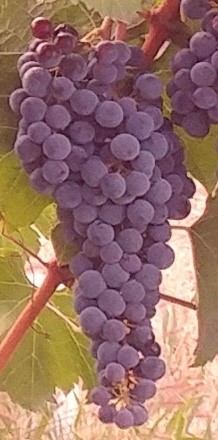} \label{fig.2c}
  }
    \subfigure[]{
  \includegraphics[width = 3cm,height = 6.5cm]{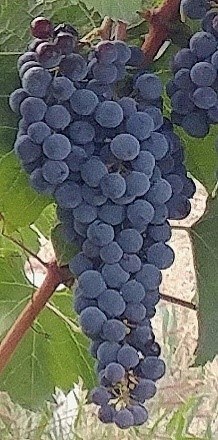} \label{fig.2d}
  }
  \subfigure[]{
  \includegraphics[width = 3cm,height = 6.5cm]{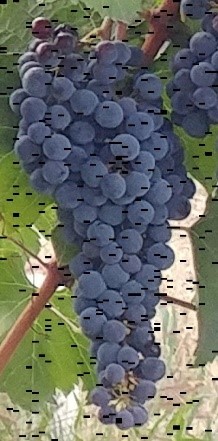} \label{fig.2e}
  }
  \caption{Illustrations of the dataset augmentation: (a) original image, (b) rotation, (c) channel enhancement, (d) Gaussian blur/noise, and (e) rectangle pixel discard.}
  \label{fig:fig2}
\end{figure}

The raw imagery dataset was manually annotated using the annotation tool of LabelImg \cite{labelimg2015}. The position of the grape bunch was individually selected using bounding boxes. Clustered grape bunches were also carefully separated. In addition, the “debar” approach was adopted based on our previous publication \cite{lu2022canopy} to separate individual canopies for evaluation purposes only. Once all raw images (Figure \ref{fig.2a}) were annotated, the dataset was further enriched by using data enhancement and augmentation library of Imgaug \cite{imgaug}, including the image adjustment of rotation (Figure \ref{fig.2b}), channel enhancement (Figure \ref{fig.2c}), Gaussian blur/noise (Figure \ref{fig.2d}), and rectangle pixel discard (Figure \ref{fig.2e}). During data augmentation, the annotated “key points” and “bounding boxes” were transformed accordingly. The enriched dataset can better represent the field conditions of the grape bunches. After augmentation, a dataset containing 4,418 images was developed, where a detailed description of the augmented dataset can be found in Table \ref{t2}. The actual number of augmented images were less than planned (4,590 images) because some invalid augmented images were identified and discarded. The finalized dataset was further divided into train (80$\%$), validation (10$\%$), and test sets (10$\%$), respectively, for development of grape bunch detection models. Finally, the in-field manual counting of grape bunches was completed during the harvest season on 10/1/2019 after the last dataset was acquired. 

\begin{table}[!h]
\caption{Grape dataset collected under different weather, berry maturity, and sunlight direction/intensity conditions.}
\center
\begin{tabular}{c c c c c}\hline
Grape variety & \multicolumn{2}{c}{Weather\textbackslash plant\textbackslash light condition} & Number of raw image & Total \\ \hline

\multirow{6}{*}{Chardonnay}& \multirow{2}{*}{Weather} & Sunny& 169 & \multirow{2}{*}{234}   \\ 
&  & Cloudy  & 65   \\ \cline{3-5} 
 &\multirow{2}{*}{Berry maturity} & Immature (white) & 83 &  \multirow{2}{*}{$155^{e}$}   \\ 
 &  & Mature (white)  & 72   \\ \cline{3-5} 
  &\multirow{3}{*}{Sunlight direction$^{a}$} & Morning$^{b}$ & 91 &  \multirow{3}{*}{234}   \\ 
 &  & Noon$^{c}$  & 75   \\ 
 &  & Afternoon$^{d}$  & 68   \\ \cline{2-5} 
 
 \multirow{6}{*}{Merlot}& \multirow{2}{*}{Weather} & Sunny& 153 & \multirow{2}{*}{225}   \\ 
&  & Cloudy  & 72   \\ \cline{3-5} 
 &\multirow{2}{*}{Berry maturity} & Immature (white)& 81 &  \multirow{2}{*}{162}   \\ 
 &  & Mature (white)  & 82   \\ \cline{3-5} 
  &\multirow{3}{*}{Sunlight direction} & Morning & 82 &  \multirow{3}{*}{225}   \\ 
 &  & Noon  & 70   \\ 
 &  & Afternoon  & 73   \\ \cline{1-5} 
 \multicolumn{5}{l}{$^{a}$All images in this study were taken from the consistent side of the canopy.}\\
\multicolumn{5}{l}{$^{b}$Morning at 8am-9am (in the direction of the light).}\\
\multicolumn{5}{l}{$^{c}$Noon at 11am-12pm (maximum solar elevation angle).}\\
\multicolumn{5}{l}{$^{d}$Afternoon at 4pm-5pm (against the direction of the light).}\\
\multicolumn{5}{l}{$^{e}$All images were periodically collected during 7/4/2019–9/30/2019, in which the dates were divided into}\\
\multicolumn{5}{l}{\quad  three growth stages including early stage (7/4/2019–7/27/2019), middle stage (8/2/2019–8/24/2019), and}\\
\multicolumn{5}{l}{\quad  late (harvest) stage (9/7/2019–9/30/2019). During the middle stage, the change of shape and color of the grapes} \\
\multicolumn{5}{l}{\quad  was inconsiderable.Therefore, we only compared the early stage (immature) and late stage (mature).}\\
\end{tabular}\\
\begin{flushleft}

\label{t1}
\end{flushleft}
\end{table}

\begin{table}[!h]
\caption{Grape imagery dataset and augmentation in this study.}
\center
\begin{tabular}{c c c c c c c }\hline
Dataset & Variety & \makecell{Color of \\ berry skin} & \makecell{Original image \\ size (pixels)} & \makecell{Planned \\ number  of \\ images (after \\ augmentation)}  & \makecell{Planned \\ number of \\ images (after \\ augmentation)} & \makecell{Actual number \\ of images \\ (after augmentation)} \\ \hline
\multirow{2}{*}{Grape} & Chardonnay & White & \multirow{2}{*}{5,312 x 2,988} & 234 & 2,340&2,263\\
&Merlot& Red& &225& 2,250& 2,155\\ \cline{1-4}
\multicolumn{4}{c}{Total}&459&4,590$^a$&4,418$^b$\\ \hline

\multicolumn{7}{l}{\makecell[l]{$^{a}$It was planned that there will be additional nine images augmented for each raw image using the augmentation \\\quad methods illustrated in Figure \ref{fig:fig2}}}.\\

\multicolumn{7}{l}{\makecell[l]{$^{b}$The actual augmented images were less than what was planned because all augmented images were examined, \\\quad and those invalid images were discarded.}} \\

\end{tabular}
\begin{flushleft}

\label{t2}
\end{flushleft}
\end{table}
\subsection{Grape bunch detection network}
\subsubsection{Swin-transformer}
\begin{figure}[!h]
  \centering
  \subfigure[]{
  \includegraphics[width = 5cm,height = 8cm]{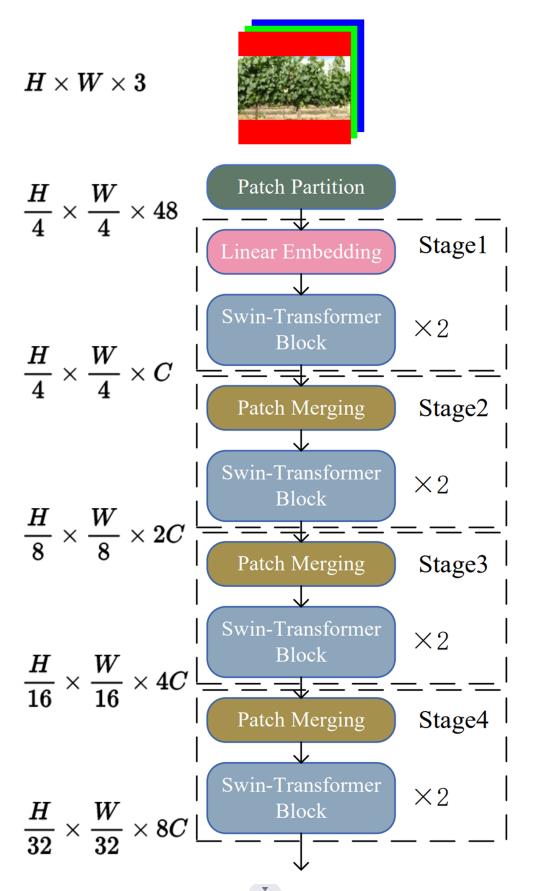} \label{fig:fig.3a}
  }
  \subfigure[]{
  \includegraphics[width = 10cm,height = 3.8cm]{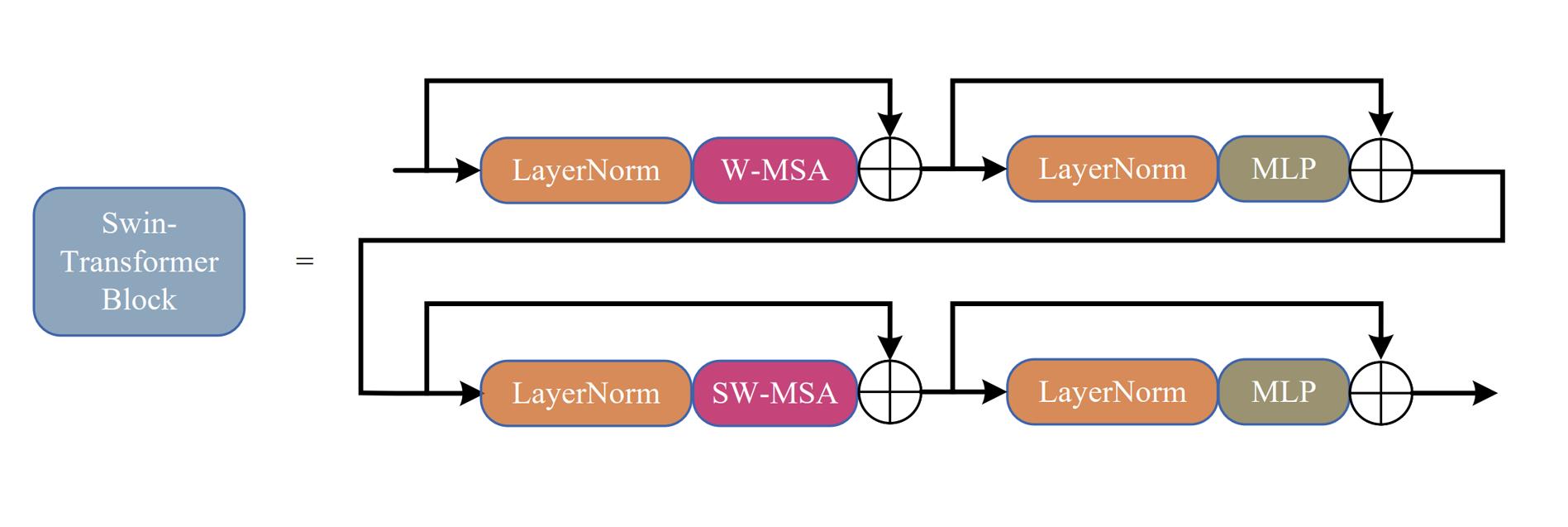} \label{fig:fig.3b}
  }
  \caption{(a) Overall architecture of Swin-transformer (H and W refer to height and width of the images; C refers to the number of feature channels) and (b) two successive Swin-transformer blocks (W-MSA and SW-MSA refer to Window-based Multi-head Self Attention and Shifted Window-based MSA; MLP refers to Multi-Layer Perceptron).}
  \label{fig:fig3}
\end{figure}

First, the Swin-transformer architecture \cite{liu2021swin} was introduced in Figure \ref{fig:fig.3a}. It can split the input RGB image into non-overlapping, small patches through a patch partition module. Each patch was treated as a “token” whose features were set as the concatenation of the raw pixel values in the RGB image (i.e., 3 channels). In this study, a patch size of 4 × 4 was used and, therefore, the feature dimension per patch was 4×4×3 = 48. A linear embedding layer was then applied to this raw value feature to project it to an arbitrary dimension (denoted as C in \ref{fig:fig.3a}). Swin-transformer was built through replacing the standard Multi-head Self Attention (MSA) module in a regular Transformer block by a MSA module based on “windows” (i.e., W-MSA) and “shifted windows” (i.e., SW-MSA), while other layers kept the same (\ref{fig:fig.3b}). This module was followed by a 2-layer Multi-Layer Perceptron (MLP) with nonlinearity of Rectified Linear Unit (ReLU) in between. A Normalization Layer (LayerNorm) and a residual connection were applied before and after each MSA module and MLP layer. 

\subsubsection{YOLOv5}
YOLOv5 (specifically, YOLOv5s) is a recent detection model in YOLO family \cite{glenn20227002879}, which has fast inference (detection) speed with up to 140 frames per second (fps). In addition, YOLOv5s is a lightweight model with fewer model parameters, which is approximately 10$\%$ of the generic YOLOv4, indicating that this model might be more suitable for deployment on embedded devices for real-time object detection. Combined with all these advantages, this study attempted to detect grape bunches in dense canopies using the improved YOLOv5. 

\begin{figure}[!h]
  \centering
  \subfigure[]{
  \includegraphics[width = 12cm,height = 8cm]{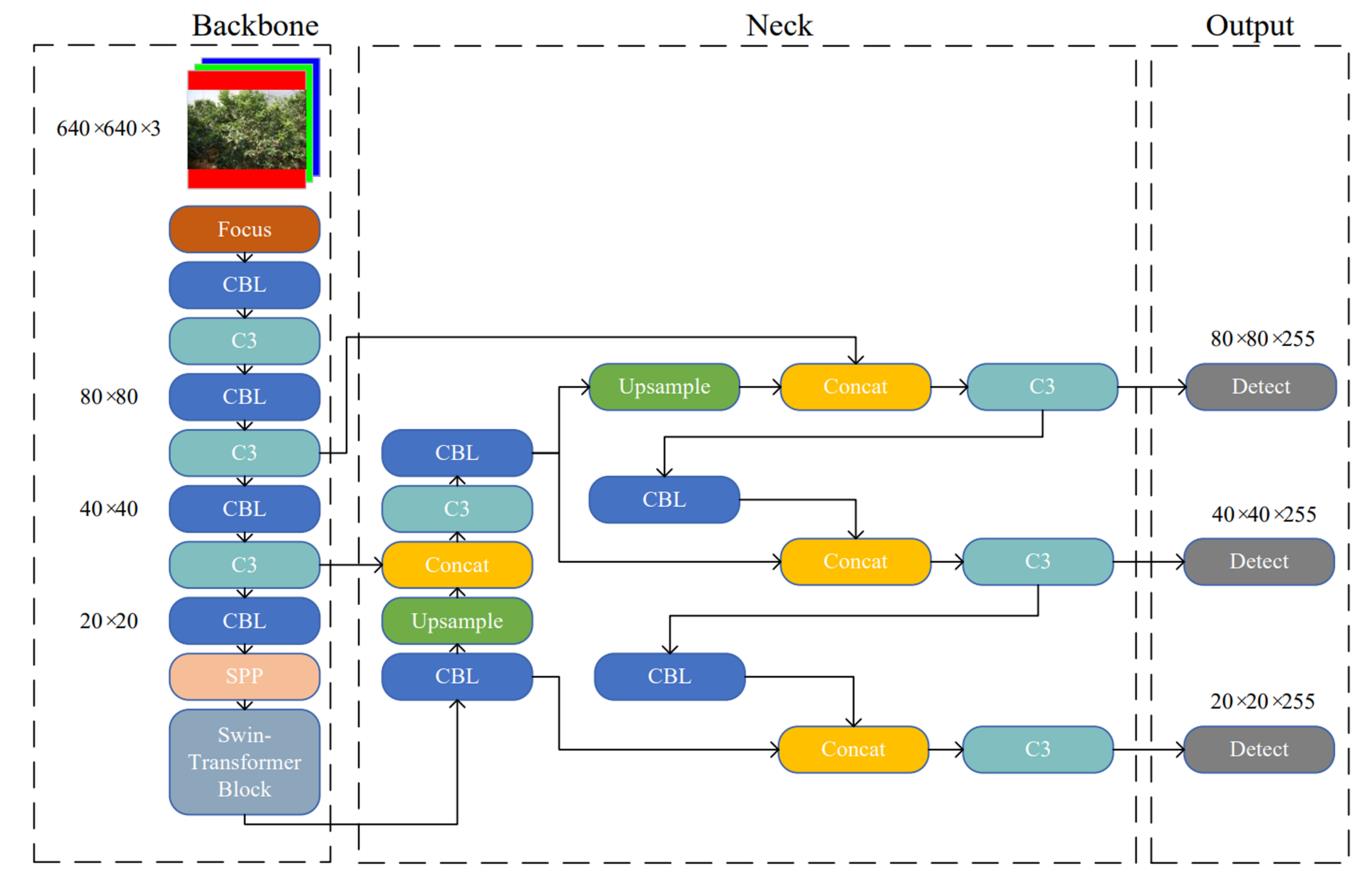} \label{fig.4a}
  }
  \subfigure[]{
  \includegraphics[width = 6cm,height = 2.8cm]{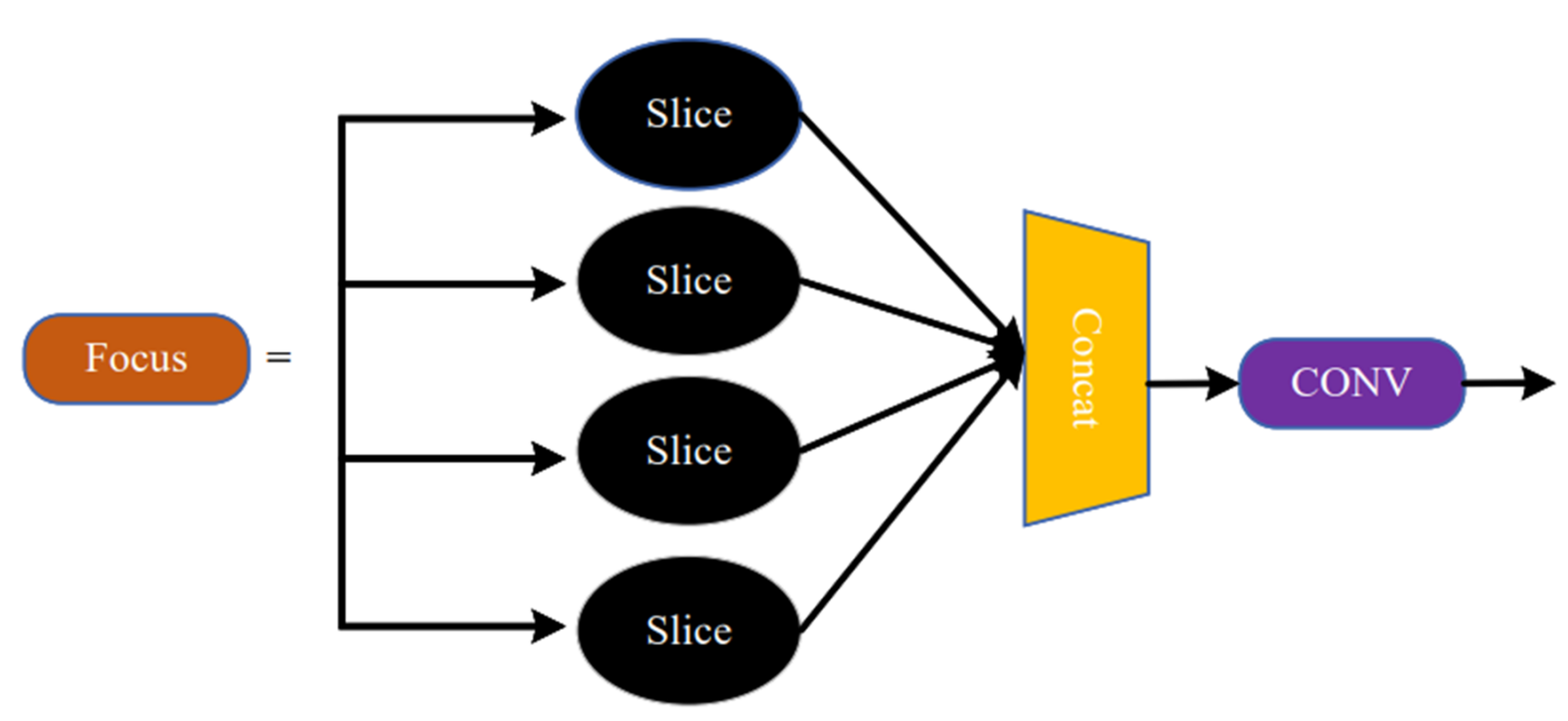} \label{fig.4b}
  }
   \subfigure[]{
  \includegraphics[width = 6cm,height = 2.8cm]{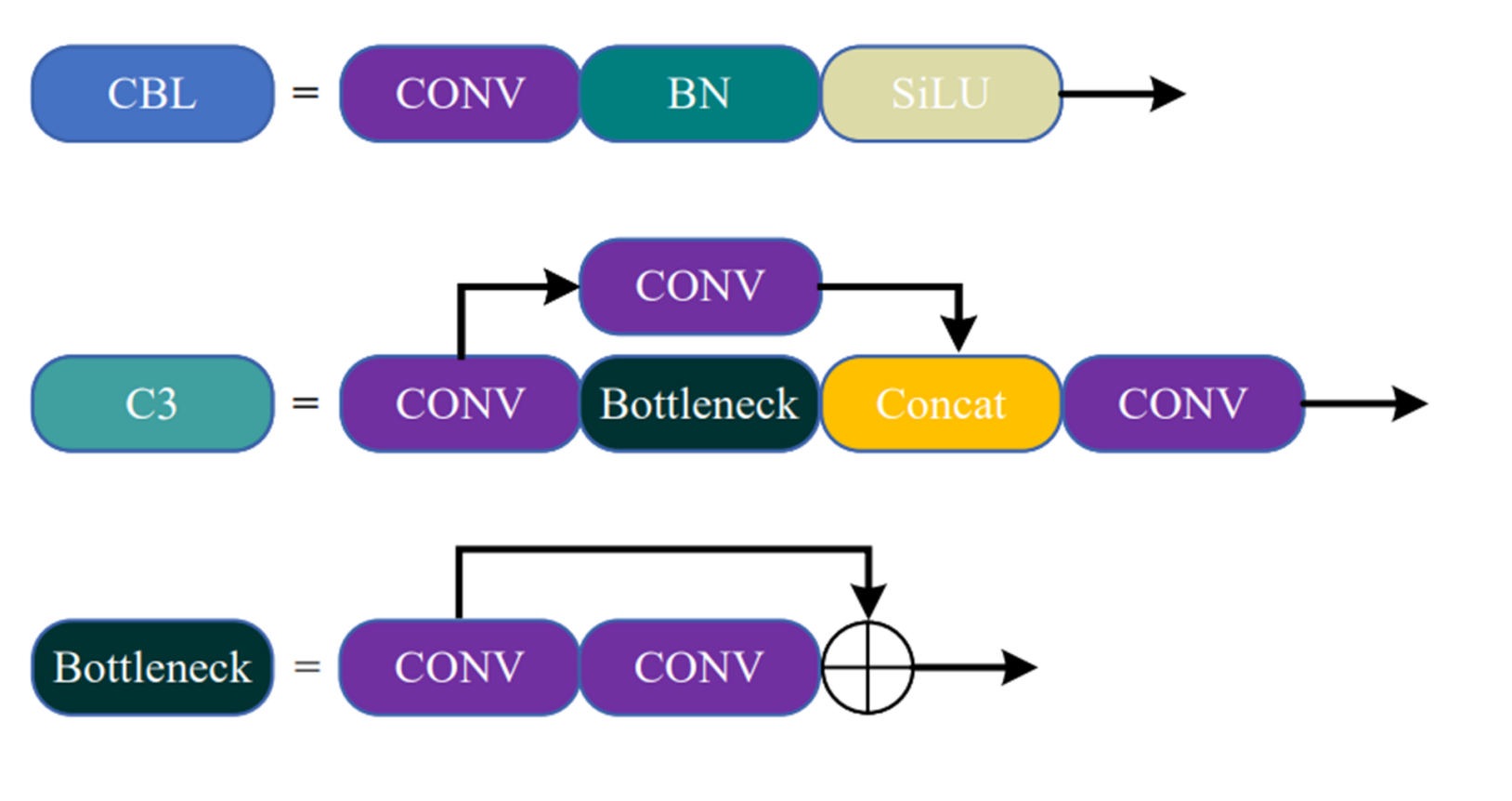} \label{fig.4c}
  }
   \subfigure[]{
  \includegraphics[width = 6cm,height = 2.8cm]{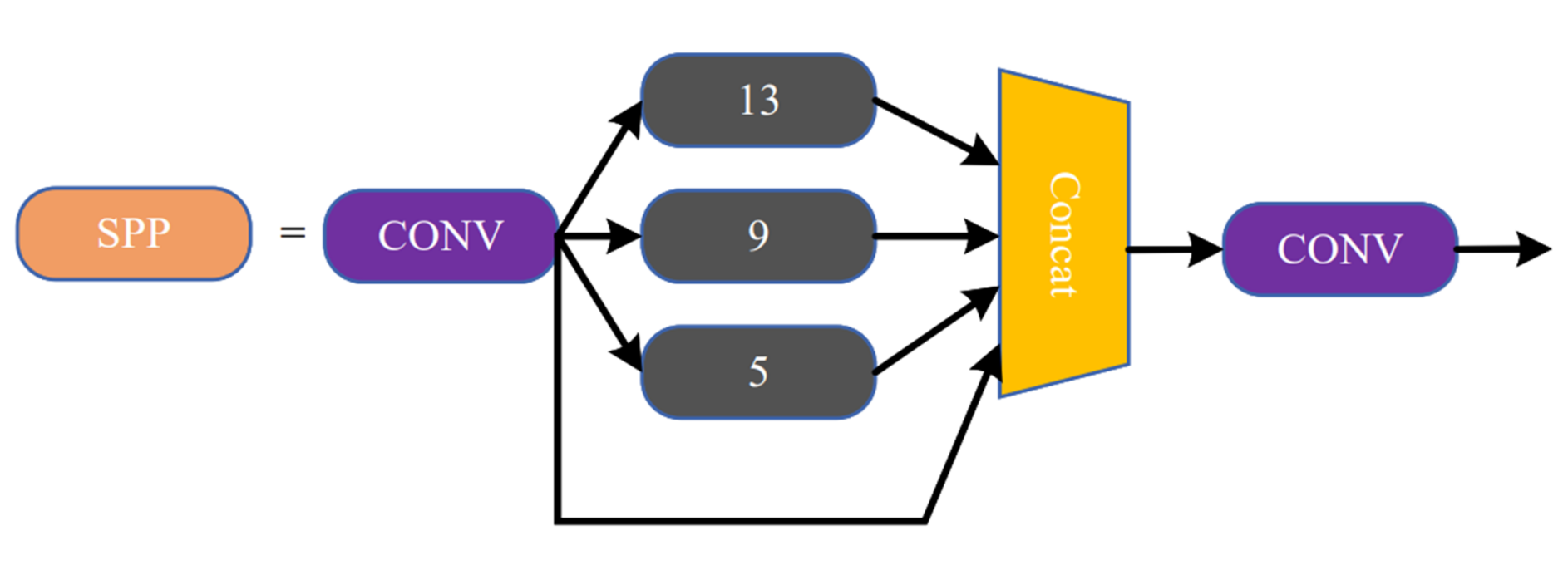} \label{fig.4d}
  }
  \caption{(a) Integrated architecture of Swin-transformer-YOLOv5, layers of (b) Focus, (c) CBL and Cross Stage Partial (CSP) Bottleneck with 3 convolutions (C3), and (d) Spatial Pyramid Pooling (SPP), where CONV, Concat, BN, SiLU, and ADD () refer to Convolutional, Concatenate, Batch Normalization, activation function of Sigmoid Linear Unit, and feature fusion with the number of channels unchanged. }
  \label{fig:fig4}
\end{figure}

In general, YOLOv5 framework includes three parts: backbone, neck, and detection (or output) networks (Figure \ref{fig:fig4}).The backbone network was used to extract feature maps from the input images with multiple convolutions and merging. A three-layer feature map was then generated in the backbone network in the sizes of 80×80, 40×40, and 20×20 (Figure \ref{fig.4a}; left). After backbone network, the neck network contained a series of feature fusion layers that can mix and combine image features. All feature maps in different sizes generated by the backbone network were fused to obtain more context information and reduce the information loss. The characteristic pyramid structure of Feature Pyramid Network (FPN) and Path Aggregation Network (PANet) were adopted during the merging process, where strong semantic features were transferred from top to bottom feature maps using FPN structure. Meanwhile, strong localization features were transferred from lower to higher feature maps using PANet. Overall, the ability of feature fusion in the neck network was enhanced by using FPN and PANet together (Figure \ref{fig.4a}; middle). Finally, the detection network was used to give the detection results. It consisted of three detection layers, with the corresponding output feature maps of 80×80, 40×40, and 20×20, which was used to detect objects in the input images. Each detection layer ultimately can output a 21-channel vector and then generate and mark the predicted bounding box and category of the target in the original input images for final detections (Figure \ref{fig.4a}; right).

Moreover, the Focus module of YOLOv5 can slice and concatenate images (Figure \ref{fig.4b}), which was designed to reduce the computational load of the model and speed up the training process. It can first split the input 3-channel image into four slices using the Slice operation. The four slices were concatenated using the Concat operation, and a Convolutional layer (CONV) was then used to generate the output feature map. Figure \ref{fig.4c} gave the explanations on some layers/modules in backbone network, including CBL and C3, in which CBL was a standard convolutional module consisting of CONV, Batch Normalization (BN), and activation function of Sigmoid Linear Unit (SiLU); C3 was Cross Stage Partial (CSP) Bottleneck with 3 CONVs. The initial input was split into two branches, and thus the number of channels of the feature maps were halved by CONV operation in each branch. The output feature maps of the two branches were again connected through the Concat operation. The final output feature map of C3 was generated by CONV. C3 was used to improve inference (test) speed by reducing the size of the model, while maintaining desired performances in extracting useful features from images. Finally, Spatial Pyramid Pooling (SPP) module was used to improve the receptive field by converting feature maps of arbitrary size into feature vectors of fixed size (Figure \ref{fig.4d}). The feature map was first output through a CONV layer with the kernel size of 1×1. It was then connected with the output feature map subsampled by three parallel max pooling layers, followed by a CONV layer to output the final feature map.

\subsubsection{Integration of Swin-transformer and YOLOv5}
To take advantages of both Swin-transformer and YOLOv5, two models were integrated (i.e., Swin-transformer-YOLOv5 or Swin-T-YOLOv5) by replacing the last C3 layer (i.e., with CSP Bottleneck and three CONVs) in the original YOLOv5 with Swin-transformer encoder blocks (Figure \ref{fig.4a}). Because the resolution of feature maps was 20×20 at the end of the backbone network, applying Swin-transformer on low-resolution feature maps can reduce computational load and save memory space. This integration can compensate for the shortcoming of YOLOv5 as one of the typical CNNs in lack of capturing global and contextual information due to the limited receptive field \cite{oksuz2020imbalance}, while Swin-transformer can be used to capture long-distance dependencies and retain different local information \cite{liu2021swin}. Therefore, our proposed scheme combined YOLOv5s and Swin-transformer, so that the new structure can inherit their advantages and preserve both global and local features. Furthermore, the self-attention mechanism was used to improve the detection accuracy of the integrated model. This integration might be particularly useful for the occluded grape bunches in dense foliage vineyard canopies. Pre-trained YOLOv5s using COCO dataset was adopted during training to improve the generalization ability of the proposed network.
\begin{table}[!h]
\caption{Major hyper-parameters used in this study for YOLOv3, YOLOv4, YOLOv5, and Swin-transformer-YOLOv5 (Swin-T-YOLOv5).}
\center
\begin{tabular}{c c c c c c}\hline
Hyper-parameter&\makecell{Faster R-\\CNN}&YOLOv3&YOLOv4&YOLOv5&\makecell{Swim-T-\\YOLOv5}\\ \hline
Optimization algorithm &SGD$^a$ &SGD &SGD &SGD &SGD\\
Initial learn rate &$1\times10^{3}$ &$1\times10^{3} $&$1\times10^{-3}$ &$1\times10^{-2}$&$1\times10^{-2}$\\
Mini-batch size &256 &16 &16 &32 &32\\
Number of epochs &100 &100 &100 &100 &100\\
\makecell{Intersection over Union (IoU)\\(train and validation)} &0.3 &0.213 &0.213 &0.2 &0.2\\
Weight decay &$5\times10^{-4}$& $5\times10^{-4}$ &$5\times10^{-4}$ &$5\times10^{-4}$ &$5\times10^{-4}$ \\ \hline
\multicolumn{6}{l}{\textit{a}: SGD refers to Stochastic Gradient Descent.}
\end{tabular}

\label{t3}

\end{table}

The training, validation, and testing steps were carried out on a workstation with an Intel® Xeon® Silver 4114 CPU, 64 GB RAM, NVIDIA RTX3090 GPU (24 GB VRAM), and Ubuntu 20.04 LTS Operating System. Python was used to write program code and call required libraries, such as CUDA, cuDNN, and OpenCV, on top of PyTorch 1.8.1 framework. To comprehensively evaluate the performances, our proposed Swin-T-YOLOv5 was compared against Faster R-CNN \cite{ren2015faster}, YOLOv3 \cite{redmon2018yolov3}, YOLOv4 \cite{bochkovskiy2020yolov4}, and YOLOv5 \cite{glenn20227002879}, where the training hyperparameters of each model were shown in Table \ref{t3}.

\subsubsection{Evaluation metrics}
The performance of each model was evaluated using its \textit{Precision (P)}, \textit{Recall (R)}, \textit{F1-score}, \textit{mean Average Precision (mAP)} (Equations (1-4)), and inference (detection) speed per image, in which \textit{mAP} served as a key metric to assess the overall performance of a model:
\begin{equation}
P = \frac{TP}{TP+FP}  
\end{equation}
\begin{equation}
R =\frac{TP}{TP+FN} 
\end{equation}
\begin{equation}
  F1 = \frac{2\times P\times R}{P+R}  
\end{equation}
\begin{equation}
  mAP = \frac{1}{n} \sum_{k=1}^{k=n} {AP_k}
\end{equation}

where True Positives (TP) represent the positive samples correctly predicted by the model, True Negatives (TN) represent the negative samples correctly predicted by the model, False Positives (FP) represent the positive samples incorrectly predicted by the model, and False Negatives (FN) represent the negative samples incorrectly predicted by the model;\textit{$AP_{k}$} represents the \textit{AP} of class \textit{k};\textit{n}represents the number of classes.

Additionally, \textit{$r^{2}$}and \textit{root mean square error} (\textit{RMSE};Eation (5)) was adopted to compare the results predicted by the models and ground truth data from both in-field manual counting and manual labeling:
\begin{equation}
  RMSE = \sqrt{\frac{\sum_{i=1}^{N}{(x_{i}-\hat{x}_i)^{2}}} {N}}
\end{equation}
where i represents the variable, N represents the number of data points (plants), \textit{$x_{i}$} represents the actual count of grape bunches (in-field or label),  \textit{$\hat{x}$} represents the estimated count of grape bunches using Swin-T-YOLOv5. The P-R curves were also used for visually demonstrating the performance of the models, where P and R were shown on vertical and horizontal axes, respectively. The Intersection over Union (IoU) and the confidence score were both set to 0.5 for test set.

\section{Result}
\subsection{Swin-transformer-YOLOv5 training and validation}
\begin{figure}[!h]
  \centering
  
  \includegraphics[width = 12cm,height = 12cm]{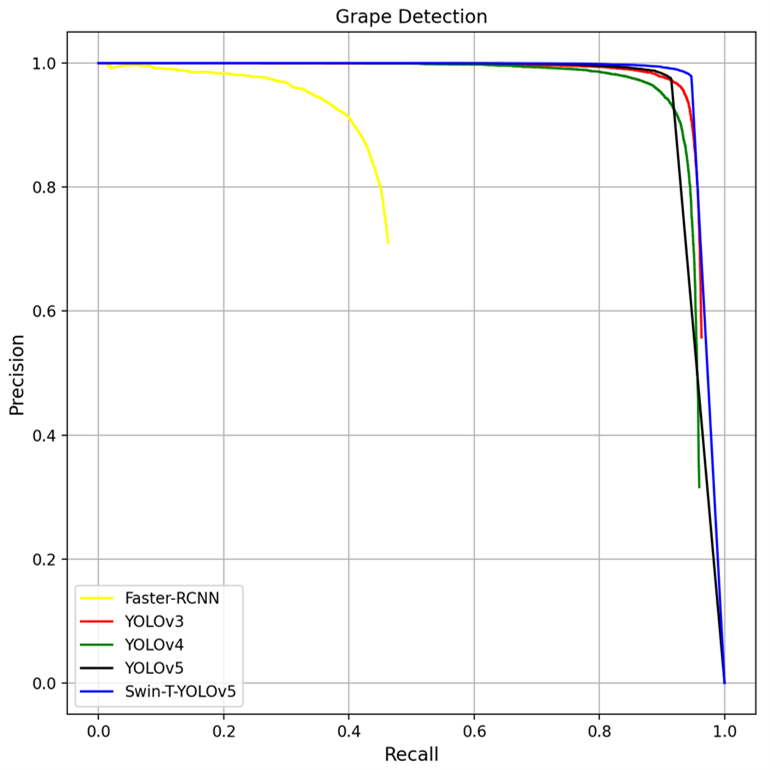} 
  
  \caption{Precision-Recall (P-R) curves of Faster R-CNN, YOLOv3, YOLOv4, YOLOv5, and Swin-transformer-YOLOv5 (Swin-T-YOLOv5) in detecting grape bunches. }
  \label{fig5}
\end{figure}

All models were trained and validated for a comprehensive comparison. Table \ref{t4} shows the detailed comparison results using the previously defined evaluation metrics. Overall, our proposed Swin-T-YOLOv5 outperformed all other models, with the mAP of 97.4$\%$, F1-score of 0.96, and inference speed of 13.2 ms per image. The mAP of Swin-T-YOLOv5 was 34.8$\%$, 2.1$\%$, 3.2$\%$, and 2.1$\%$ better than Faster R-CNN, YOLOv3, YOLOv4, and YOLOv5, respectively. Although its inference time was slightly slower (1.4 ms) than the original YOLOv5, it was still faster than the rest of the models by 0.6-336.8 ms per image. Moreover, P-R curves were given in Figure \ref{fig5} that Swin-T-YOLOv5 had the curve (in blue color) that was nearest to the upper-right point, indicating the best performance among all models. While Faster R-CNN (in yellow color) performed the worst. 
\pagebreak
\begin{table}[!h]
\caption{Comparison of training and validation results for Faster R-CNN, YOLOv3, YOLOv4, YOLOv5, and Swin-transformer-YOLOv5 (Swin-T-YOLOv5).}
\center
\begin{tabular}{c c c c c c}\hline
Network& Precision($\%$)& Recall($\%$)& \makecell{\textit{mAP}$^{a}$\\ \textit{(IoU = 0.5)}}& F1-score & \makecell{Inference speed \\per image (ms)$^{b}$}\\ \hline
Faster R-CNN& 60.1 &59.1 &62.6 &0.59 &350.0\\
YOLOv3 &96.1 &93.4 &95.3 &0.94 &13.8\\
YOLOv4 &75.5 &92.8 &94.2 &0.82 &14.2\\
YOLOv5 &96.6 &91.4 &95.3 &0.94 &\textbf{11.8}\\ 
\textbf{Swin-T-YOLOv5}& \textbf{97.9}& \textbf{94.7}& \textbf{97.4} &\textbf{0.96} &13.2\\ \hline
\multicolumn{5}{l}{$^{a}$mAP refers to mean Average Precision.}\\
\multicolumn{5}{l}{$^{b}$Inference time may vary depending on the hardware configurations.}\\
\end{tabular}

\label{t6}

\label{t4}

\end{table}

\subsection{Swin-transformer-YOLOv5 testing}
\subsubsection{Testing under two weather conditions}
All models were tested under different conditions as listed in Table \ref{t1}, including two weather conditions (i.e., sunny and cloudy), two berry maturity stages (i.e., immature and mature), and three sunlight directions/intensities (i.e., morning, noon, and afternoon), to verify the superiority of the proposed Swin-T-YOLOv5. Detailed model comparison results were given in Table \ref{t5} using the test set under two weather conditions for both grape varieties of Chardonnay and Merlot. Compared to Faster R-CNN, YOLOv3, YOLOv4, and YOLOv5, Swin-T-YOLOv5 achieved the best performance under both conditions in terms of mAP (95.36$\%$-97.19$\%$) and F1-score (0.86-0.89). Swin-T-YOLOv5 performed slightly better under cloudy sky condition with higher mAP (+1.83$\%$) and F1-score (+0.03) compared with sunny sky condition. While the inference speed of Swin-T-YOLOv5 (13.2 ms per image) was not the best among all, it was 1.4 ms slower than YOLOv5 only.

\begin{figure}[!h]
  \centering
  
  \includegraphics[width = 15cm,height = 15cm]{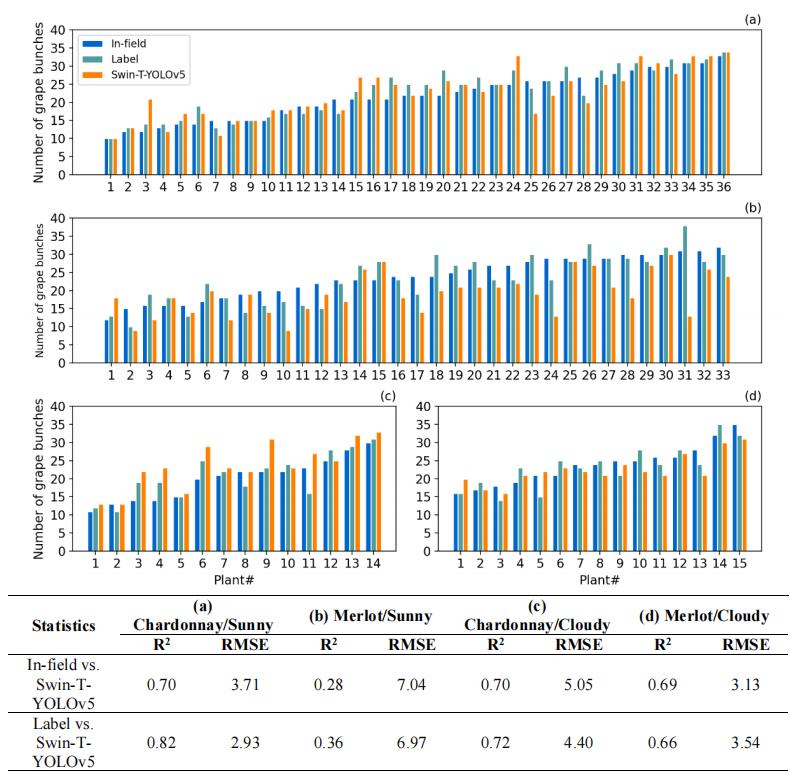} 
  
  \caption{The number of grape bunches comparison between in-field manual counting (gTruth), manual label, and detection using Swin-transformer-YOLOv5 (Swin-T-YOLOv5) with (a) Chardonnay (sunny), (b) Merlot (sunny), (c) Chardonnay (cloudy), and (d) Merlot (cloudy). RMSE refers to root mean square error.}
  \label{fig6}
\end{figure}

Being proven that Swin-T-YOLOv5 outperformed all other models under both sunny and cloudy sky conditions, we further compared it against the ground truth data from in-field manual counting and manual labeling (Figure\ref{fig6}) for Chardonnay and Merlot, respectively. Results showed that Swin-T-YOLOv5 performed well with Chardonnay variety under both weather conditions with 0.70-0.82 of R2 and 2.93-5.05 RMSE when the predicted results were compared against both in-field and label counts (Figure \ref{fig6}a and \ref{fig6}c). It also worked well with Merlot under cloudy condition (Figure \ref{fig6}d), however, R2 dropped to 0.28-0.36 and RMSE increased to 6.97-7.04 on Merlot under sunny condition (Figure \ref{fig6}b), indicating greater detection errors. Demonstrations of detection results under two weather conditions were provided in Figures \ref{figa1}-\ref{figa2} in Appendices. 
\begin{table}[!h]
\caption{Model comparison using the test set under two weather conditions.}
\center
\begin{tabular}{c c c c c c}\hline
Network&\makecell{Weather \\Condition}& \makecell{\textit{mAP}$^{a}$\\ \textit{(IoU = 0.5)}}& F1-score & \makecell{Inference speed \\per image (ms)$^{b}$}\\ \hline
Faster R-CNN& \multirow{5}{*}{Sunny} &59.23 &0.63 &350\\
YOLOv3 & &84.47 &0.64 &13.8\\
YOLOv4 & &90.43 &0.68 &14.2\\
YOLOv5 & &92.16 &0.82  &\textbf{11.8}\\ 
\textbf{Swin-T-YOLOv5}& & \textbf{95.36} &\textbf{0.89} &13.2\\ \hline
Faster R-CNN& \multirow{5}{*}{Cloudy} &53.54 &0.67 &350\\
YOLOv3 & &78.93 &0.72 &13.8\\
YOLOv4 & &83.45 &0.76 &14.2\\
YOLOv5 & &93.64 &0.83  &\textbf{11.8}\\ 
\textbf{Swin-T-YOLOv5}& & \textbf{97.19} &\textbf{0.89} &13.2\\ \hline
\multicolumn{5}{l}{$^{a}$mAP refers to mean Average Precision.}\\
\multicolumn{5}{l}{$^{b}$Inference time may vary depending on the hardware configurations.}\\
\end{tabular}

\label{t5}
\end{table}

\pagebreak

\subsubsection{Testing at two maturity stages}
In addition to two different weather conditions, we compared the performances of Swin-T-YOLOv5 with all other models at two berry maturity stages, including immature and mature berries for Chardonnay (i.e., white color of berry skin throughout the growth season) and Merlot (i.e., white or white-red mix when immature; red color when matured) (Figure \ref{fig:fig1}). Detailed comparison results were given in Table \ref{t6} that, as expected, Swin-T-YOLOv5 outperformed all other models at both berry maturity stages with 90.31$\%$-95.86$\%$of mAP and 0.82-0.87 of F1-score. Clearly, all detectors achieved better detection results at the mature stage, including Swin-T-YOLOv5 (5.55$\%$ higher in mAP and 0.05 higher in F1-score), when the berries were larger (i.e., less occlusions) and with more distinct color than their background, such as leaves. Comparing to the second-best model, YOLOv5 (mAPs of 89.78$\%$-91.58$\%$), the performance of Swin-T-YOLOv5 was improved more at the berry mature stage (+4.28$\%$) than immature stage (+0.53$\%$), indicating that the improvements of the model were more effective to those ready-to-harvest grape bunches. 

\begin{figure}[!h]
  \centering
  
  \includegraphics[width = 15cm,height = 15cm] {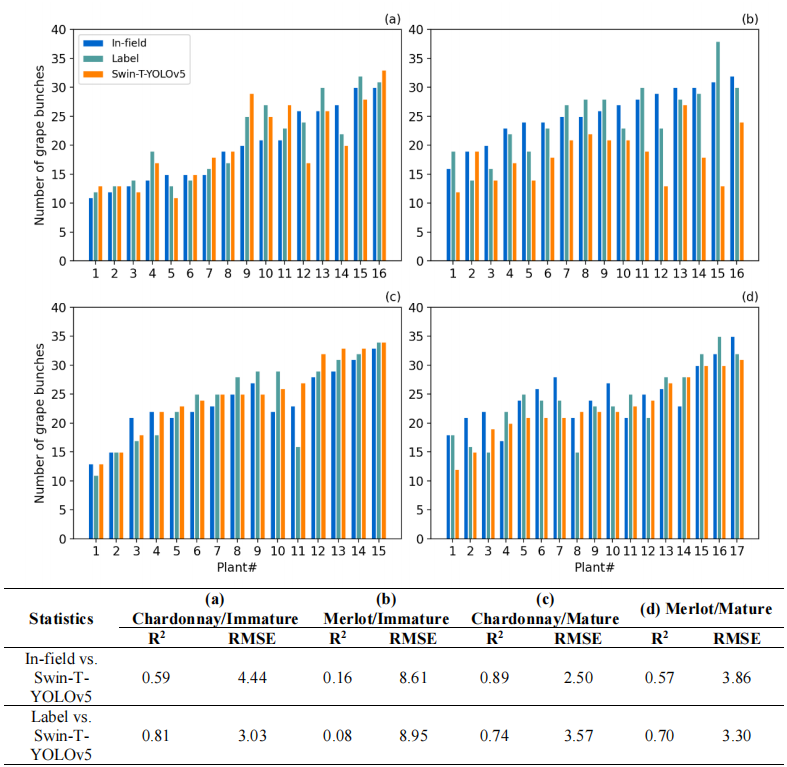} 
  
  \caption{The number of grape bunches comparison between in-field manual counting (gTruth), manual label, and detection using Swin-transformer-YOLOv5 (Swin-T-YOLOv5) for (a-b) immature and (c-d) mature berries with Chardonnay (left) and Merlot (right). RMSE refers to root mean square error.}
  \label{fig7}
\end{figure}

Figure \ref{fig7} compared the specific predicted number of grape bunches using Swin-T-YOLOv5 against the ground truth data of both in-field manual counting and manual labeling on both Chardonnay and Merlot. As observed in Table \ref{t6}, R2 was higher (0.57-0.89) and RMSE was smaller (2.50-3.86) for those matured berries (Figure \ref{fig7}c-\ref{fig7}d). Swin-T-YOLOv5 did a poor job on Merlot when the berries were immature (i.e., white or white-red mixed berries) with 0.08-0.16 of R2 and 8.61-8.95 RMSE (Figure \ref{fig7}b). Demonstrations of detection results at two berry maturity stages were provided in Figures \ref{figa3}-\ref{figa4} in Appendices. 

\begin{table}[!h]
\caption{Model comparison using the test set under two berry maturity conditions.}
\center
\begin{tabular}{c c c c c c}\hline
Network&\makecell{Berry \\maturity}& \makecell{\textit{mAP}$^{a}$\\ \textit{(IoU = 0.5)}}& F1-score & \makecell{Inference speed \\per image (ms)$^{b}$}\\ \hline
Faster R-CNN& \multirow{5}{*}{Immature} &50.12 &0.52 &350\\
YOLOv3 & &82.84 &0.60 &13.8\\
YOLOv4 & &87.24 &0.65 &14.2\\
YOLOv5 & &89.78 &0.80  &\textbf{11.8}\\ 
\textbf{Swin-T-YOLOv5}& & \textbf{90.31} &\textbf{0.82} &13.2\\ \hline
Faster R-CNN& \multirow{5}{*}{Mature} &52.35 &0.59 &350\\
YOLOv3 & &85.43 &0.76 &13.8\\
YOLOv4 & &89.40 &0.77 &14.2\\
YOLOv5 & &91.58 &0.81  &\textbf{11.8}\\ 
\textbf{Swin-T-YOLOv5}& & \textbf{95.86} &\textbf{0.87} &13.2\\ \hline
\multicolumn{5}{l}{$^{a}$mAP refers to mean Average Precision.}\\
\multicolumn{5}{l}{$^{b}$Inference time may vary depending on the hardware configurations.}\\
\end{tabular}

\label{t6}

\end{table}
\pagebreak
\subsubsection{Testing under three sunlight directions and intensities}
Finally, all models were tested under three different sunlight directions and intensities, including in the morning (8am-9am; in the direction of the light), noon (11am-12pm; maximum solar elevation angle), and afternoon (4pm-5pm; against the direction of the light) (Table \ref{t1}). The light intensity was the highest at noon and was the lowest in the morning. Specific comparison results were given in Table \ref{t7}. Among all models tested in this research, Swin-T-YOLOv5 performed the best under any sunlight condition, with the optimal mAPs of 91.96$\%$-94.53$\%$ and F1-scores of 0.83-0.86. It was also obvious that the detection results were better at noon than in the morning or afternoon with 2.49$\%$-2.57$\%$ higher mAP and 0.01-0.03 higher F1-score. Additionally, YOLOv5 still performed the second best except for during noon, where Swin-T-YOLOv5 and YOLOv4 achieved 6.08$\%$ and 1.71$\%$ better mAP than it. 

\begin{figure}[!h]
  \centering
  
  \includegraphics[width = 14.8cm,height = 15cm]{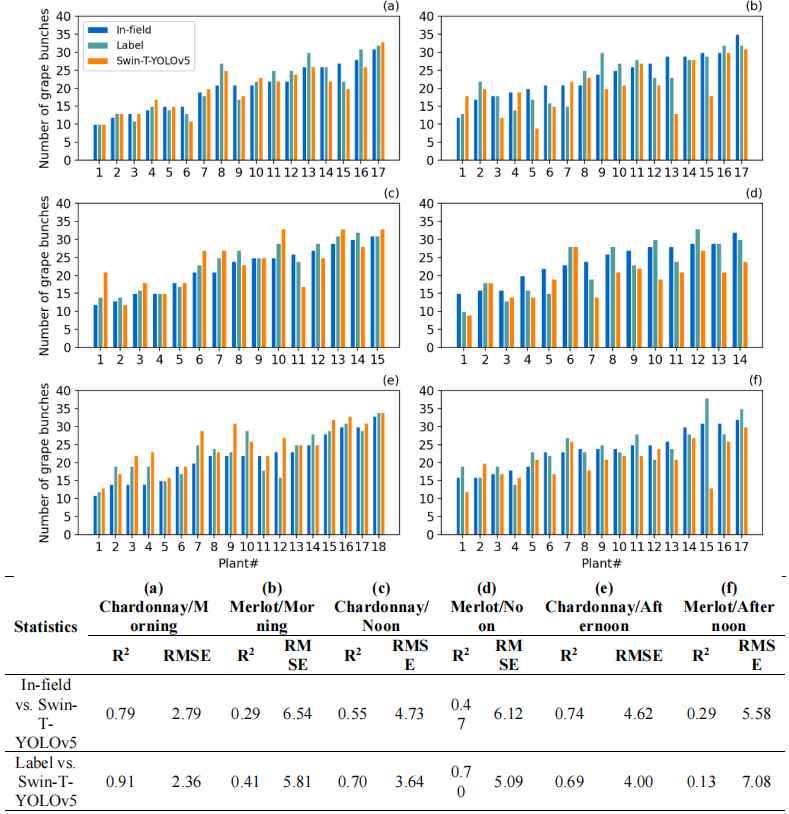} 
  
  \caption{The number of grape bunches comparison between in-field manual counting (gTruth), manual label, and detection using Swin-transformer-YOLOv5 (Swin-T-YOLOv5) during (a-b) morning, (c-d) noon, (e-f) afternoon with Chardonnay (left) and Merlot (right). RMSE refers to root mean square error.}
  \label{fig8}
 
\end{figure}

Further observations on the number of grape bunches detected by Swin-T-YOLOv5 comparing against ground truth data, from in-field manual counting and manual labeling, were provided in Figure \ref{fig8}. For Chardonnay variety, the agreement between the predictions and ground truth were relatively better (0.55-0.91 of R2 and 2.36-4.73 of RMSE; Figure \ref{fig8}a, \ref{fig8}c, and \ref{fig8}e) than Merlot variety (0.13-0.70 of R2 and 5.09-7.08 of RMSE; Figure \ref{fig8}b, \ref{fig8}d, and \ref{fig8}f) under any sunlight conditions. The results for Merlot were the best at noon with 0.47-0.70 of R2 (Figure \ref{fig8}d), while the results were the worst in the afternoon with only 0.13-0.29 of R2 when the imaging side was against the direction of the sunlight (Figure \ref{fig8}f). Visual comparisons of model performances under different sunlight conditions can be found in Figures \ref{figa5}-\ref{figa6} in Appendices. 

\begin{figure}[!h]
  \centering
  \subfigure[]{
  \includegraphics[width = 5cm,height = 5cm]{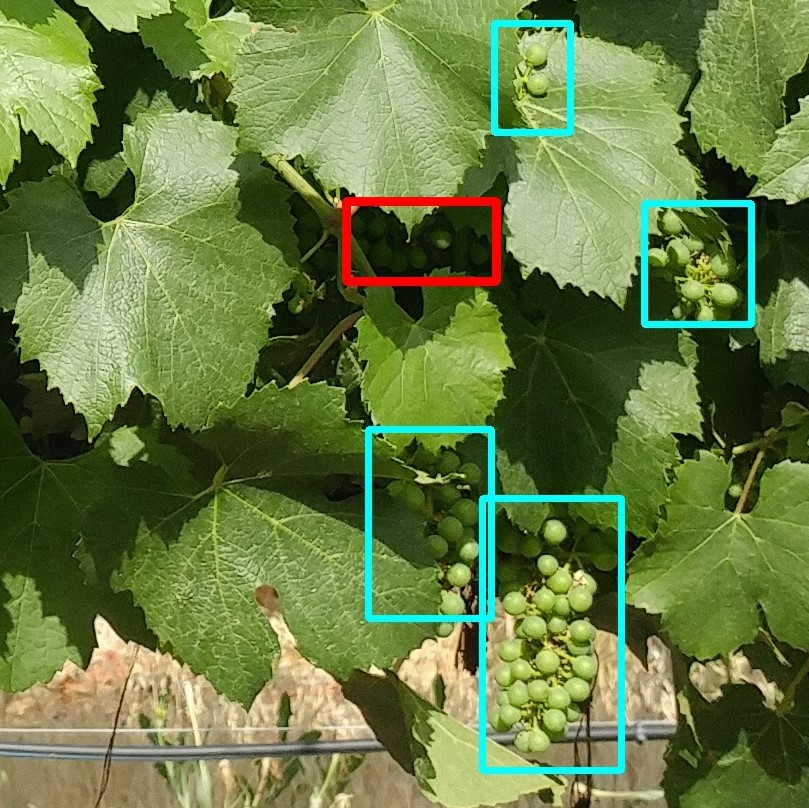} \label{fig9a}
  }
  \subfigure[]{
  \includegraphics[width = 5cm,height = 5cm]{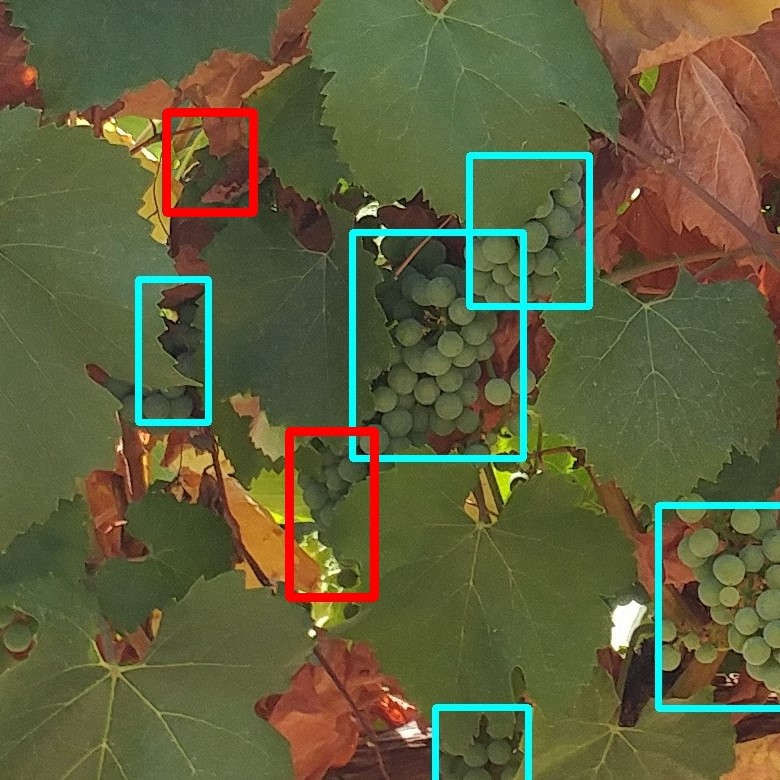} \label{fig9b}
  }
  \subfigure[]{
  \includegraphics[width = 5cm,height = 5cm]{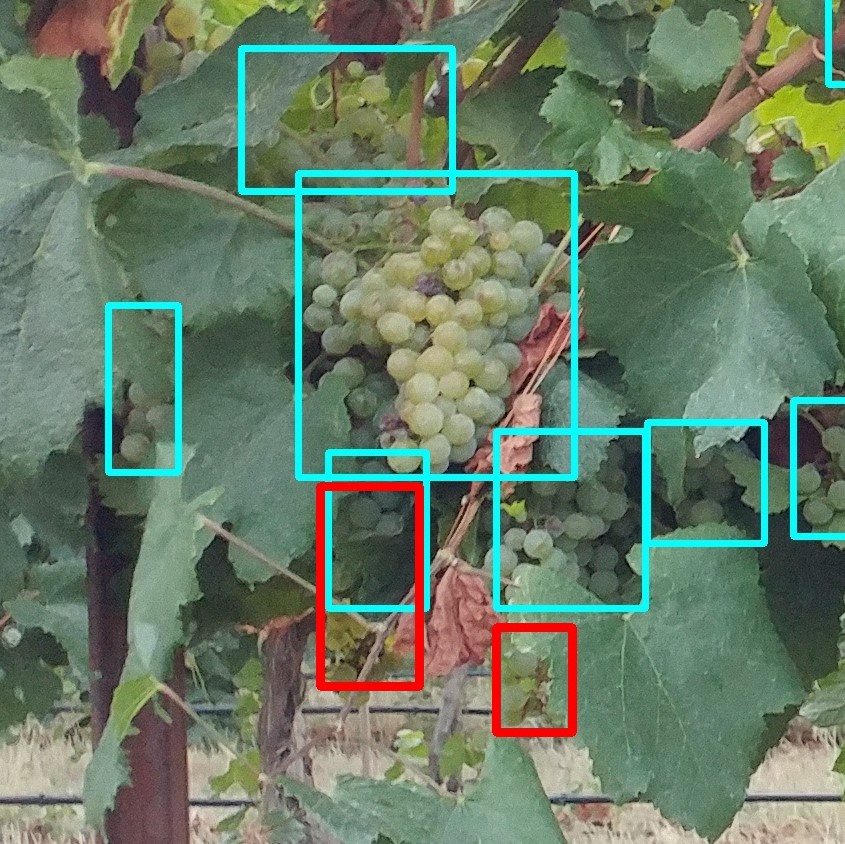} \label{fig9c}
  }
  \subfigure[]{
  \includegraphics[width = 5cm,height = 5cm]{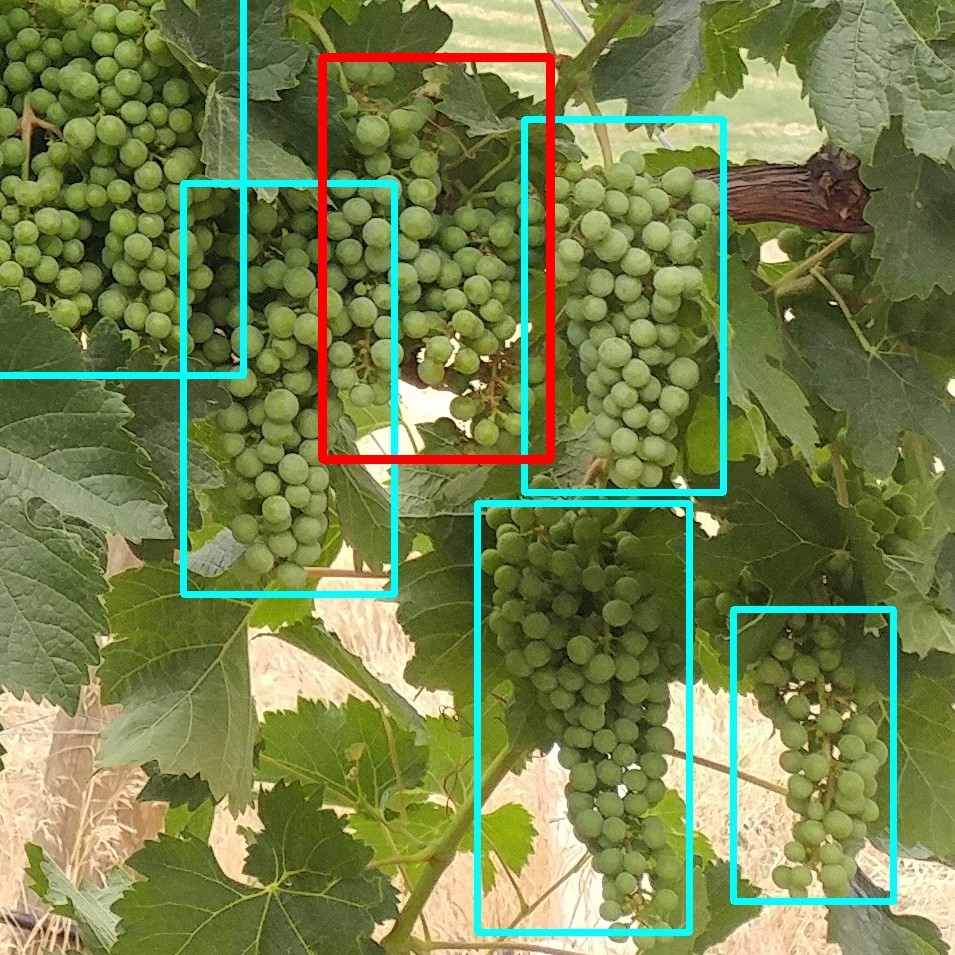} \label{fig9d}
  }
  \subfigure[]{
  \includegraphics[width = 5cm,height = 5cm]{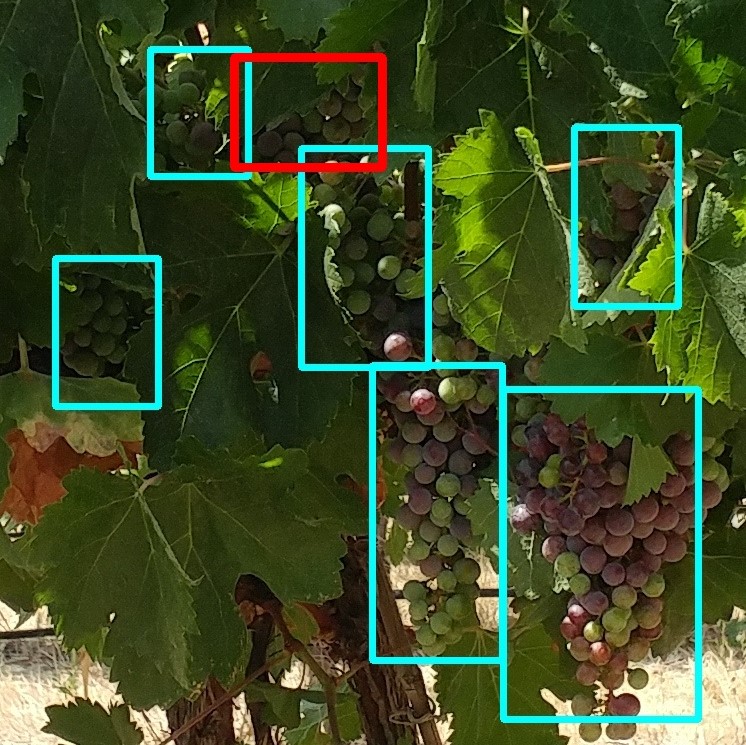} \label{fig9e}
  }
  \subfigure[]{
  \includegraphics[width = 5cm,height = 5cm]{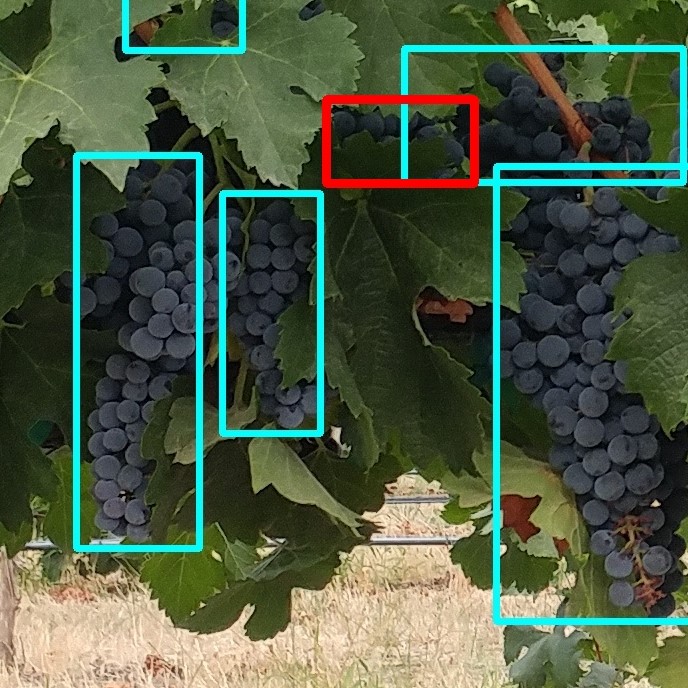} \label{fig9f}
  }
  \caption{Illustrations of failures (i.e., True Negatives (TN) or False Positives (FP) highlighted in red 
bounding boxes) in grape bunch detection using Swin-transformer-YOLOv5 (Swin-T-YOLOv5) during 
early- (left column), mid- (middle column), and harvest-stage (right column) for (a-c) “Chardonnay (white
variety)” and (d-f) “Merlot (red variety)” in zoomed-in views.}
  \label{fig9}
\end{figure}

Although Swin-T-YOLOv5 outperformed all other models in detecting grape bunches under various external or internal variations, detection failures (i.e., TNs and FPs) happened more frequently in several scenarios as illustrated in Figure 9. For example, severe occlusion (mainly by leaves) caused detection failure was the major reason for having TNs and FPs in this research as marked out using the red bounding boxes, particularly when the visible part of grape bunches were small (Figure \ref{fig9e}-\ref{fig9f}) or having the similar color compared to the background (Figure \ref{fig9a}-\ref{fig9c}). In addition, clustered grape bunches can cause detection failures, where two grape bunches were detected as one (Figure \ref{fig9d}). 

\begin{table}[!h]
\caption{Model comparison using the test set under three sunlight directions and intensities.}
\center
\begin{tabular}{c c c c c c}\hline
Network&\makecell{Sunlight \\direction}& \makecell{\textit{mAP}$^{a}$\\ \textit{(IoU = 0.5)}}& F1-score & \makecell{Inference speed \\per image (ms)$^{b}$}\\ \hline
Faster R-CNN& \multirow{5}{*}{Morning} &55.35 &0.56 &350\\
YOLOv3 & &74.57 &0.65 &13.8\\
YOLOv4 & &78.15 &0.67 &14.2\\
YOLOv5 & &89.57 &0.79  &\textbf{11.8}\\ 
\textbf{Swin-T-YOLOv5}& & \textbf{92.04} &\textbf{0.83} &13.2\\ \hline
Faster R-CNN& \multirow{5}{*}{Noon} &60.73 &0.59 &350\\
YOLOv3 & &86.31 &0.67 &13.8\\
YOLOv4 & &90.16 &0.70 &14.2\\
YOLOv5 & &88.45 &0.79  &\textbf{11.8}\\ 
\textbf{Swin-T-YOLOv5}& & \textbf{94.53} &\textbf{0.86} &13.2\\ \hline
Faster R-CNN& \multirow{5}{*}{Afternoon} &56.79 &0.52 &350\\
YOLOv3 & &76.78 &0.70 &13.8\\
YOLOv4 & &81.12 &0.73 &14.2\\
YOLOv5 & &87.46 &0.80  &\textbf{11.8}\\ 
\textbf{Swin-T-YOLOv5}& & \textbf{91.96} &\textbf{0.85} &13.2\\ \hline

\multicolumn{5}{l}{$^{a}$mAP refers to mean Average Precision.}\\
\multicolumn{5}{l}{$^{b}$Inference time may vary depending on the hardware configurations.}\\
\end{tabular}

\label{t7}

\end{table}
\pagebreak

\section{Discussion}
Compared to other mid-to-large sized fruits, such as apple, citrus, pear, and kiwifruit, grape bunches in the vineyards have more complex structural shapes and silhouette characteristics to be accurately detected using machine vision systems. The previous studies on identifying and counting grape bunches commonly employed CNNs-only object detectors, which the detection accuracies and model robustness suffered from severe canopy occlusions and varying light conditions \cite{cecotti2020grape}, \cite{li2021real}). Accurate and fast identification of overlapped grape bunches in dense-foliage canopies under natural lighting environment remains to be a key challenge in vineyards. Therefore, this research proposed the combination of architectures from a conventional CNN model (YOLOv5) and a vision Transformer model (Swin-transformer) that can inherit the advantages of both models to preserve global and local features when detecting grape bunches. By replacing several CONV and CSP Bottleneck blocks with Swin-transformer encoder blocks in the original YOLOv5 (Figure \ref{figa4}), the newly integrated detector (i.e., Swin-T-YOLOv5) worked as expected in overcoming the drawbacks of CNNs in capturing the global features due to their limited receptive fields (Figure \ref{fig5}). 

Our proposed Swin-T-YOLOv5 was tested on two different wine grape varieties (Chardonnay in white berry skin and Merlot in red berry skin), two different weather/sky conditions (sunny and cloudy), two different berry maturity stages (immature and mature), and three different sunlight directions/intensities (morning, noon, and afternoon) for its detection performance (Table \ref{t1}). A comprehensive evaluation was made by comparing Swin-T-YOLOv5 against various commonly used detectors, including Faster R-CNN, YOLOv3, YOLOv4, and YOLOv5 (Table \ref{t4}). Results verified that our proposed Swin-T-YOLOv5 outperformed all other models under any listed environmental conditions with achieved 90.31$\%$-97.19$\%$ mAPs. The best and worst results were obtained under cloudy weather and berry immature conditions, respectively, with 6.88$\%$ in difference. 

Specifically, Swin-T-YOLOv5 performed the best under cloudy weather condition with the highest mAP of 97.19$\%$, which was 1.83$\%$ higher than sunny weather condition (Table \ref{t5}), although the difference was inconsiderable. While testing the models at different berry maturity stages, Swin-T-YOLOv5 performed much better when the berries were matured with a 95.86$\%$ of mAP than immature berries (with 5.55$\%$ lower mAP; Table \ref{t6}). The results were reasonable because the berries tended to be smaller in size and lighter in color during early growth stage and thus more difficult to be detected. Moreover, Swin-T-YOLOv5 achieved better mAP at noon (94.53$\%$; with the maximum solar elevation angle) than other two timings in a day (Table \ref{t7}), while the afternoon sunlight condition (i.e., against the direction of the light) more negatively affected the model with a lower mAP of 91.96$\%$ than in the morning (i.e., in the direction of the light). Apparently, the effectiveness of the berry maturities and light directions can be the major reasons for impacting the performances of the models, while weather conditions almost did not change the detection results. The improvements from original YOLOv5 to the proposed Swin-T-YOLOv5 varied based on the conditions (0.53$\%$-6.08$\%$), however, the maximum increment happened when comparing them at noon (Table \ref{t7}). Overall, it was confirmed that the Swin-T-YOLOv5 achieved the best results in terms of mAP (up to 97.19$\%$) and F1-score (up to 0.89) among all compared models in this research for wine grape bunch detections in vineyards. Its inference speed was the second best (13.2 ms per image) only after YOLOv5’s (11.8 ms per image) under any test conditions.

To further assess the model performance, we compared the predicted number of grape bunches by Swin-T-YOLOv5 with two sets of ground truth values from both in-field manual counting and manual labeling on the images. The R2 and RMSE between Swin-T-YOLOv5 and in-field counting had the similar trends of the ones between Swin-T-YOLOv5 and manual labeling in general, potentially because of some of those heavily occluded grape bunches were not taken into consideration for labeling during the annotation process. However, the values changed vastly for the two different grape varieties (Chardonnay and Merlot) under various conditions (Figures \ref{fig6}-\ref{fig8}). First, it was clear that Swin-T-YOLOv5 did not perform well on Merlot variety when the weather condition was sunny with 0.28-0.36 of R2 and 6.97-7.04 of RMSE, which were considerably worse than cloudy weather condition with 0.66-0.69 of R2 and 3.13-3.54 of RMSE. While for Chardonnay, the model performed well under either condition with 0.70-0.82 of R2 and 2.93-5.05 of RMSE (Figure \ref{fig6}). Similarly, the performance of Swin-T-YOLOv5 was poor on Merlot variety when the berries were immature with 0.08-0.16 of R2 and 8.61-8.95 of RMSE, which were again much worse than mature condition with 0.57-0.70 of R2 and 3.30-3.86 of RMSE. For Chardonnay, Swin-T-YOLOv5 achieved better results under either condition with 0.59-0.89 of R2 and 2.50-4.44 of RMSE (Figure \ref{fig7}). In addition, Swin-T-YOLOv5 underperformed on Merlot variety when the sunlight condition was in the afternoon with 0.13-0.29 of R2 and 5.58-7.08 of RMSE, which were slightly worse than in the morning with 0.29-0.41 of R2 and 5.81-6.54 of RMSE. Comparatively, it worked better on Merlot at noon with 0.47-0.70 of R2 and 5.09-6.12 of RMSE. For Chardonnay, the predictions were more accurate comparing with ground truth values with 0.55-0.91 of R2 and 2.36-4.73 of RMSE (Figure \ref{fig8}). This was possibly because Merlot variety had a more complex combination of grape bunches when the berries were immature with either white or white-red mixed color (Figure \ref{fig.1e}), which may cause more detection errors under more challenging test conditions, such as when imaging against the direction of the light. In general, detecting grape bunches of Merlot variety was more challenging than Chardonnay variety under any test conditions in this research.

\section{Conclusion}
This This research attempted to propose an optimal and real-time wine grape bunch detection model in natural vineyards by architecturally integrating YOLOv5 and Swin-transformer detectors, called Swin-T-YOLOv5. The research was carried out on two different grape varieties, Chardonnay (white color of berry skin when matured) and Merlot (red color of berry skin when matured), throughout the growth season from 7/4/2019 to 9/30/2019 under various testing conditions, including two different weather/sky conditions (i.e., sunny and cloudy), two different berry maturity stages (i.e., immature and mature), and three different sunlight directions/intensities (i.e., morning, noon, and afternoon). Further assessment was made by comparing the proposed Swin-T-YOLOv5 with other commonly used detectors, including Faster R-CNN, YOLOv3, YOLOv4, and YOLOv5. Based on the obtained results, the following conclusions can be drawn:
 \begin{enumerate}
            \item The proposed Swin-T-YOLOv5 was integrated by replacing the last C3 layer (i.e., with CSP Bottleneck and three CONVs) in the original YOLOv5 with Swin-transformer encoder blocks to inherit the advantages from both models for global and local information preservation. Validation results verified the advancement of Swin-T-YOLOv5 with the best Precision of 98$\%$, Recall of 95$\%$, mAP of 97$\%$, and F1-score of 0.96; 
            \item Swin-T-YOLOv5 outperformed all other studied models under any test conditions in this research:
             \begin{enumerate}
            \item Two weather conditions: During sunny weather, Swin-T-YOLOv5 achieved 95$\%$ of mAP and 0.86 of F1-score, which were up to 11$\%$ and 0.22 higher than all other YOLO family models (i.e., YOLOv3, YOLOv4, or YOLOv5); During cloudy weather, it achieved 97$\%$ of mAP and 0.89 of F1-score, which were up to 18$\%$ and 0.17 higher than the same ones; 
            \item Two berry maturity stages: During immature berry stage, Swin-T-YOLOv5 achieved 90$\%$ of mAP and 0.82 of F1-score, which were up to 7$\%$ and 0.22 higher than all other YOLO family models; During mature berry stage, it achieved 96$\%$ of mAP and 0.87 of F1-score, which were up to 10$\%$ and 0.11 higher than the same ones; 
            \item Three sunlight directions/intensities: In the morning, Swin-T-YOLOv5 achieved 92$\%$ of mAP and 0.83 of F1-score, which were up to 17$\%$ and 0.18 higher than all other YOLO family models; At noon, it achieved 95$\%$ of mAP and 0.86 of F1-score, which were up to 8$\%$ and 0.19 higher than the same ones; In the afternoon, it achieved 92$\%$ of mAP and 0.85 of F1-score, which were up to 15$\%$ and 0.15 higher than the same ones; 
        \end{enumerate}
            \item Swin-T-YOLOv5 performed differently on Chardonnay and Merlot varieties when comparing the predictions against the ground truth data (i.e., in-field manual counting and manual labeling). For Chardonnay variety, Swin-T-YOLOv5 provided desired predictions under almost all test conditions, with up to 0.91 of R2 and 2.36 of RMSE. For Merlot variety, Swin-T-YOLOv5 performed better under several test conditions (e.g., 0.70 of R2 and 3.30 of RMSE for matured berries), while underperformed when detecting immature berries (0.08 of R2 and 8.95 of RMSE). 

        \end{enumerate}

\bibliographystyle{unsrt} 

\bibliography{references}

\newpage
\section*{Appendices}
\begin{figure}[!h]
  \centering
  \subfigure[]{
  \includegraphics[width = 7cm,height = 4.2cm]{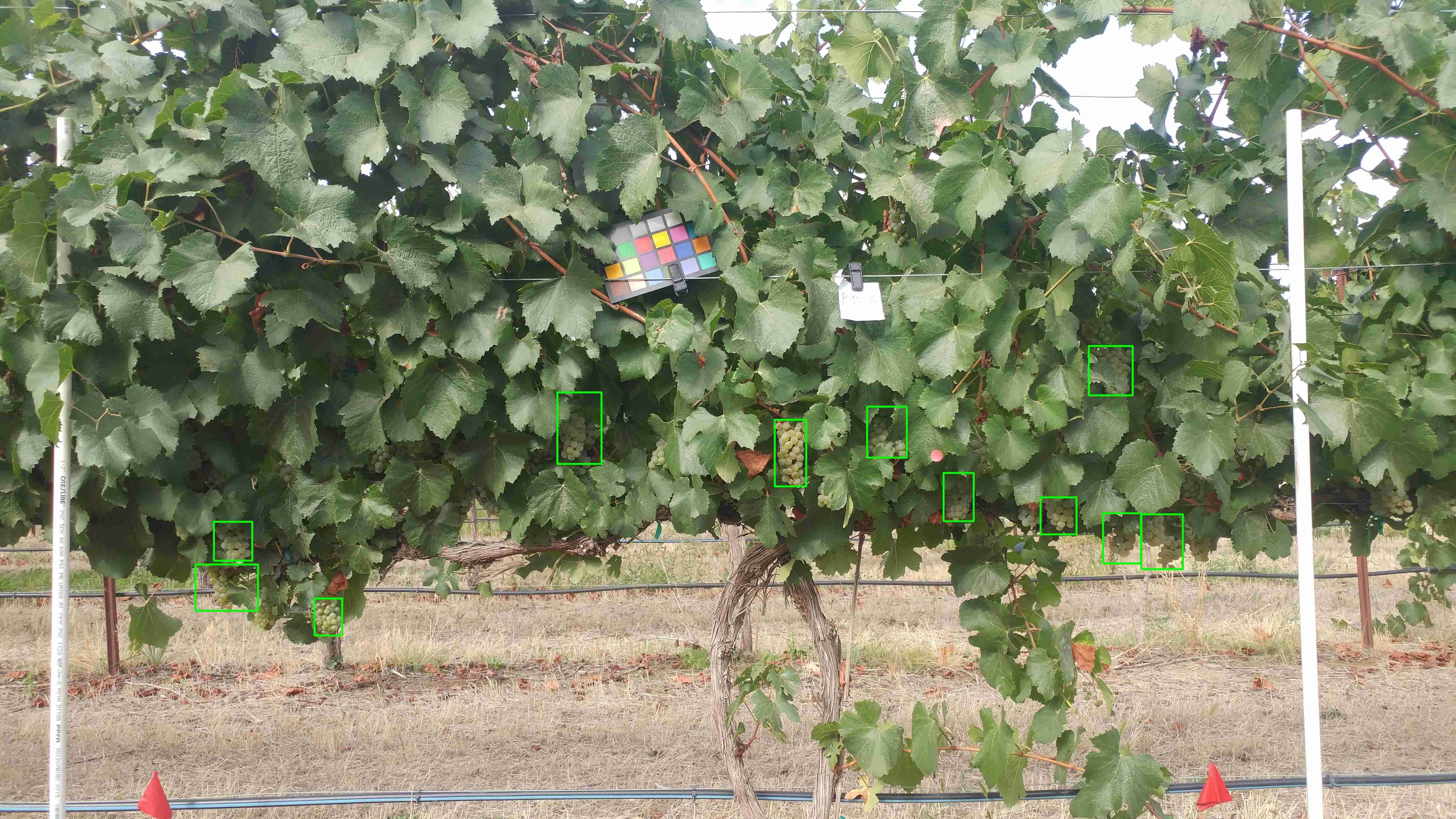} \label{fig.a1a}
  }
  \subfigure[]{
  \includegraphics[width = 7cm,height = 4.2cm]{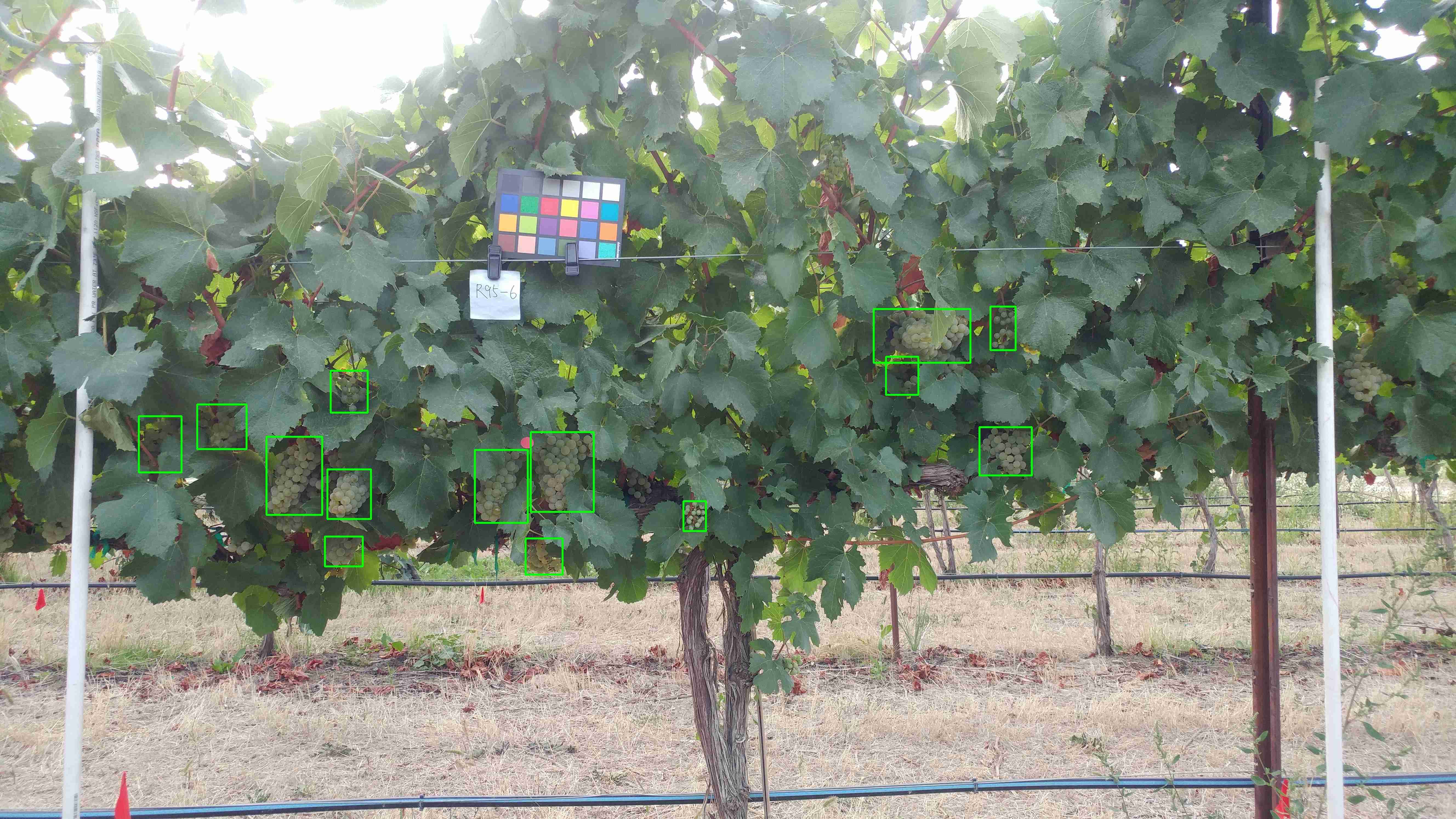} \label{fig.a1b}
  }
  \subfigure[]{
  \includegraphics[width = 7cm,height = 4.2cm]{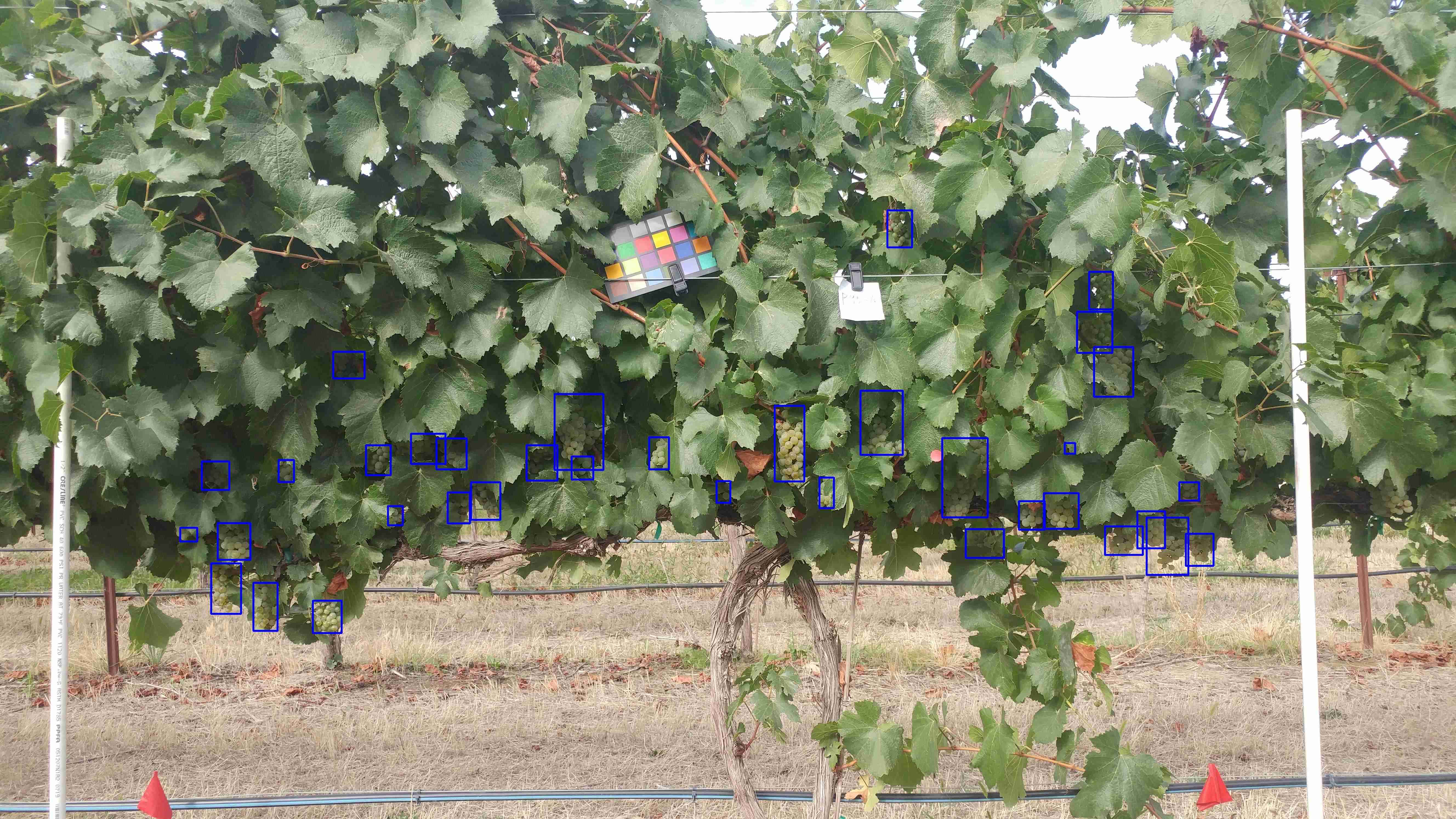} \label{fig.a1c}
  }
  \subfigure[]{
  \includegraphics[width = 7cm,height = 4.2cm]{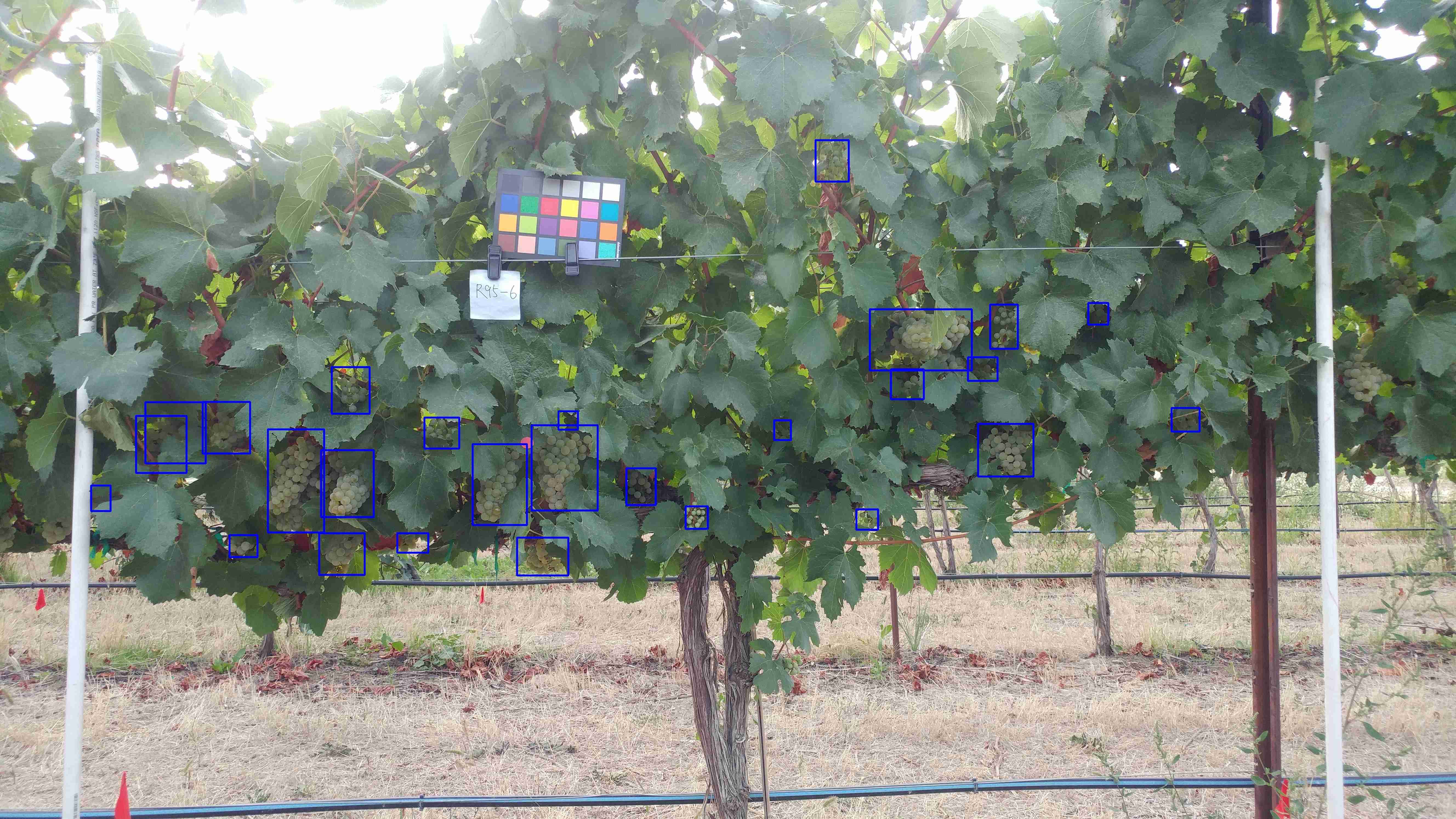} \label{fig.a1d}
  }
  \subfigure[]{
  \includegraphics[width = 7cm,height = 4.2cm]{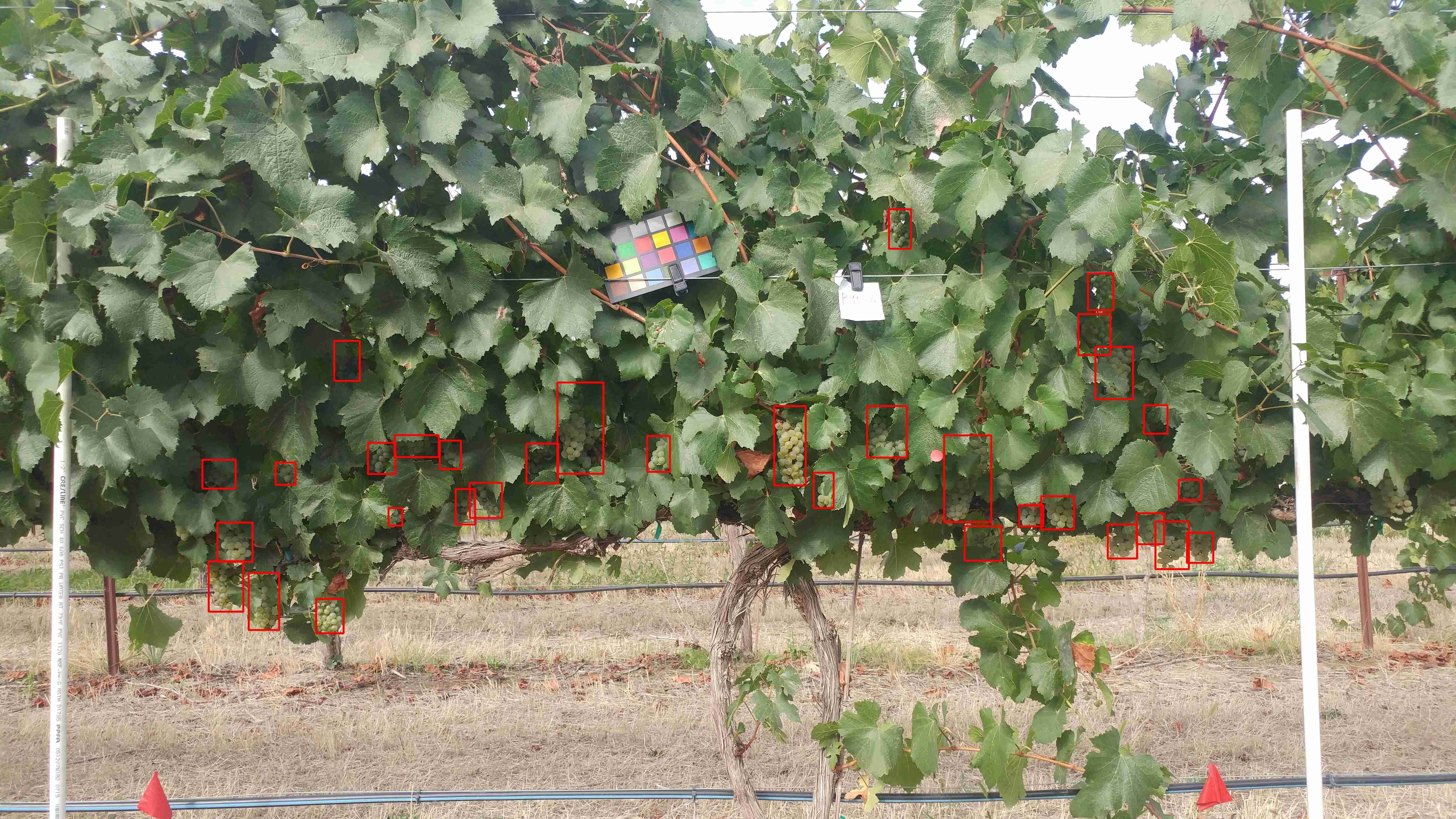} \label{fig.a1e}
  }
  \subfigure[]{
  \includegraphics[width = 7cm,height = 4.2cm]{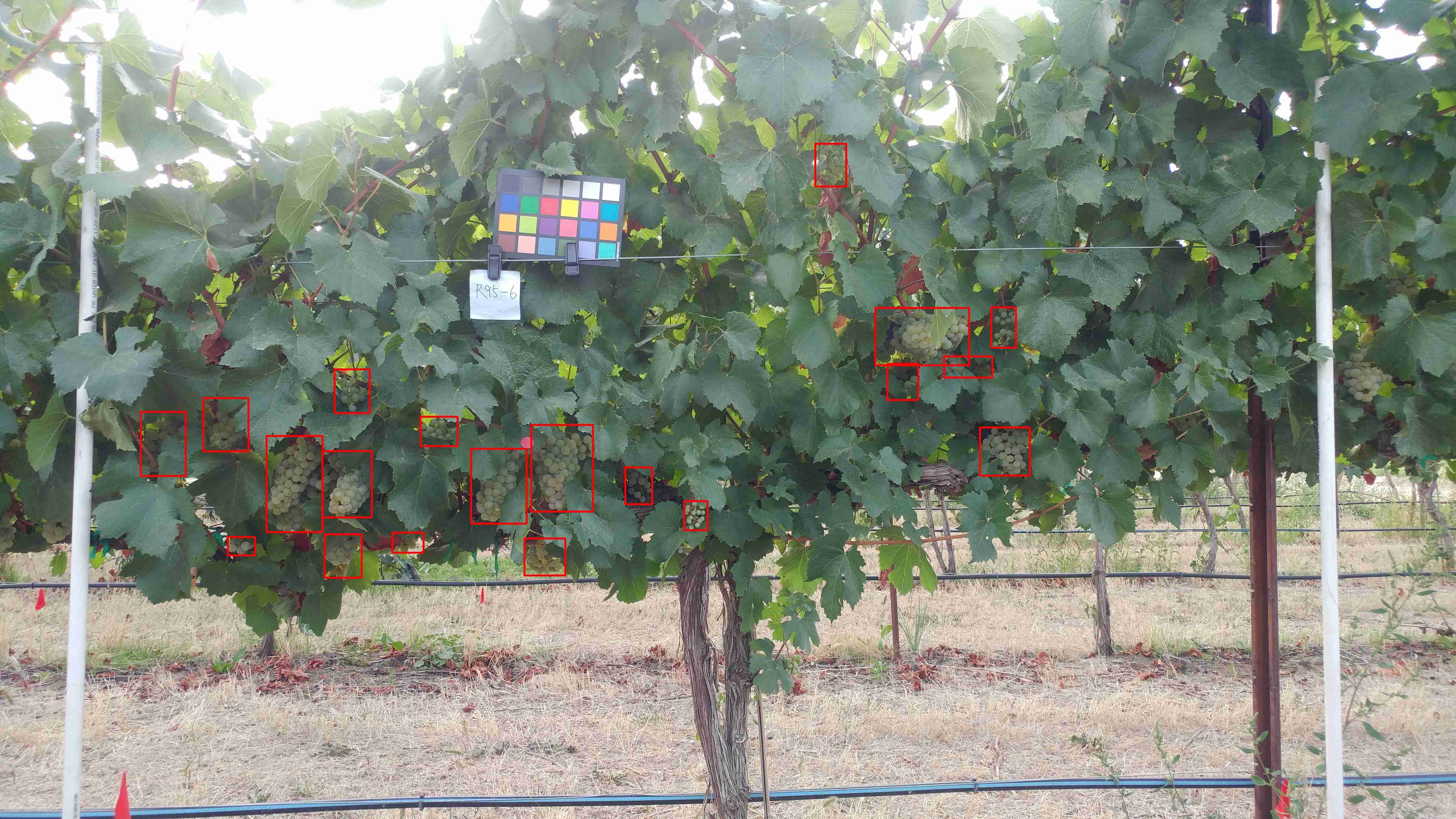} \label{fig.a1f}
  }
  \subfigure[]{
  \includegraphics[width = 7cm,height = 4.2cm]{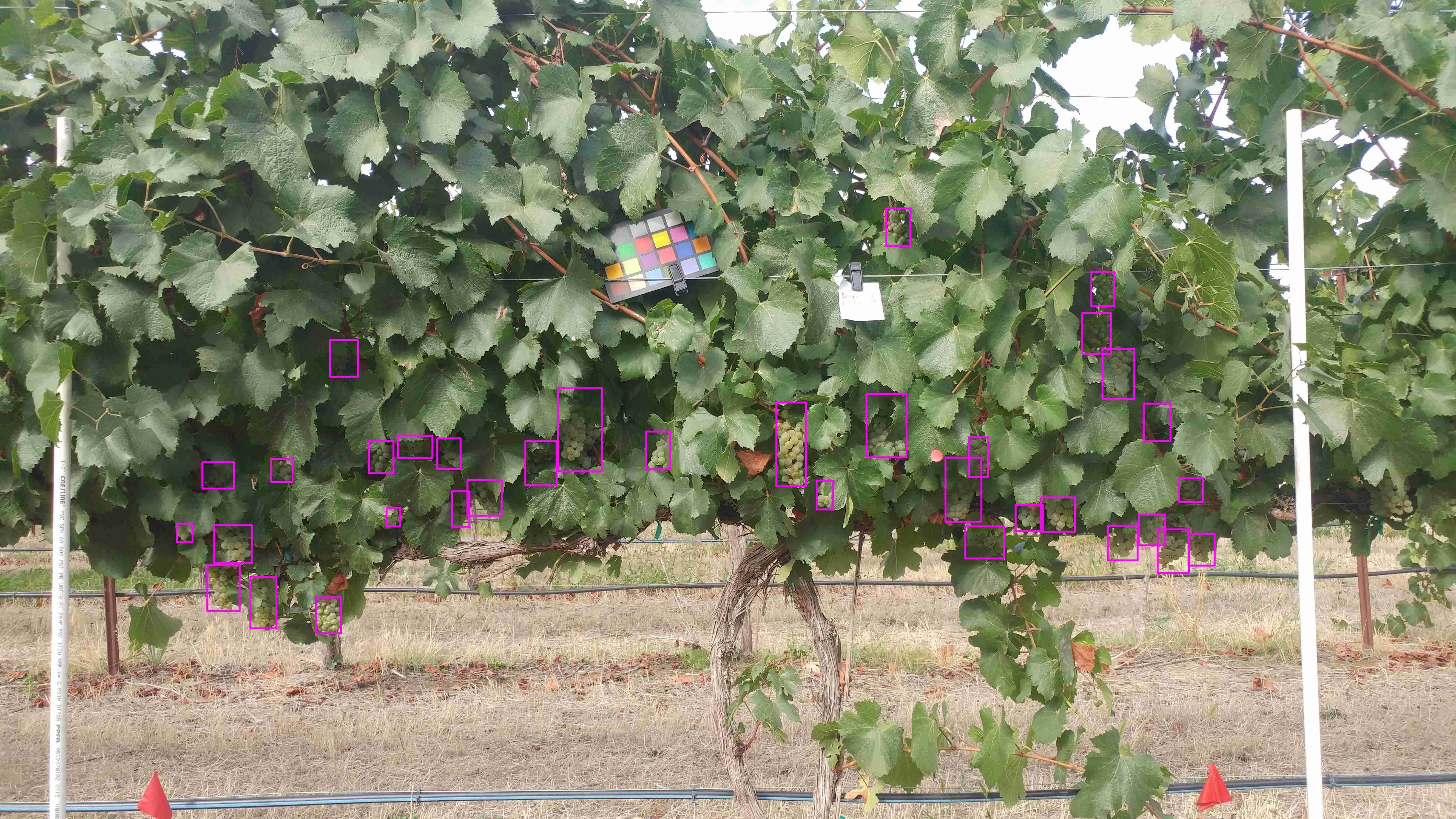} \label{fig.a1g}
  }
  \subfigure[]{
  \includegraphics[width = 7cm,height = 4.2cm]{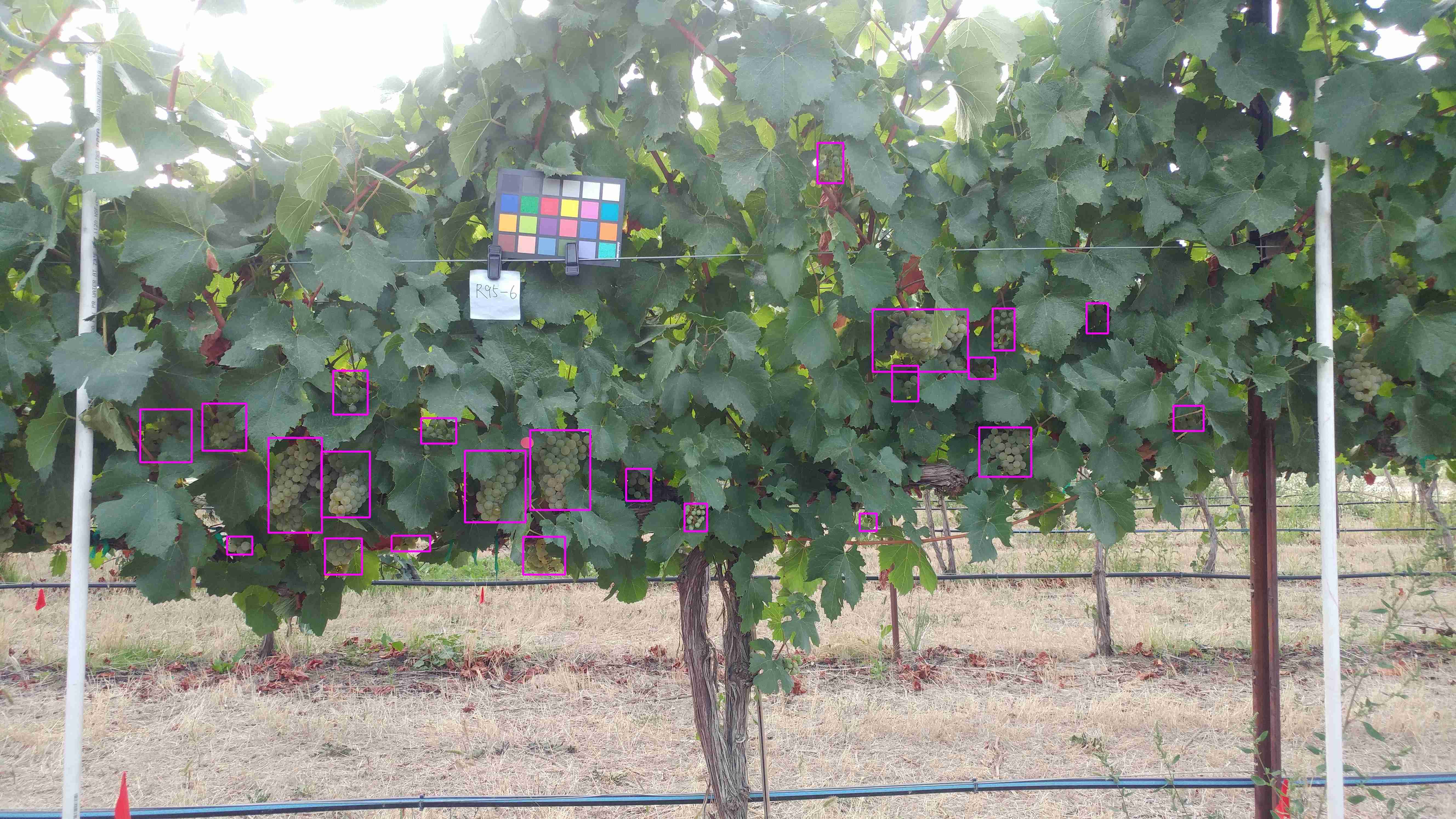} \label{fig.a1h}
  }
\end{figure}
\begin{figure}[H]
  \centering

  \subfigure[]{
  \includegraphics[width = 7cm,height = 4.2cm]{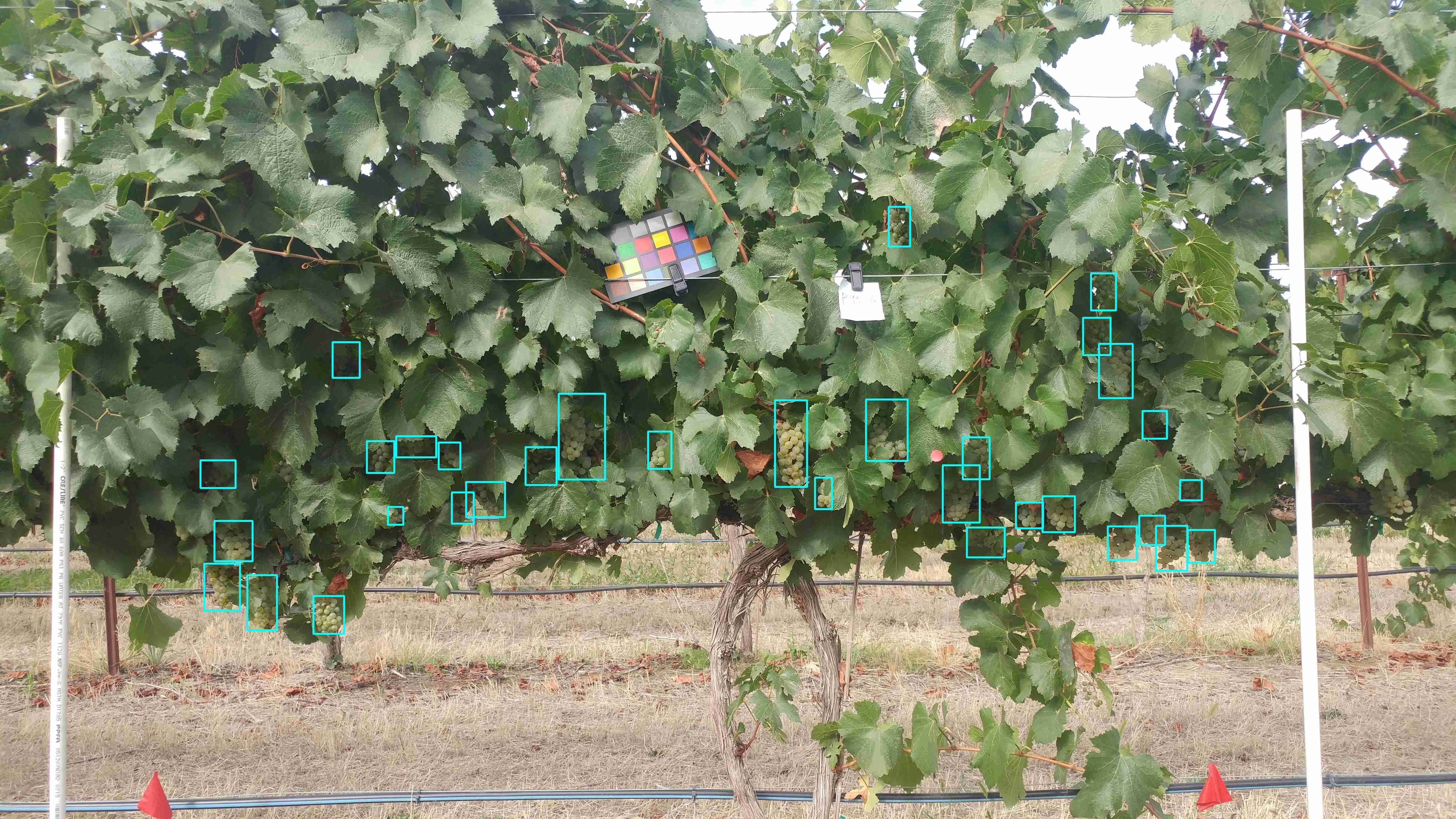} \label{fig.a1i}
  }
  \subfigure[]{
  \includegraphics[width = 7cm,height = 4.2cm]{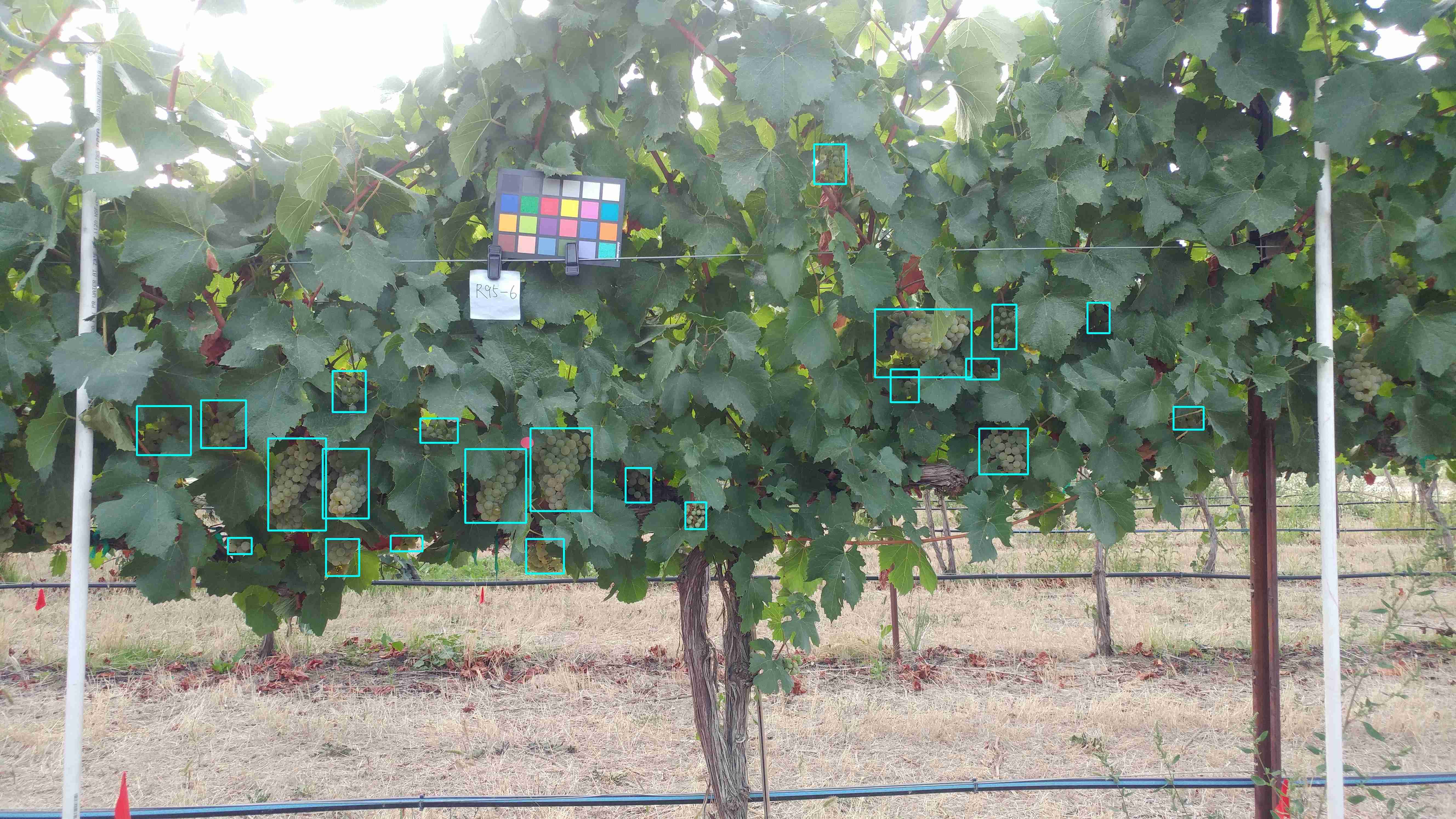} \label{fig.a1j}
  }
  \renewcommand*{\thefigure}{A.1}
  \caption{Demonstrations of detection results on the test set of Chardonnay (white variety) using (a-b) Faster R-CNN (bounding boxes in green color), (c-d) YOLOv3 (in blue color), (e-f) YOLOv4 (in red color), (g-h) YOLOv5 (in magenta color), and (i-j) Swin-transformer-YOLOv5 (in cyan color) under sunny (left) and cloudy (right) weathers.}
  \label{figa1}
\end{figure}
\newpage
\begin{figure}[!h]
  \centering
  \subfigure[]{
  \includegraphics[width = 7cm,height = 4.2cm]{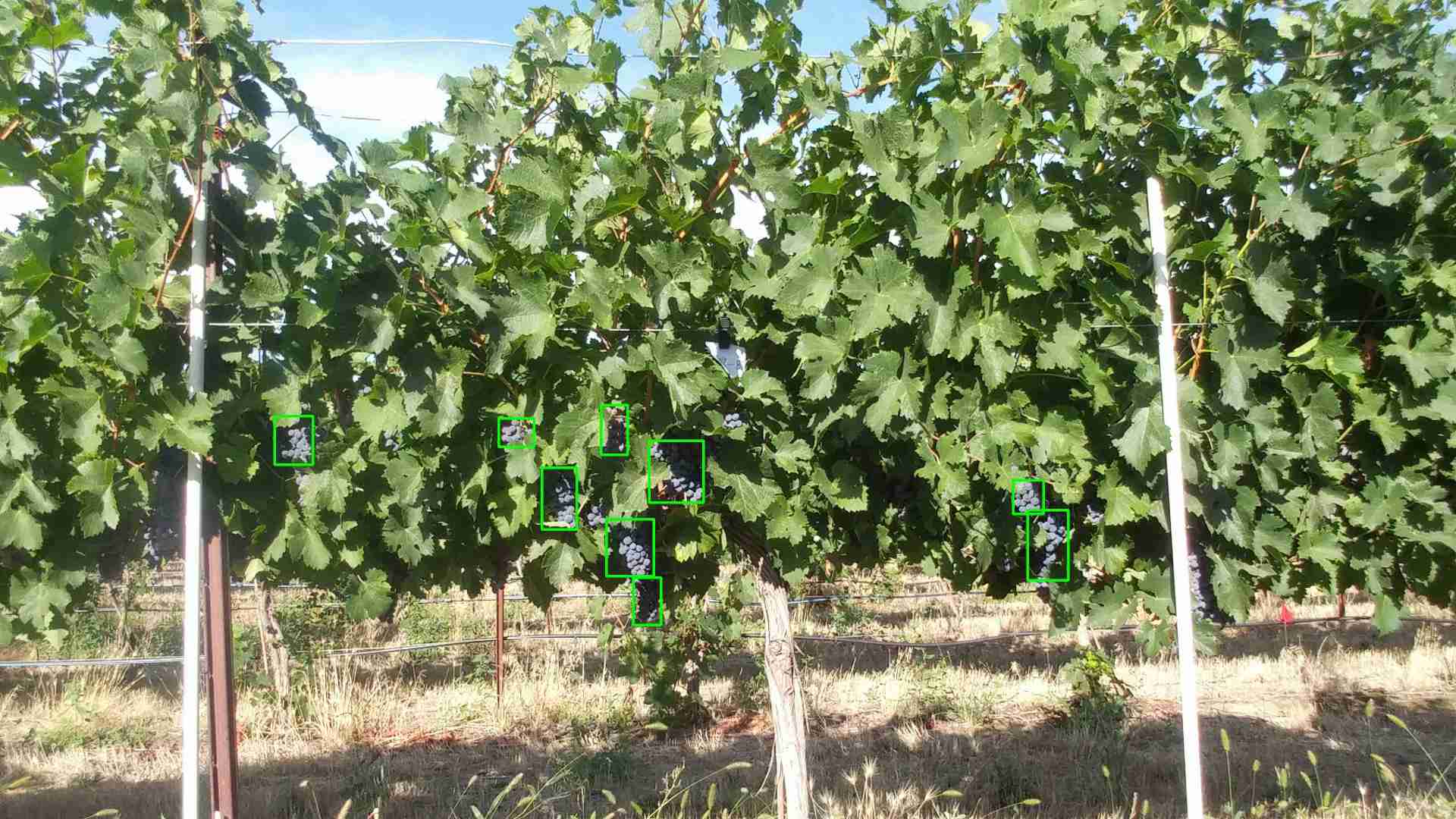} \label{fig.a2a}
  }
  \subfigure[]{
  \includegraphics[width = 7cm,height = 4.2cm]{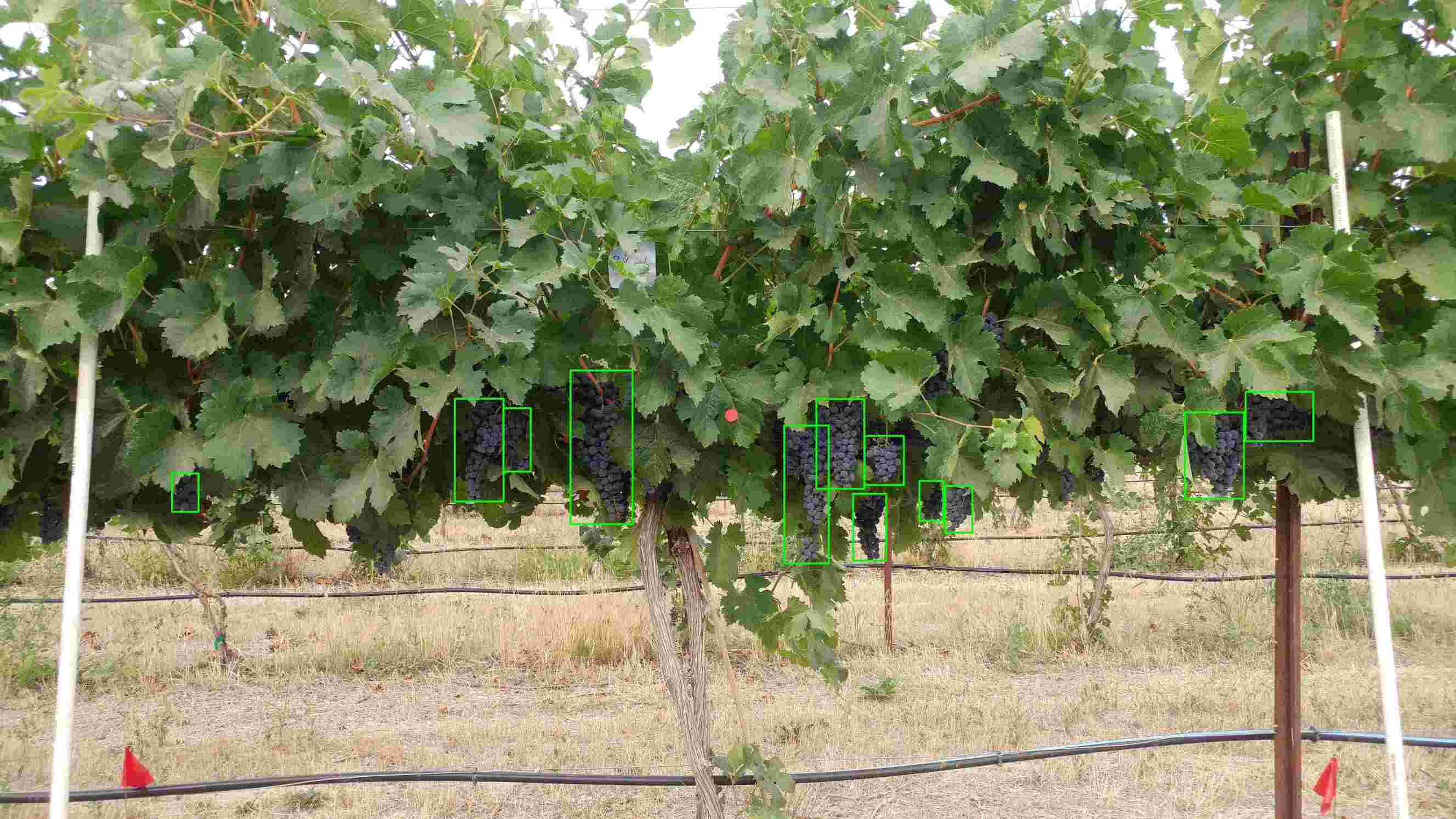} \label{fig.a2b}
  }
  \subfigure[]{
  \includegraphics[width = 7cm,height = 4.2cm]{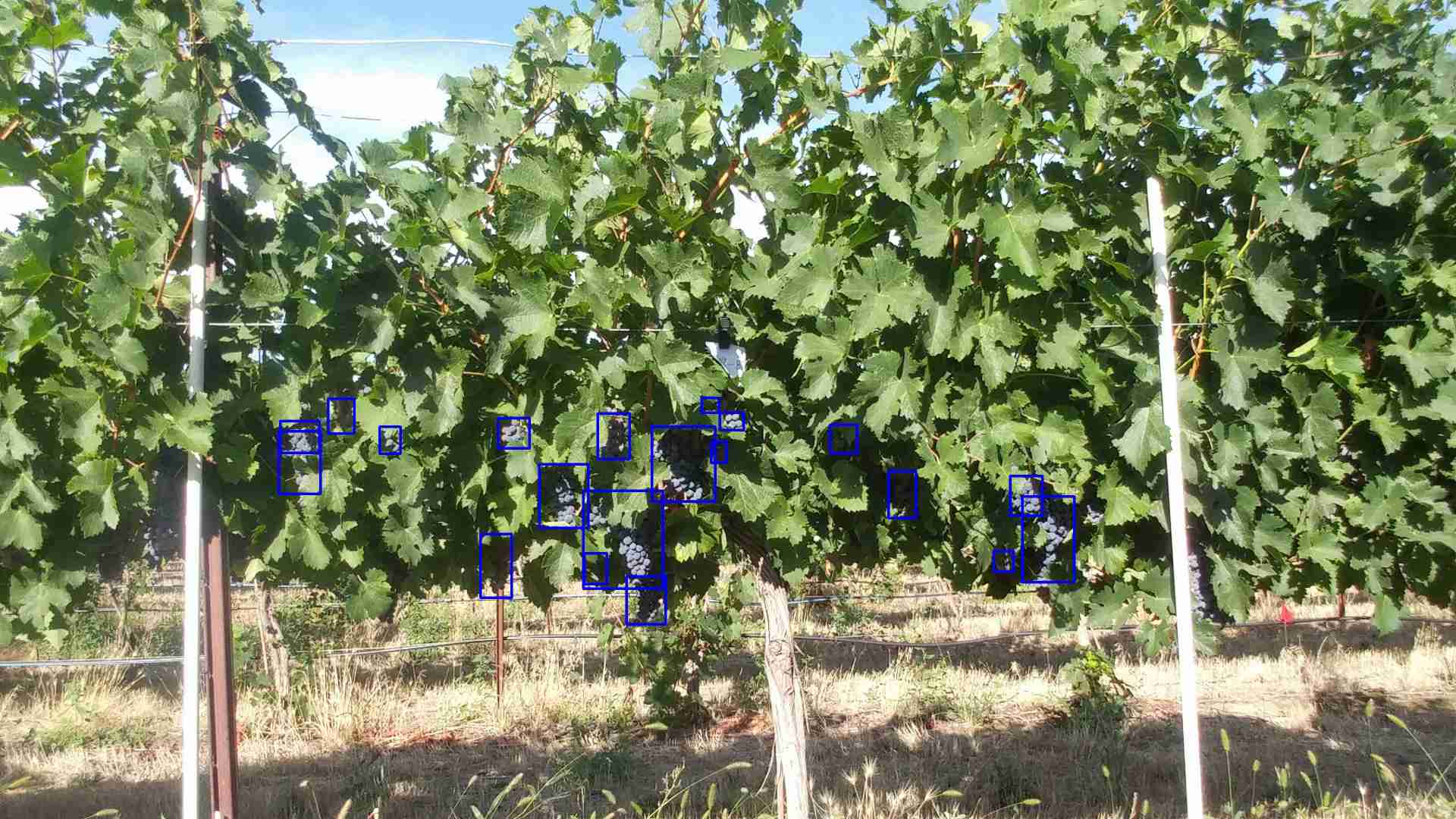} \label{fig.a2c}
  }
  \subfigure[]{
  \includegraphics[width = 7cm,height = 4.2cm]{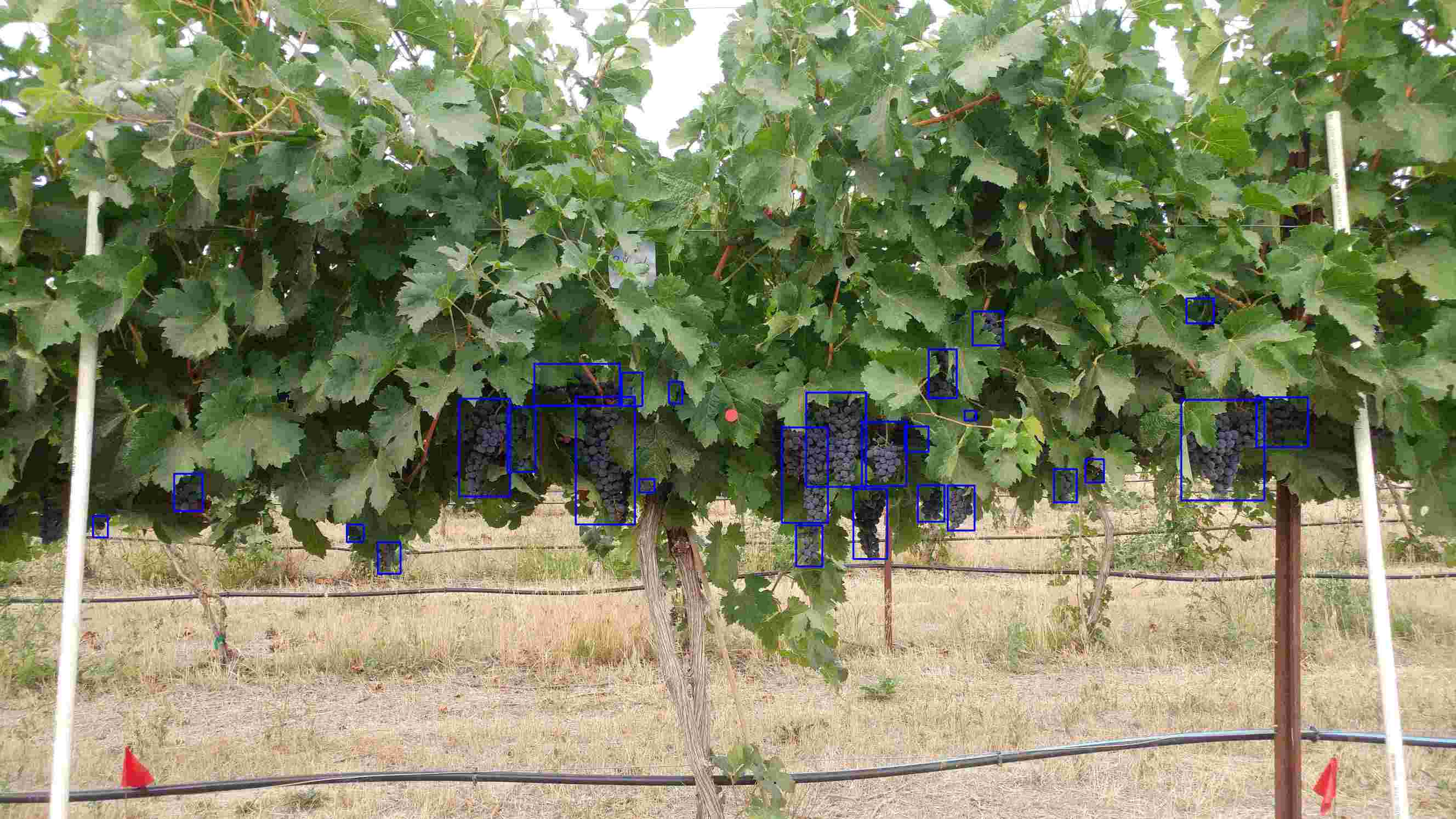} \label{fig.a2d}
  }
  \subfigure[]{
  \includegraphics[width = 7cm,height = 4.2cm]{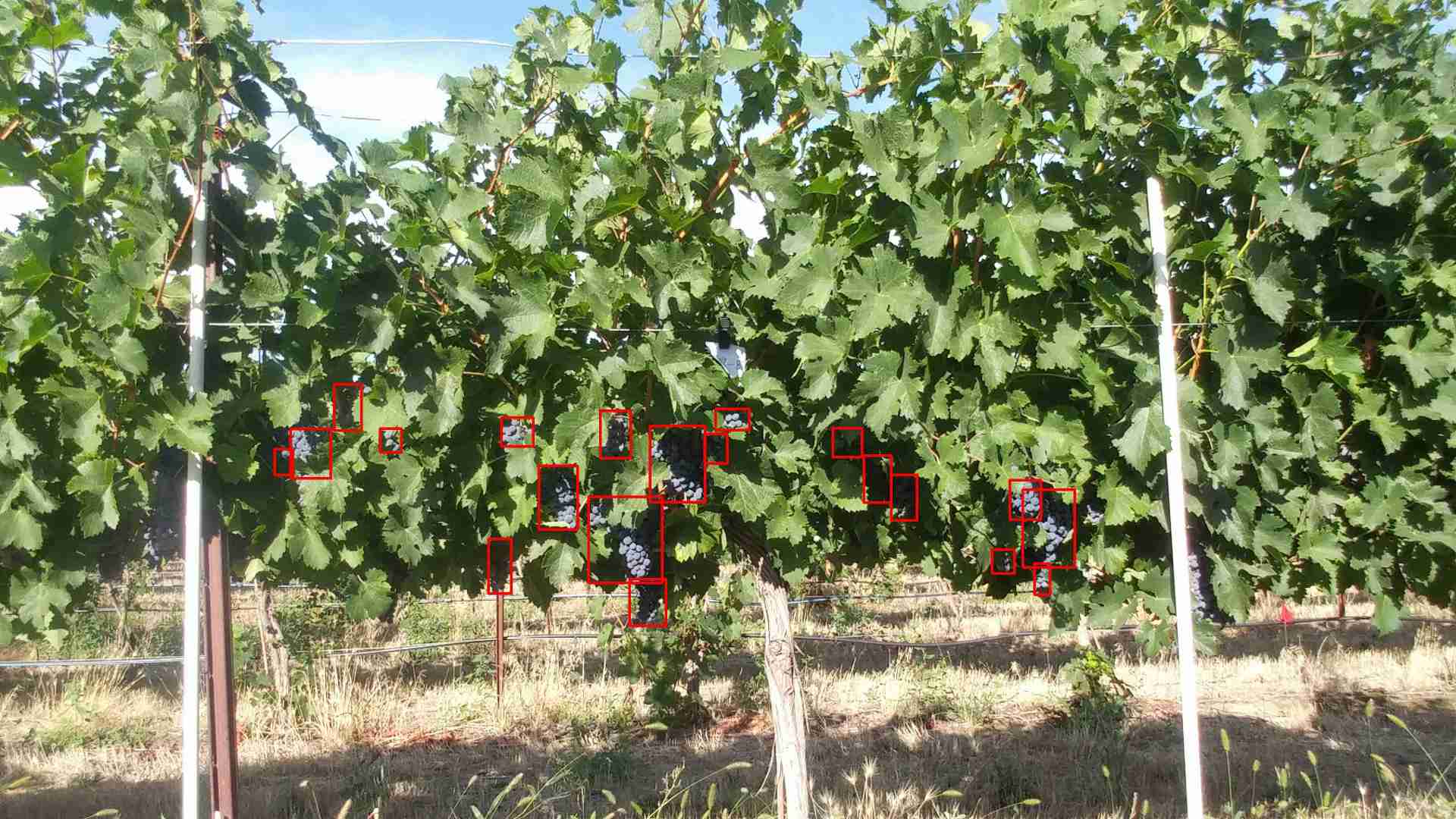} \label{fig.a2e}
  }
  \subfigure[]{
  \includegraphics[width = 7cm,height = 4.2cm]{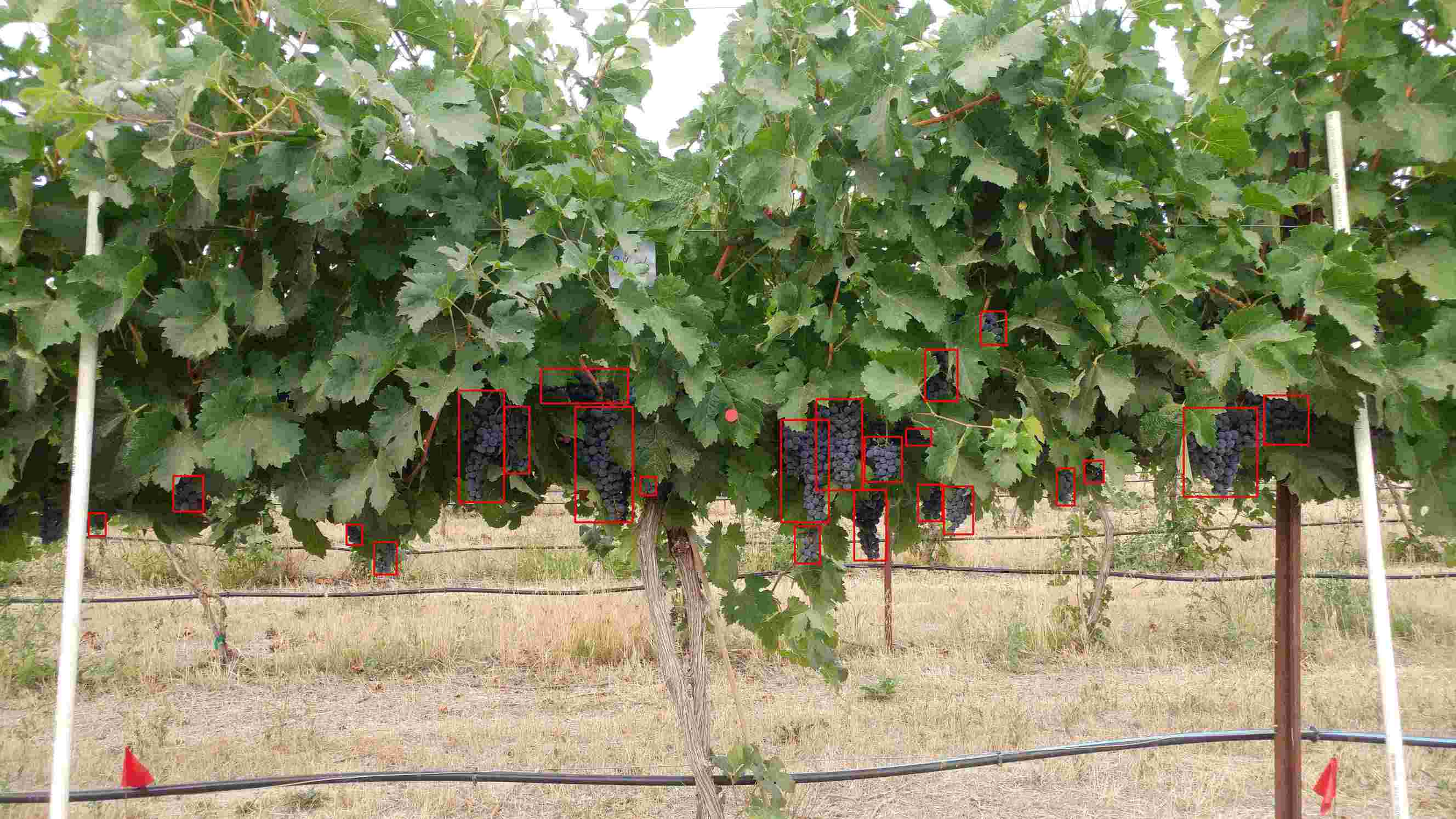} \label{fig.a2f}
  }
  \subfigure[]{
  \includegraphics[width = 7cm,height = 4.2cm]{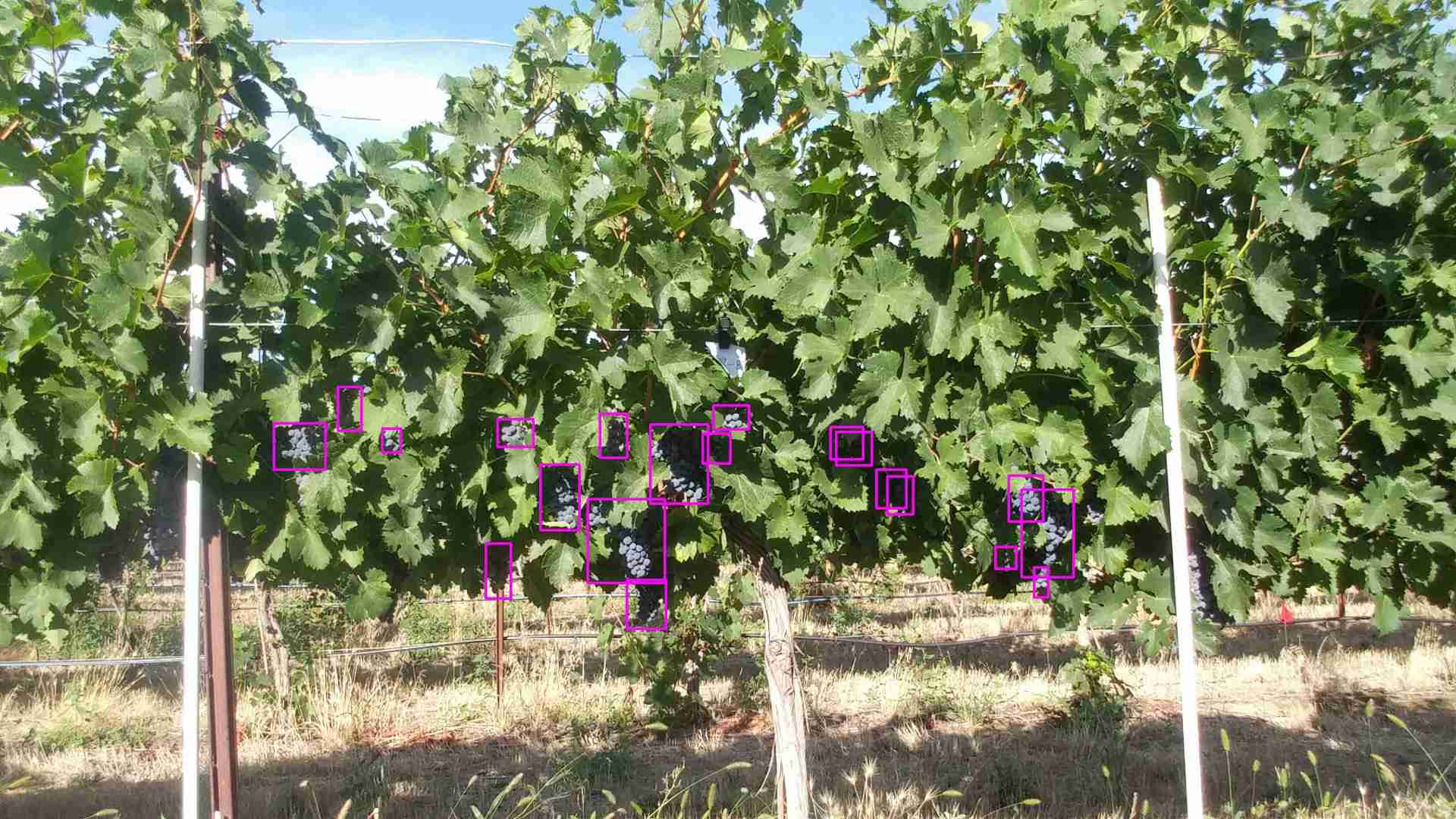} \label{fig.a2g}
  }
  \subfigure[]{
  \includegraphics[width = 7cm,height = 4.2cm]{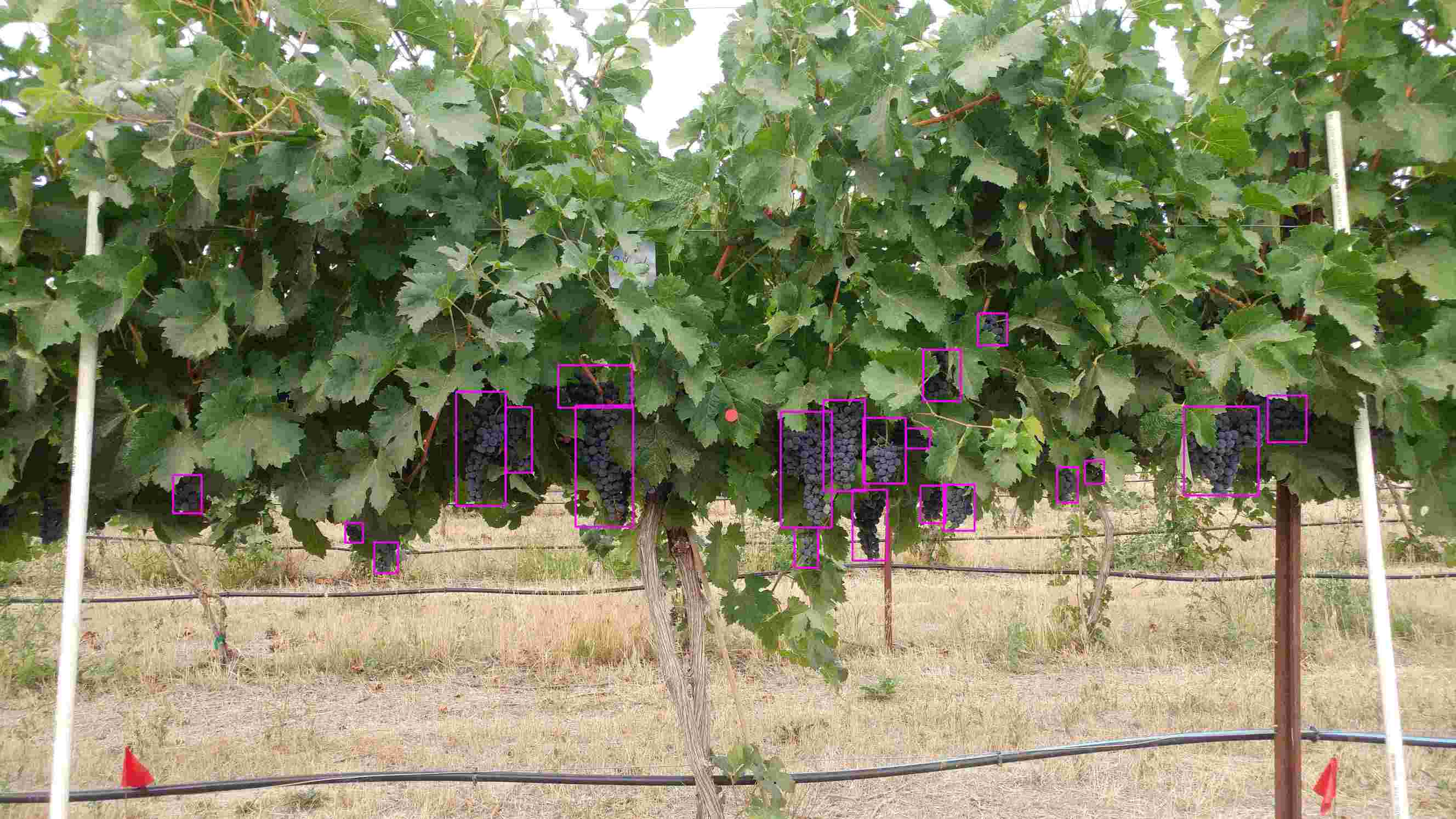} \label{fig.a2h}
  }
\end{figure}
\begin{figure}[H]
  \centering

  \subfigure[]{
  \includegraphics[width = 7cm,height = 4.2cm]{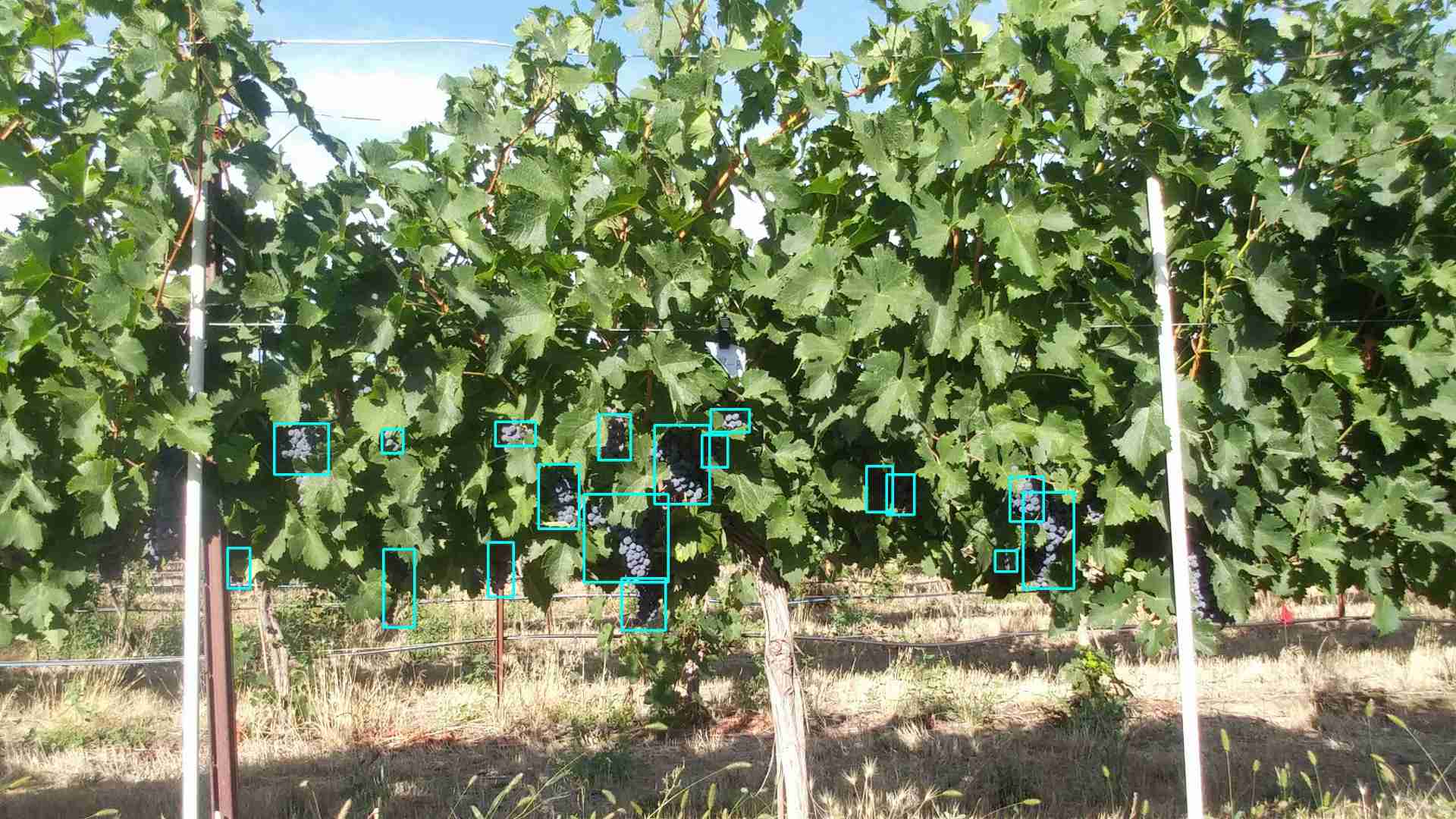} \label{fig.a2i}
  }
  \subfigure[]{
  \includegraphics[width = 7cm,height = 4.2cm]{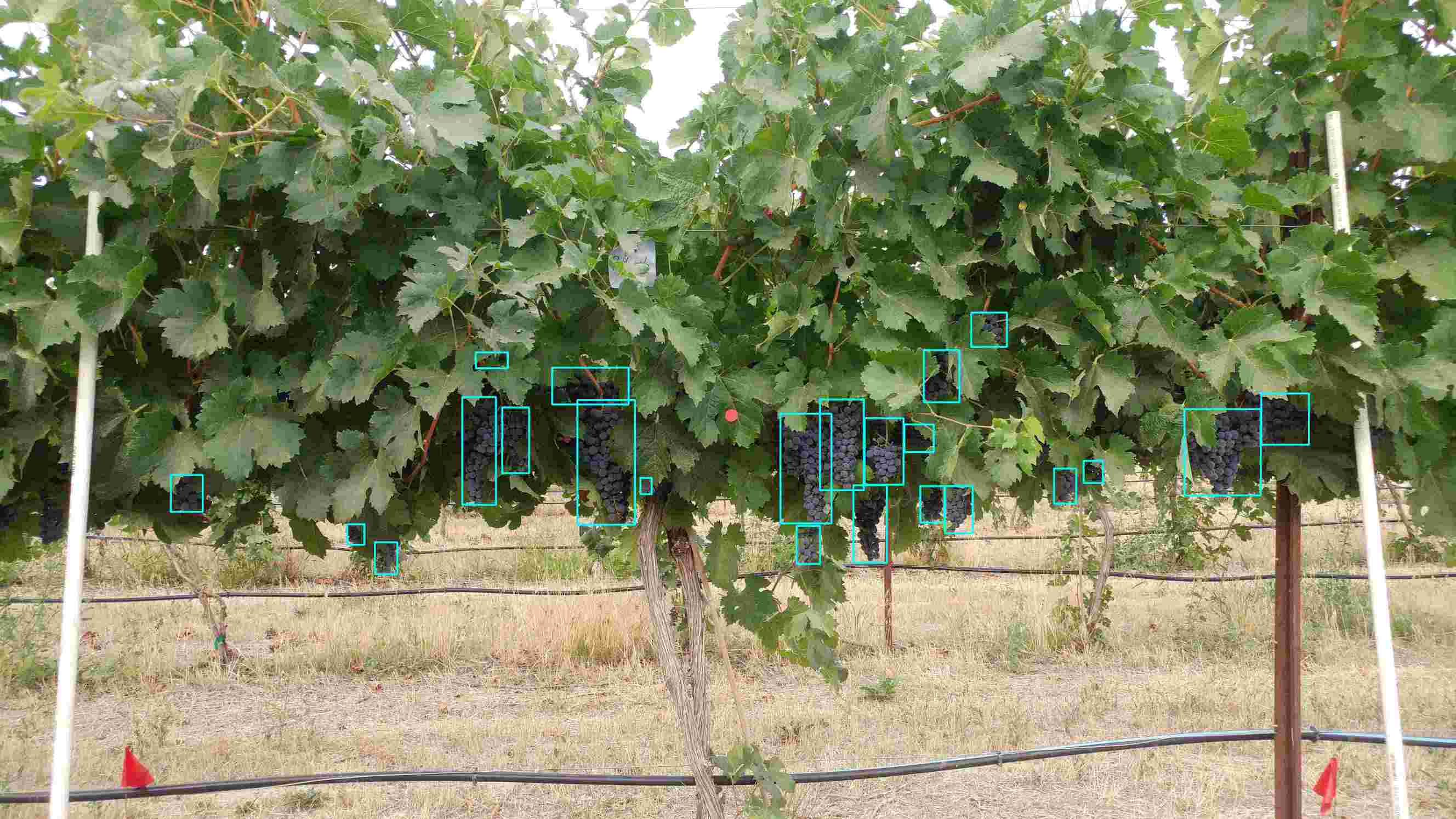} \label{fig.a2j}
  }
  \renewcommand*{\thefigure}{A.2}
  \caption{Demonstrations of detection results on the test set of Merlot (red variety) using (a-b) Faster R-CNN (bounding boxes in green color), (c-d) YOLOv3 (in blue color), (e-f) YOLOv4 in red color, (g-h) YOLOv5 (in magenta color), and (i-j) Swin-transformer-YOLOv5 (in cyan color) under sunny (left) and cloudy (right) weathers.}
  \label{figa2}
\end{figure}
\newpage
\begin{figure}[!h]
  \centering
  \subfigure[]{
  \includegraphics[width = 7cm,height = 4.2cm]{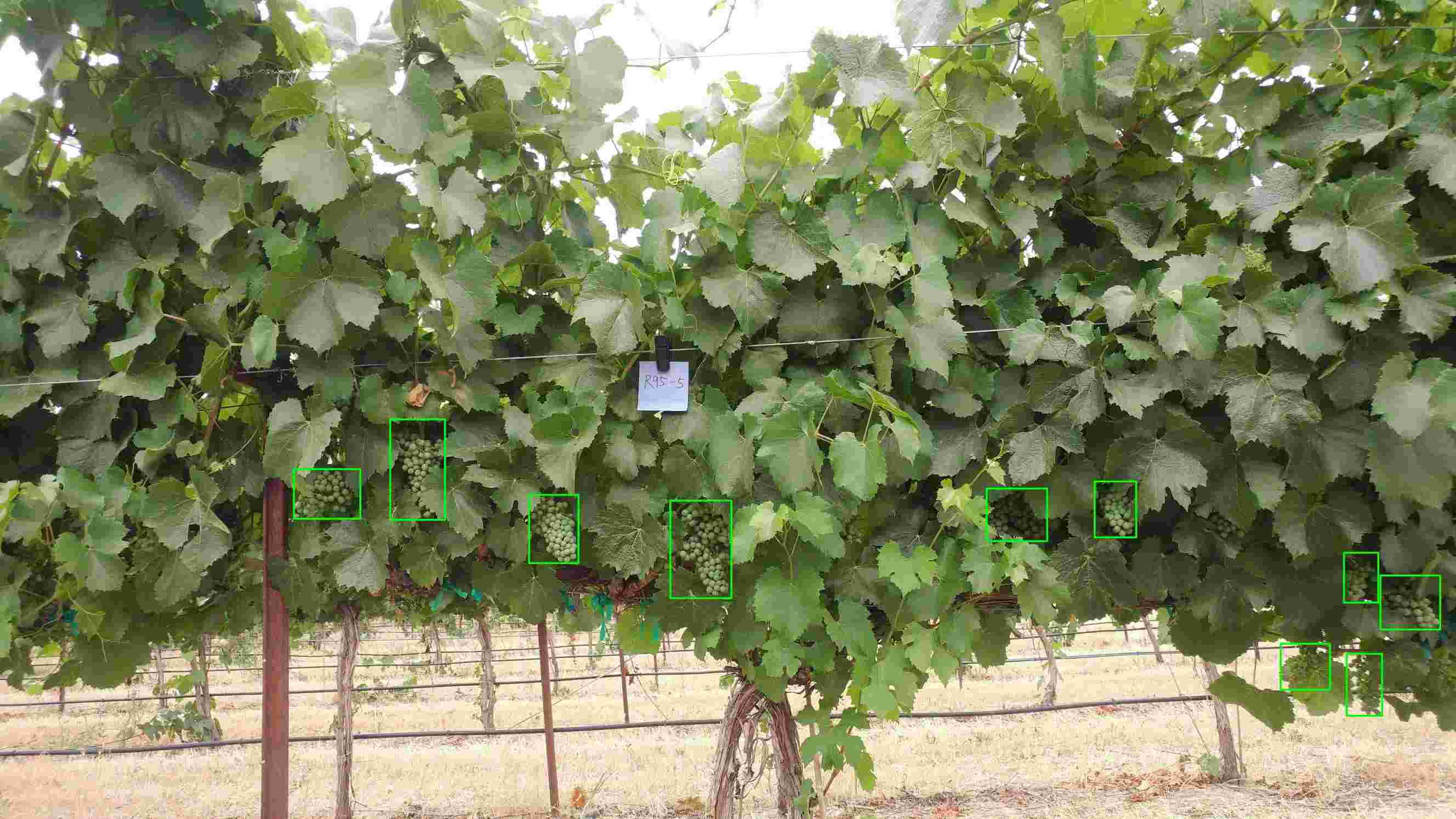} \label{fig.a3a}
  }
  \subfigure[]{
  \includegraphics[width = 7cm,height = 4.2cm]{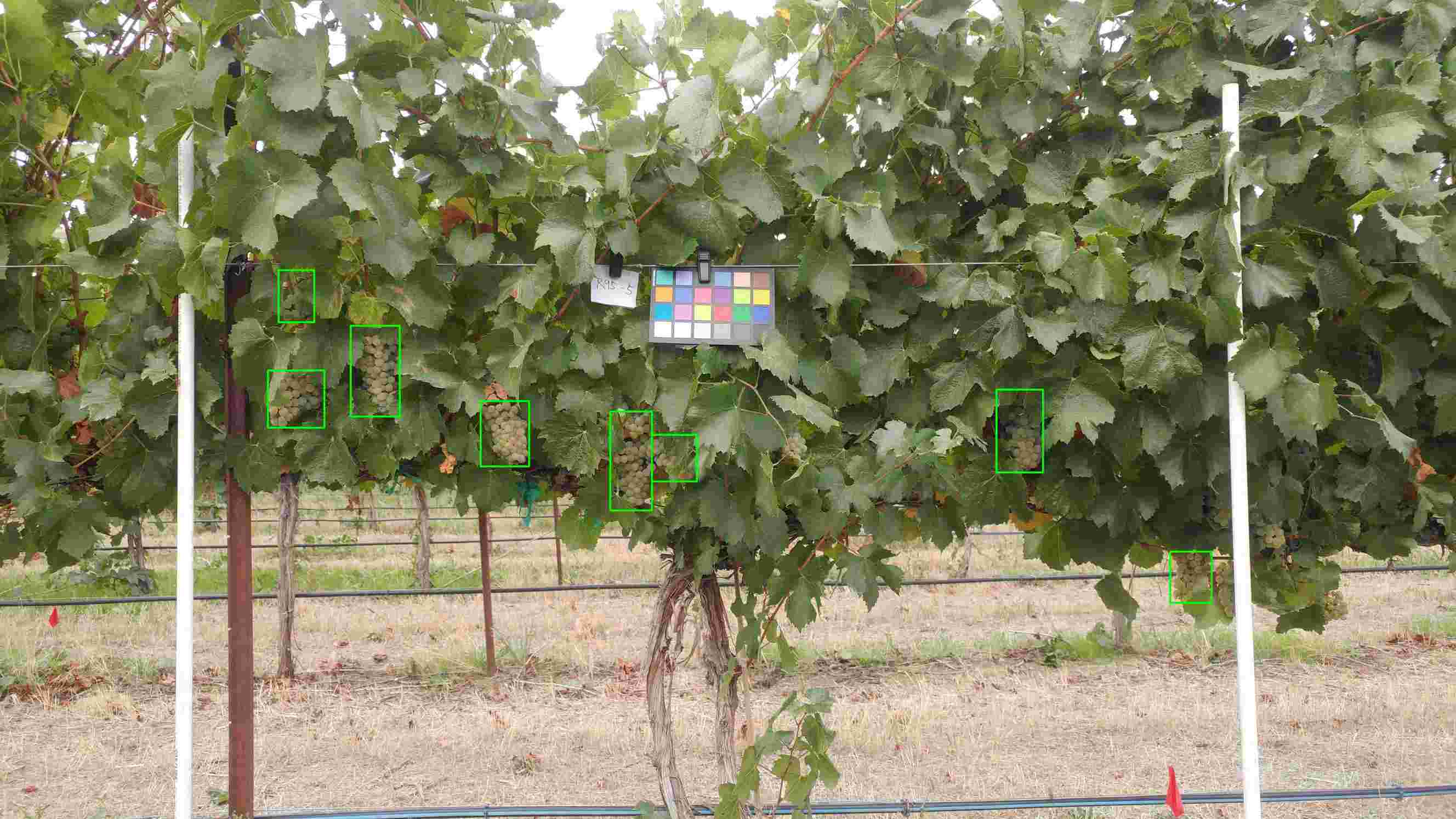} \label{fig.a3b}
  }
  \subfigure[]{
  \includegraphics[width = 7cm,height = 4.2cm]{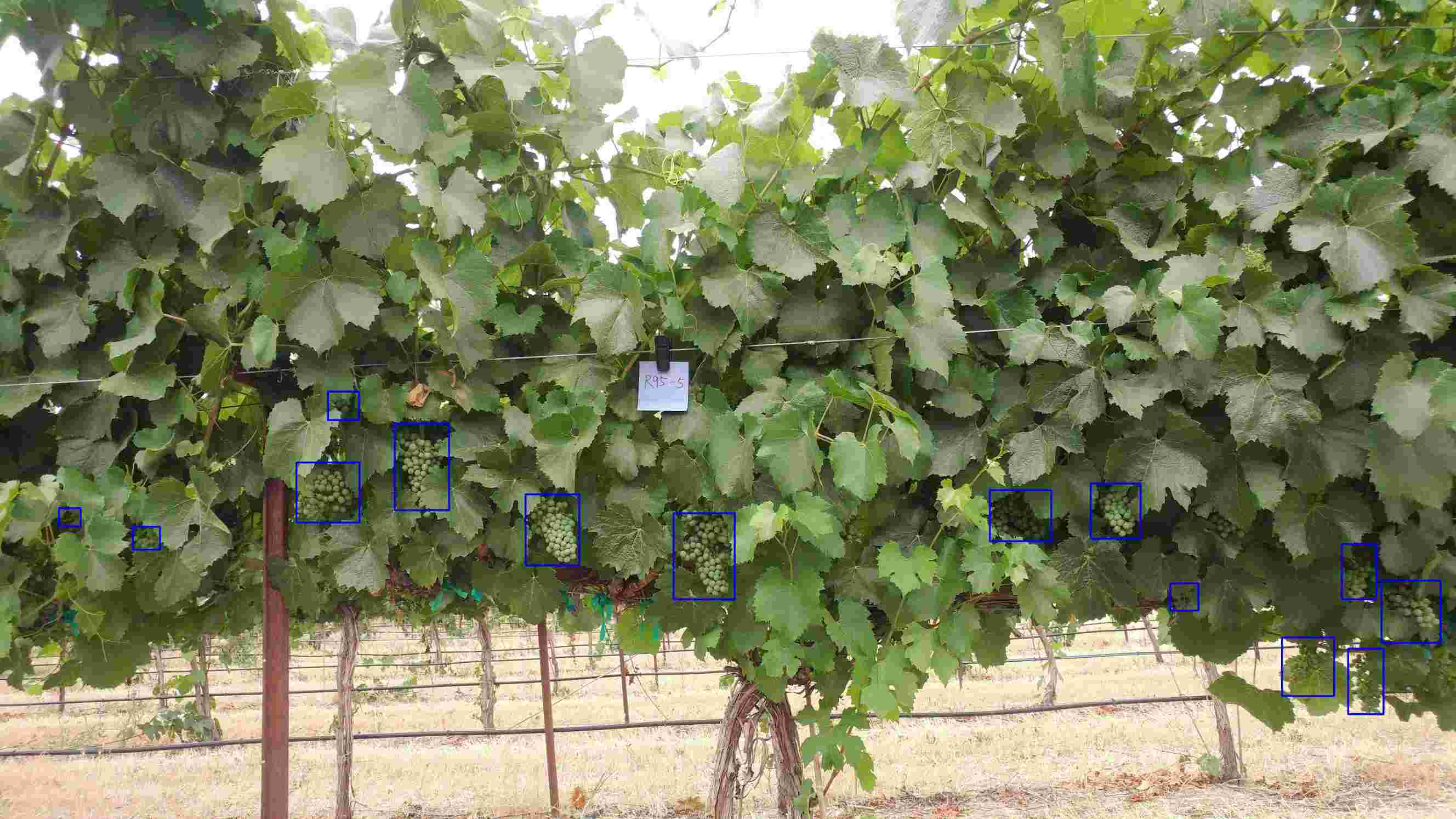} \label{fig.a3c}
  }
  \subfigure[]{
  \includegraphics[width = 7cm,height = 4.2cm]{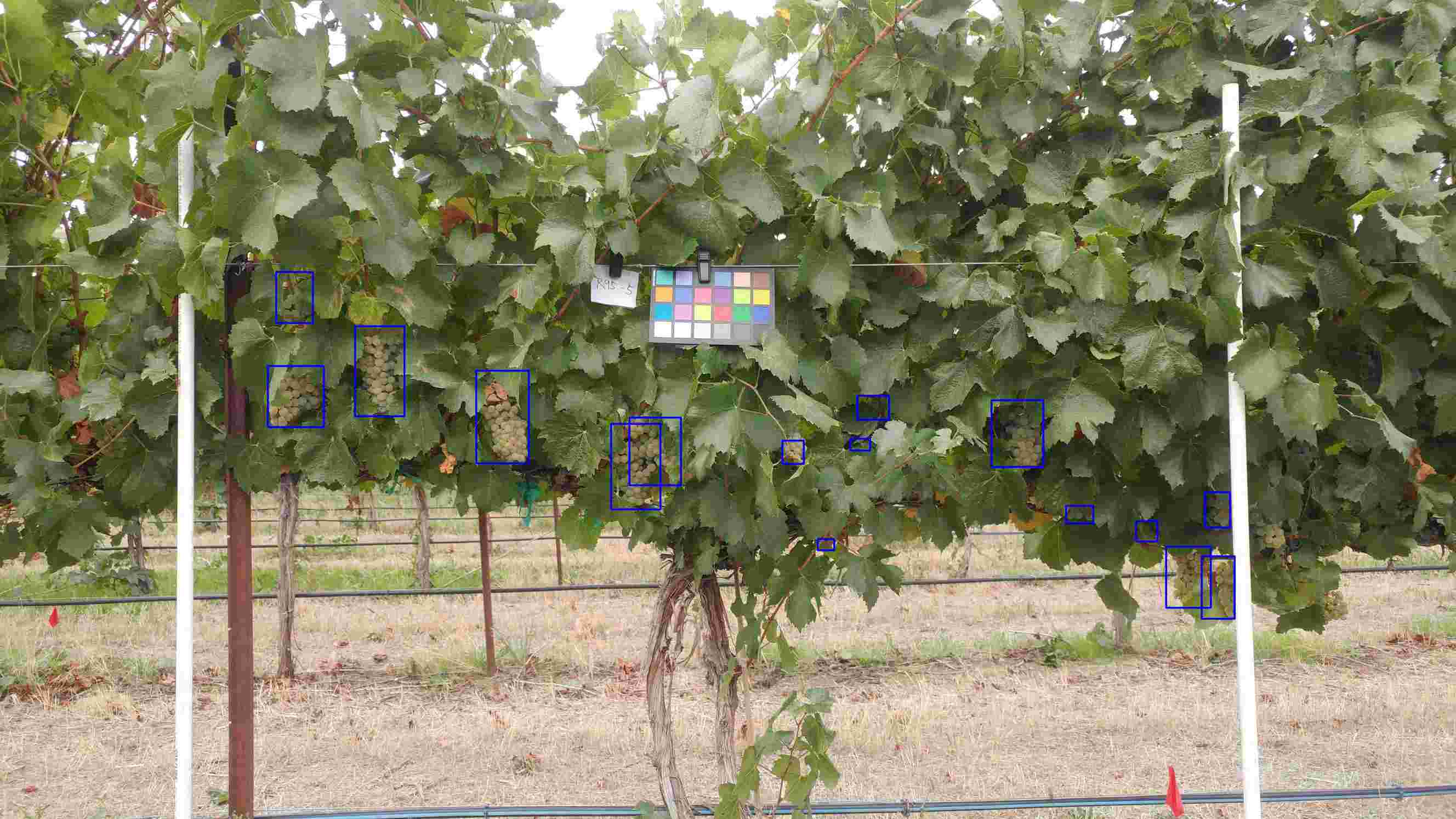} \label{fig.a3d}
  }
  \subfigure[]{
  \includegraphics[width = 7cm,height = 4.2cm]{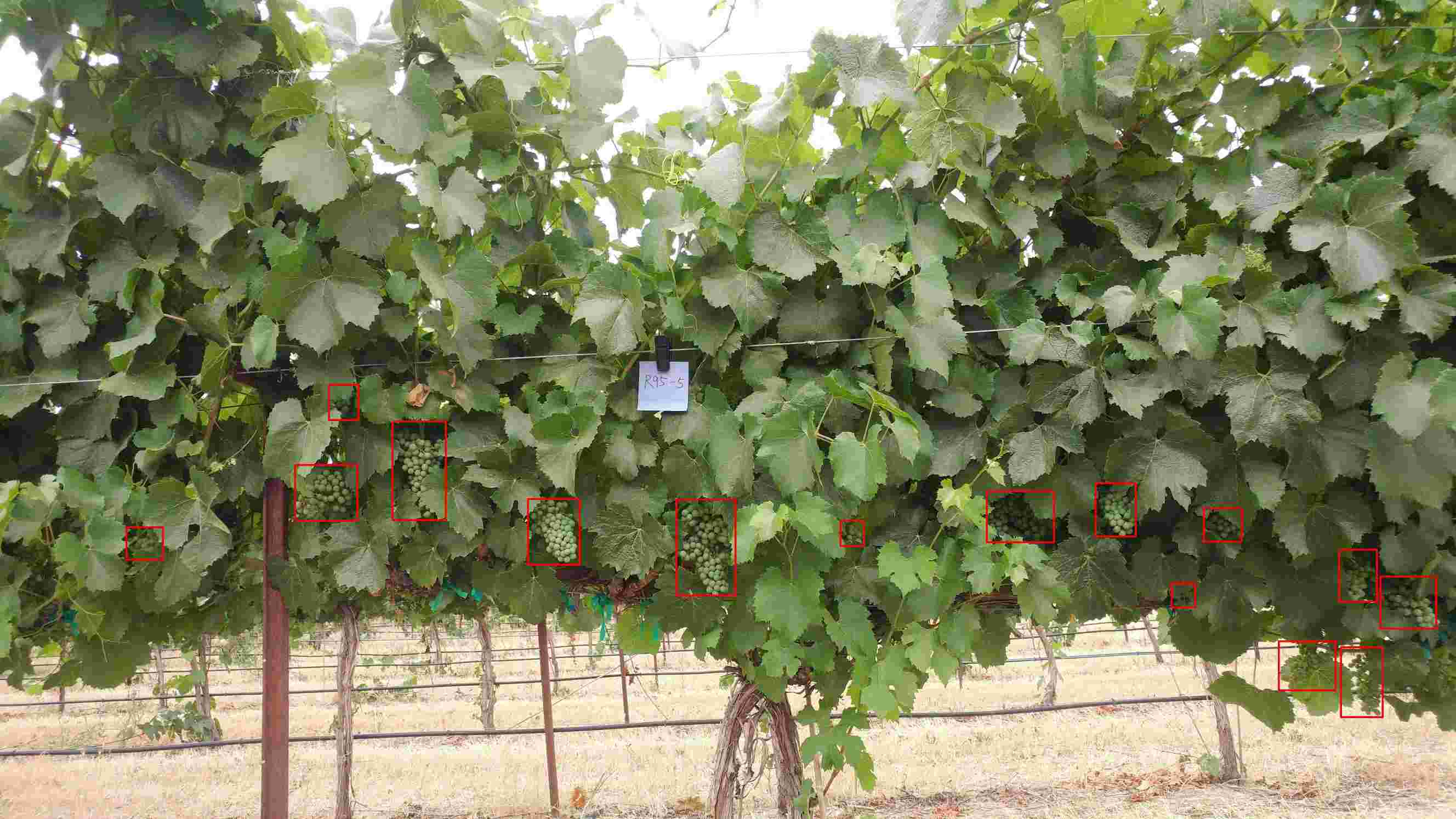} \label{fig.a3e}
  }
  \subfigure[]{
  \includegraphics[width = 7cm,height = 4.2cm]{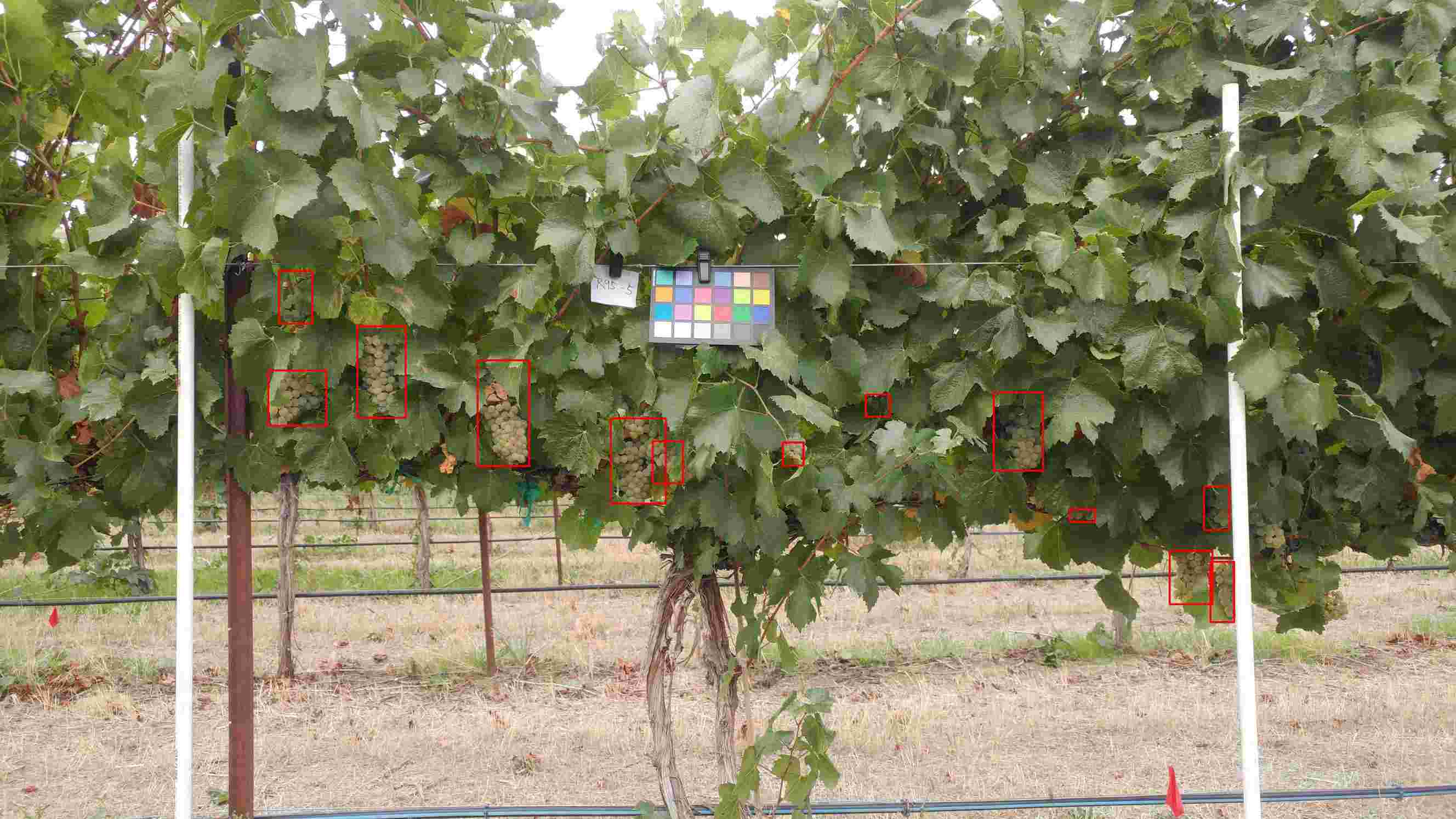} \label{fig.a3f}
  }
  \subfigure[]{
  \includegraphics[width = 7cm,height = 4.2cm]{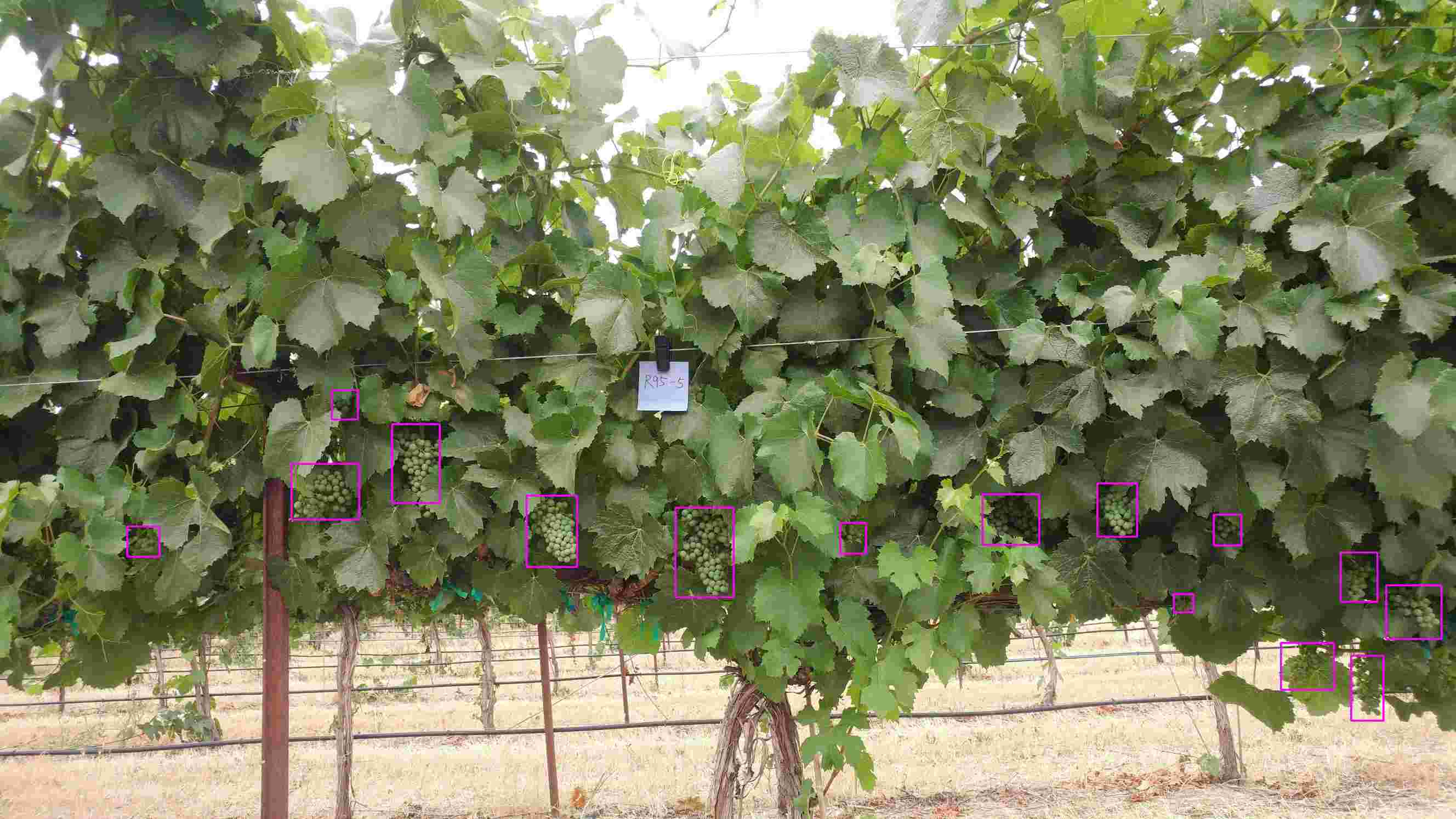} \label{fig.a3g}
  }
  \subfigure[]{
  \includegraphics[width = 7cm,height = 4.2cm]{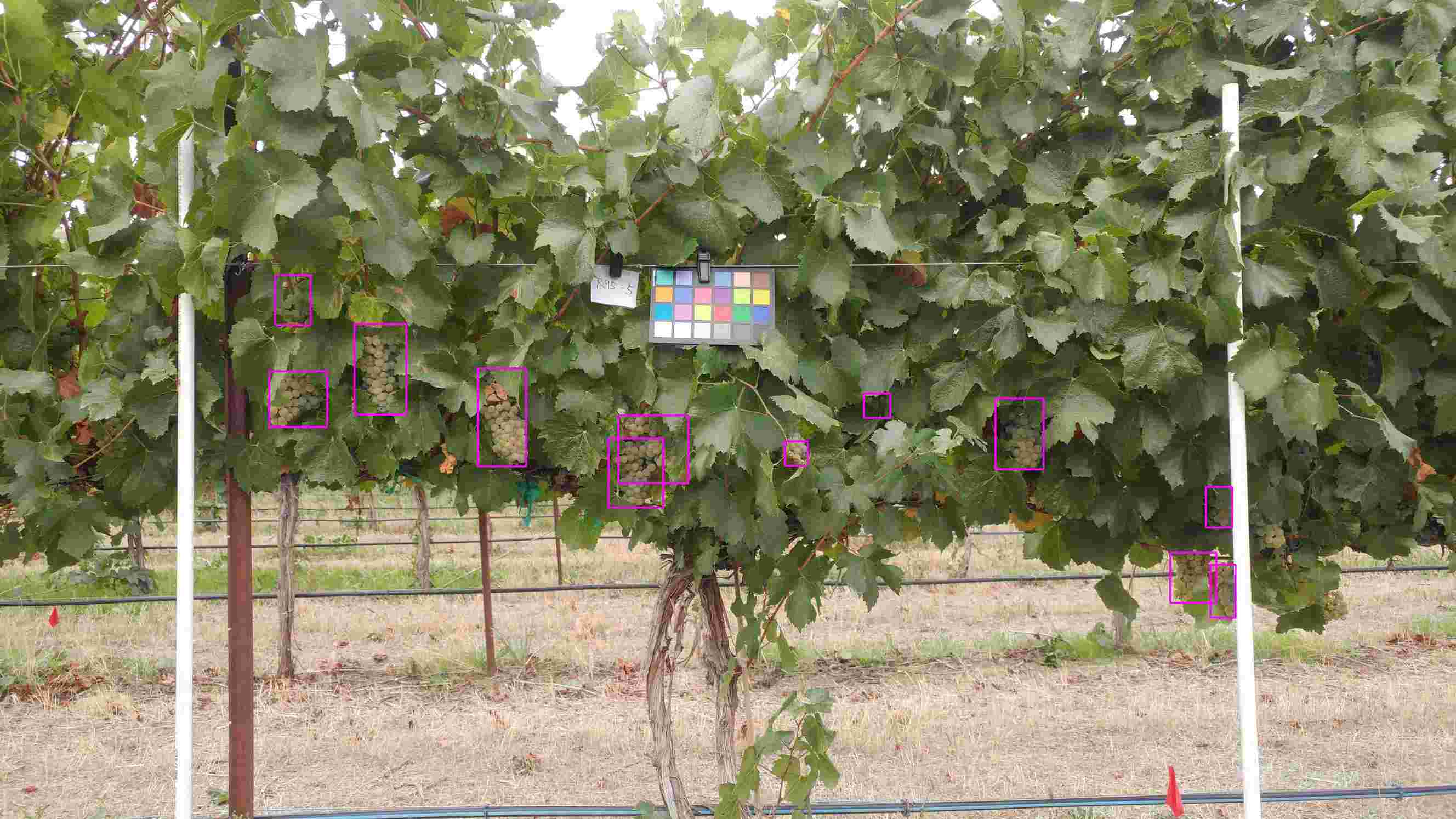} \label{fig.a3h}
  }
\end{figure}
\begin{figure}[H]
  \centering

  \subfigure[]{
  \includegraphics[width = 7cm,height = 4.2cm]{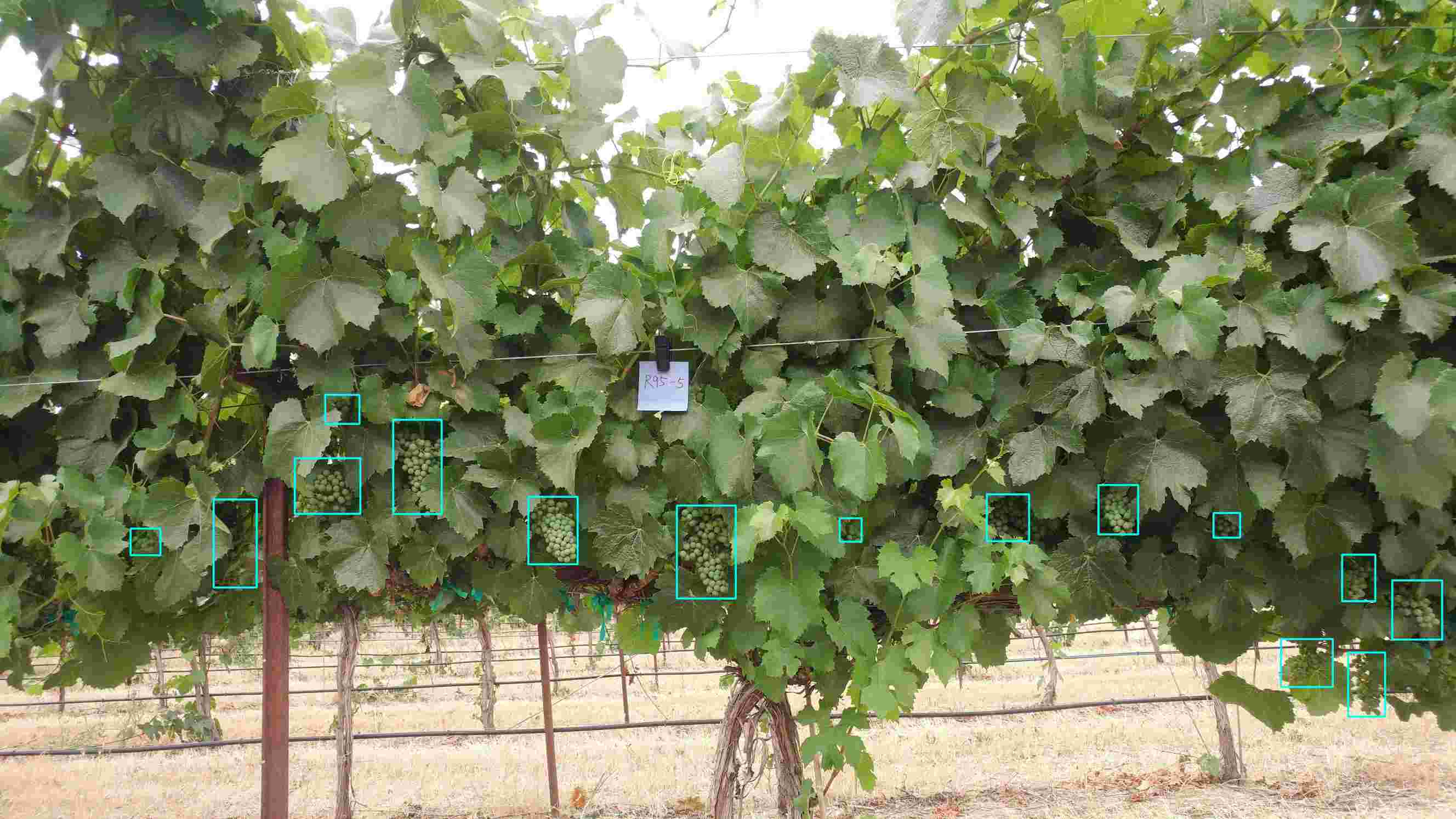} \label{fig.a3i}
  }
  \subfigure[]{
  \includegraphics[width = 7cm,height = 4.2cm]{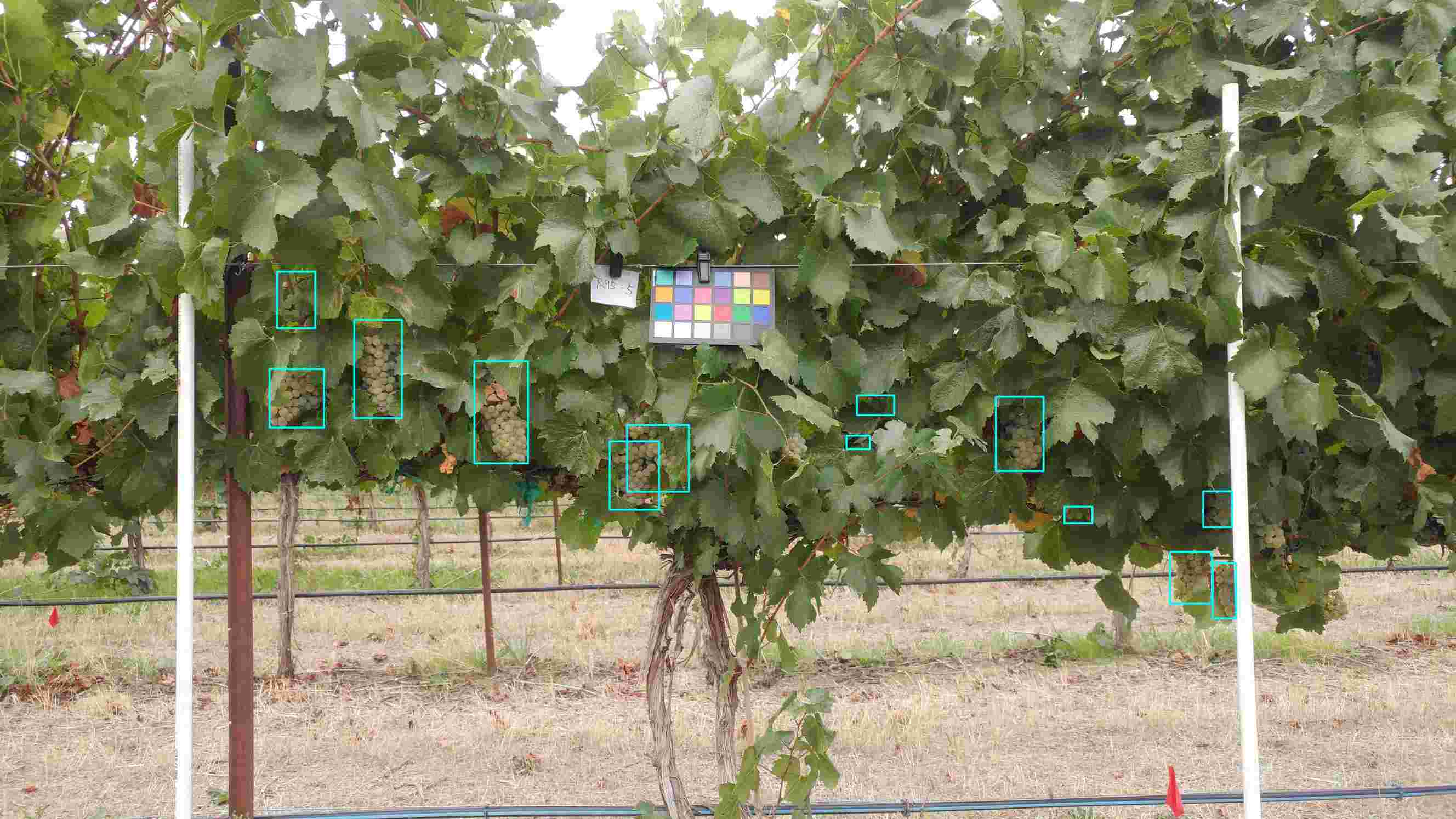} \label{fig.a3j}
  }
  \renewcommand*{\thefigure}{A.3}
  \caption{Demonstrations of detection results on the test set of Chardonnay (white variety) using (a-b) Faster R-CNN (bounding boxes in green color), (c-d) YOLOv3 (in blue color), (e-f) YOLOv4 (in red color), (g-h) YOLOv5 (in magenta color), and (i-j) Swin-transformer-YOLOv5 (in cyan color) at immature (left) and mature (right) stages.}
  \label{figa3}
\end{figure}
\newpage
\begin{figure}[!h]
  \centering
  \subfigure[]{
  \includegraphics[width = 7cm,height = 4.2cm]{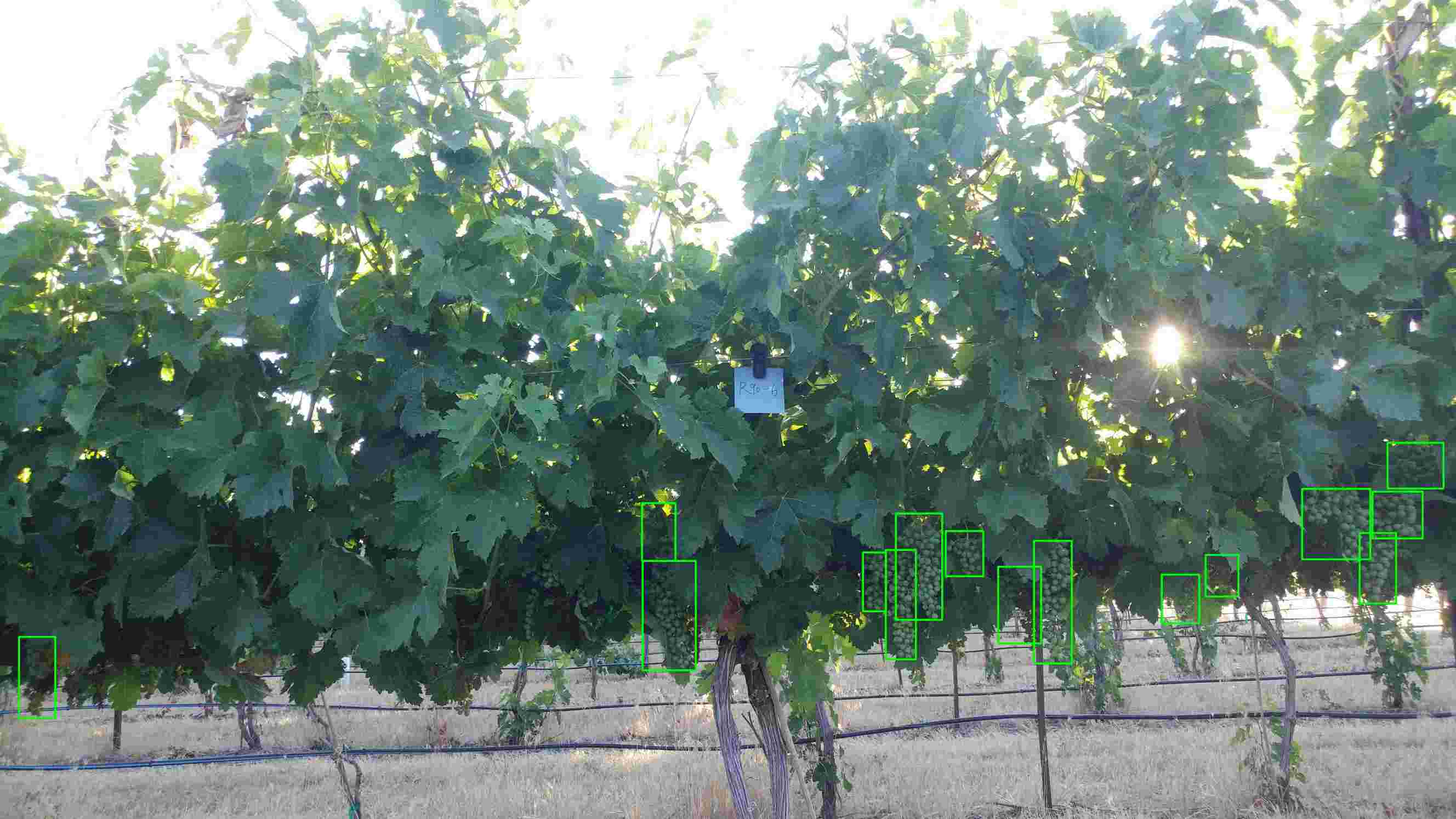} \label{fig.a4a}
  }
  \subfigure[]{
  \includegraphics[width = 7cm,height = 4.2cm]{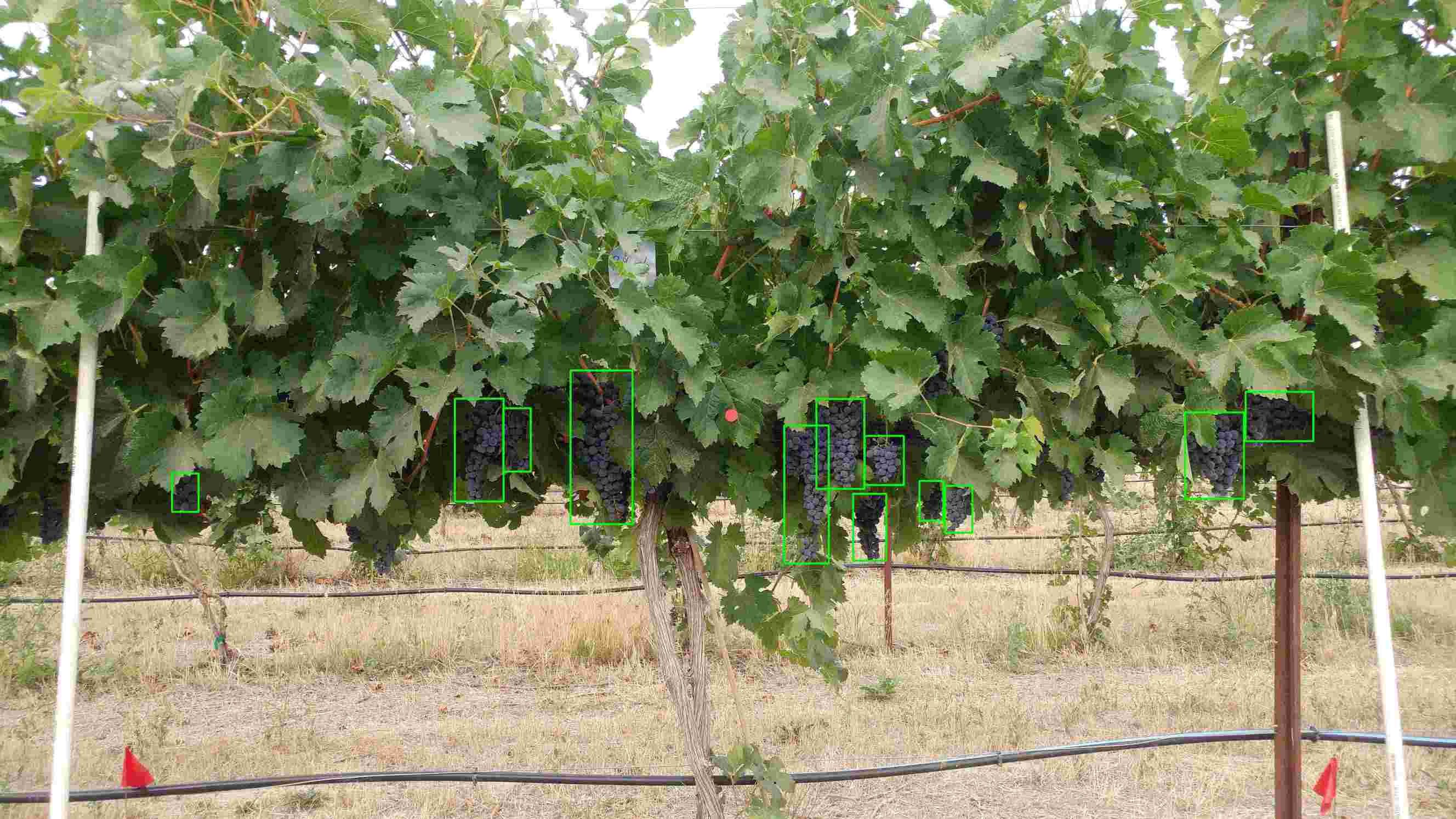} \label{fig.a4b}
  }
  \subfigure[]{
  \includegraphics[width = 7cm,height = 4.2cm]{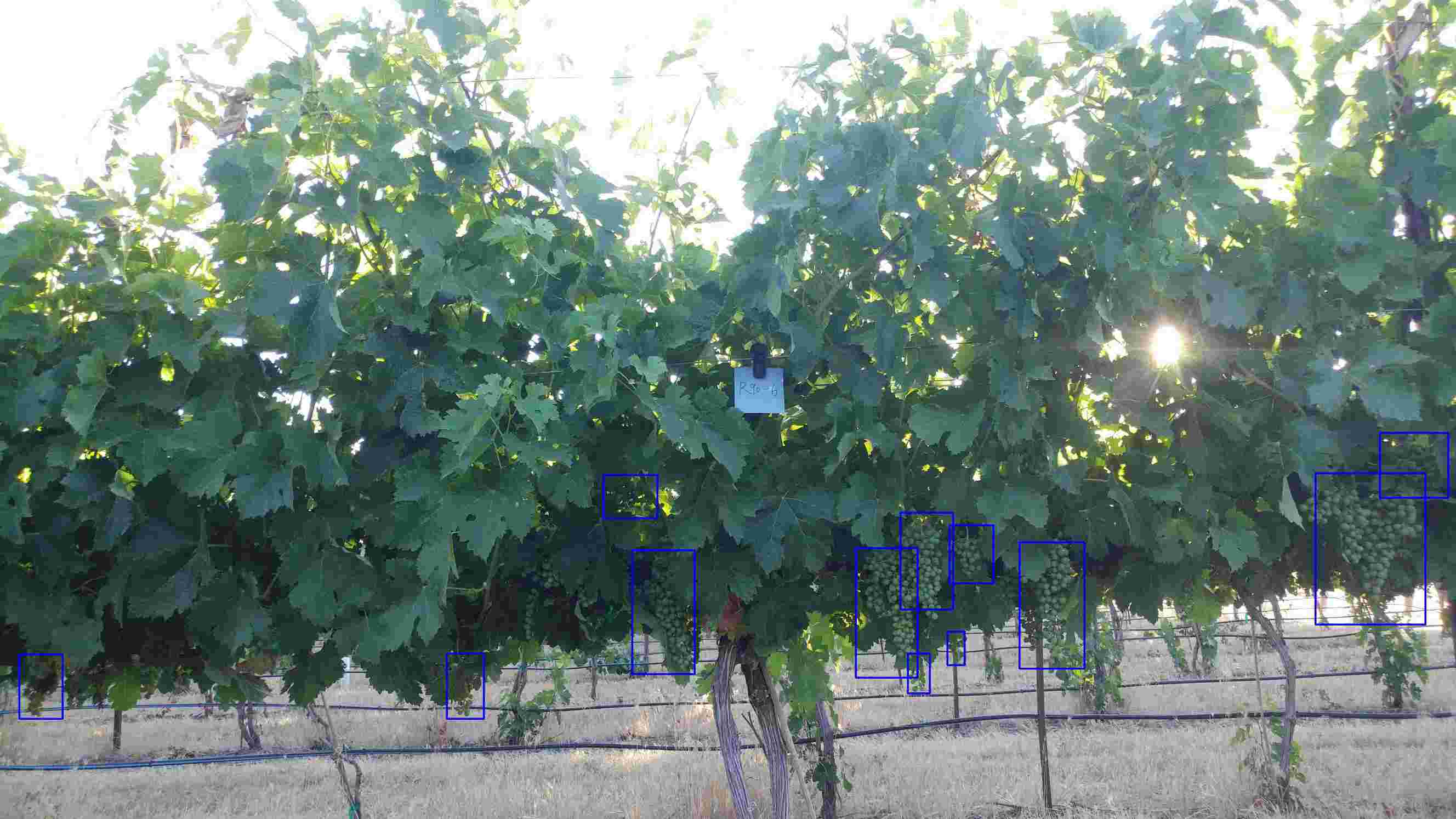} \label{fig.a4c}
  }
  \subfigure[]{
  \includegraphics[width = 7cm,height = 4.2cm]{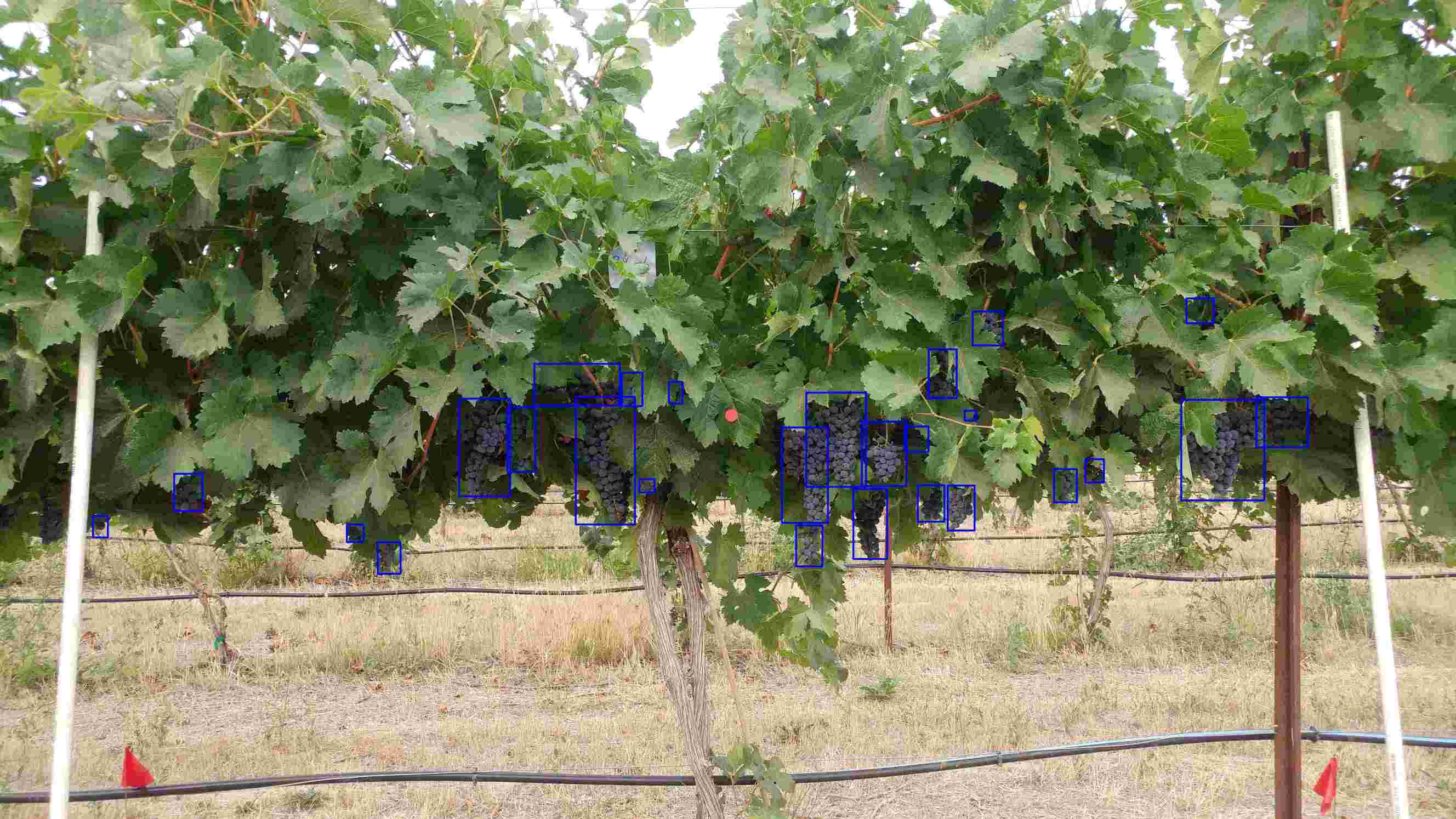} \label{fig.a4d}
  }
  \subfigure[]{
  \includegraphics[width = 7cm,height = 4.2cm]{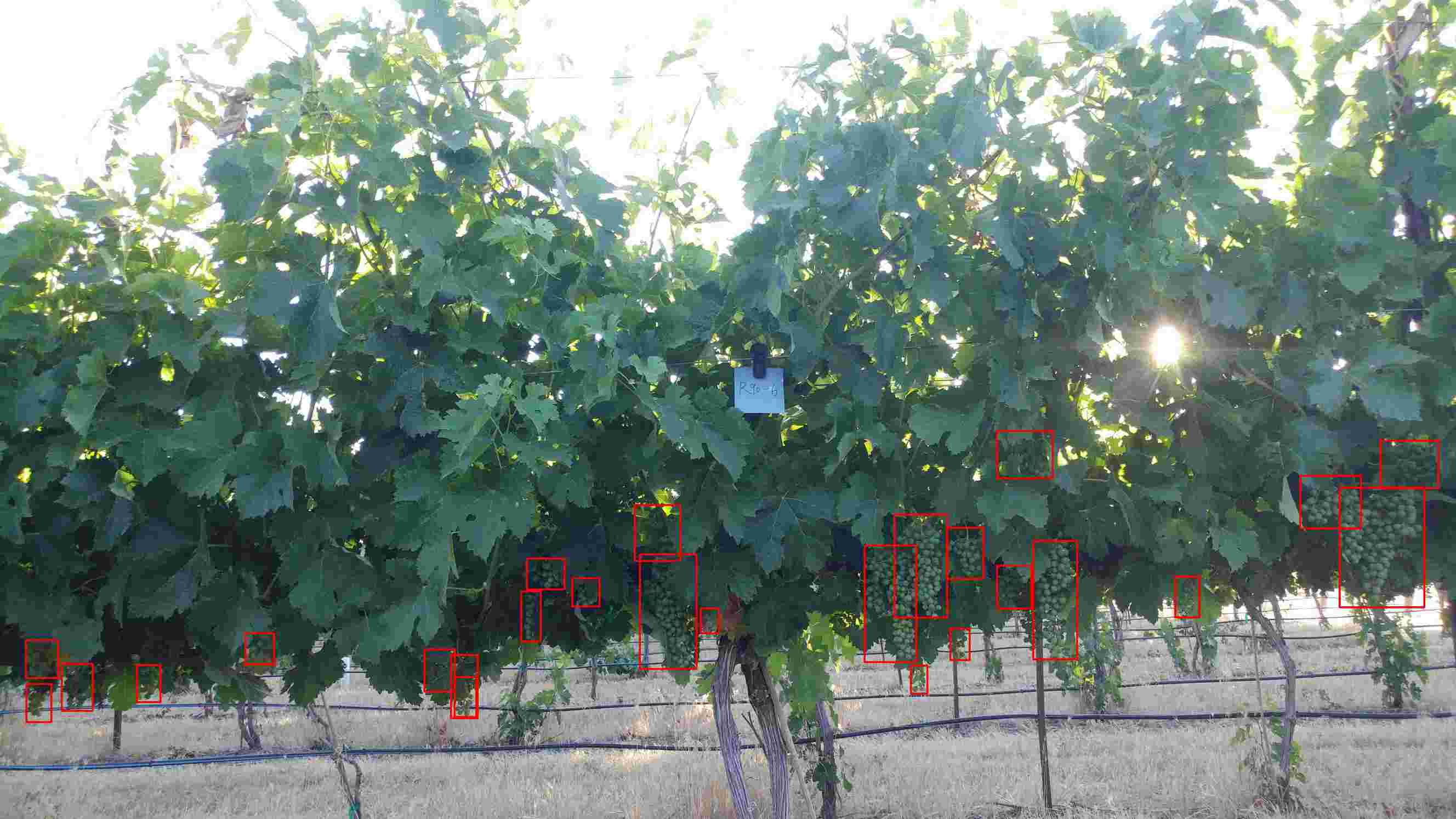} \label{fig.a4e}
  }
  \subfigure[]{
  \includegraphics[width = 7cm,height = 4.2cm]{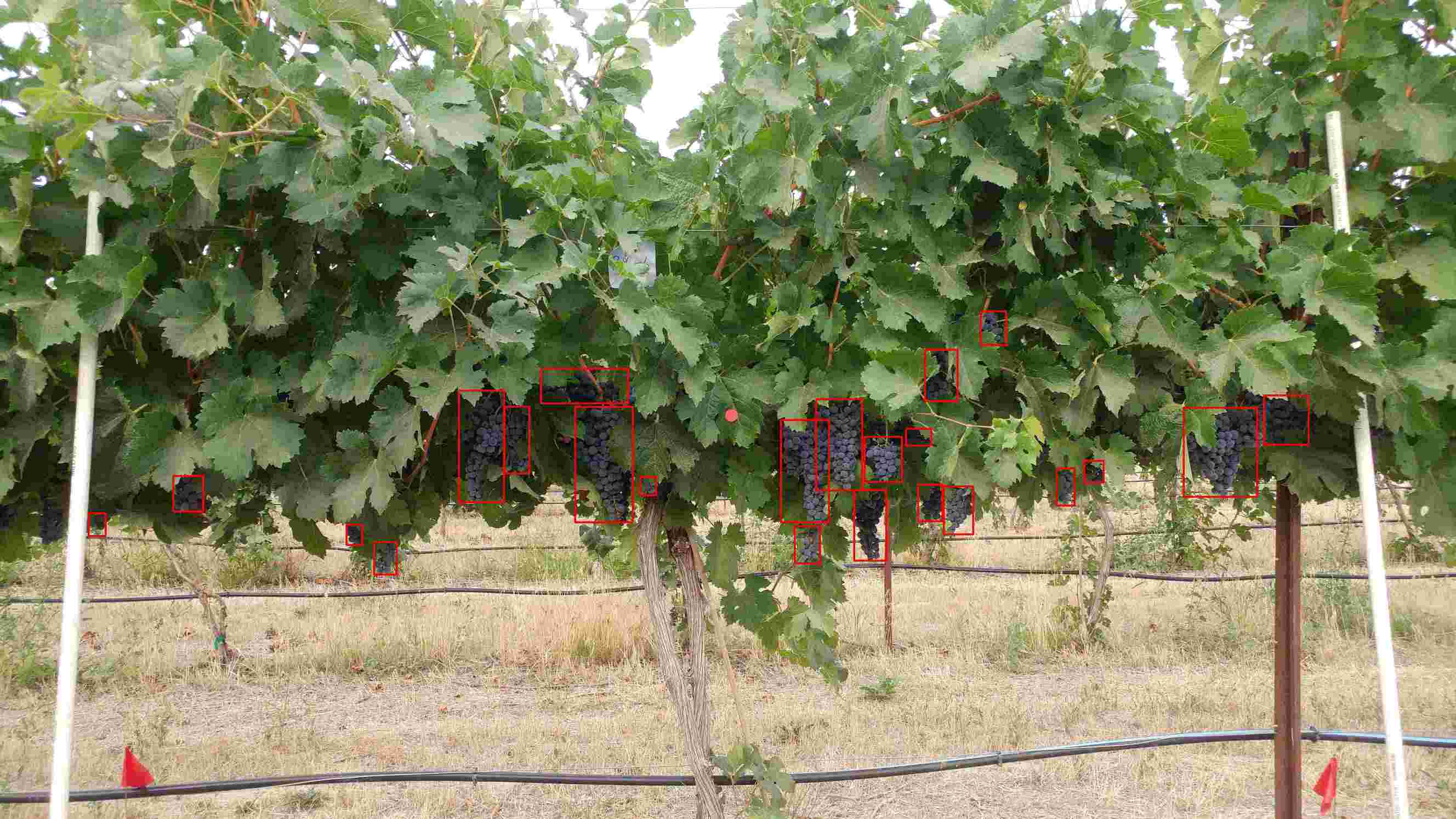} \label{fig.a4f}
  }
  \subfigure[]{
  \includegraphics[width = 7cm,height = 4.2cm]{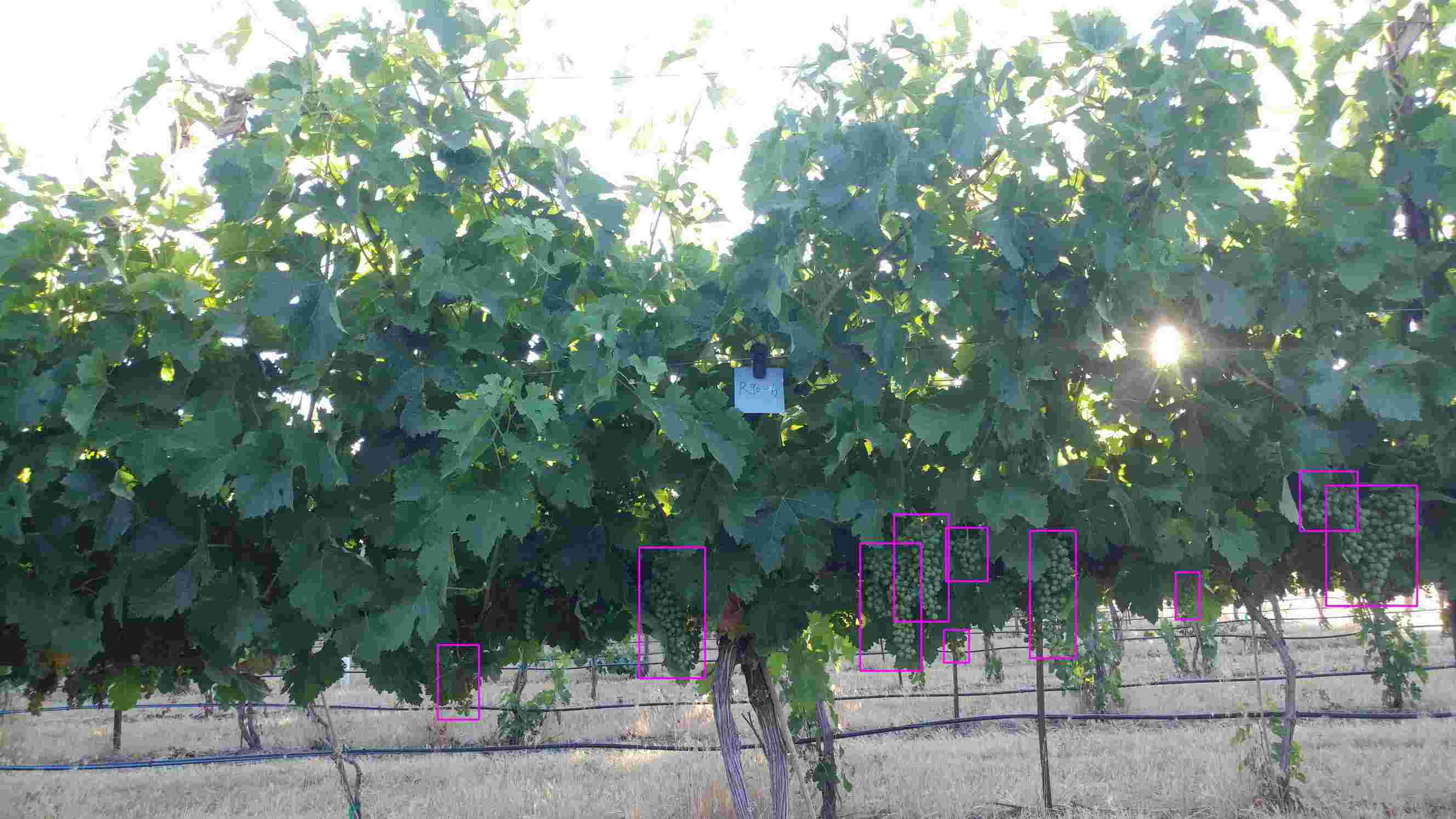} \label{fig.a4g}
  }
  \subfigure[]{
  \includegraphics[width = 7cm,height = 4.2cm]{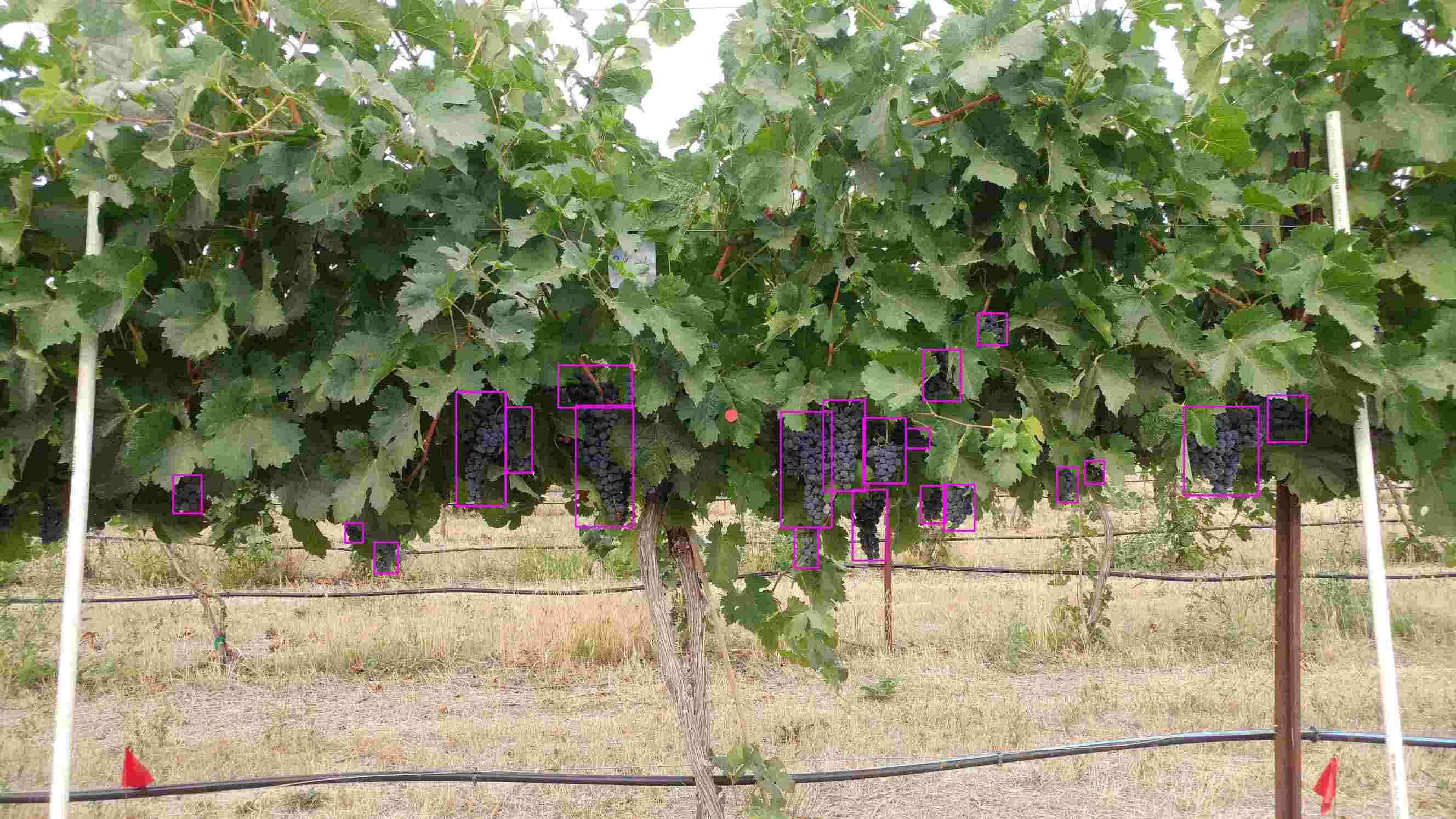} \label{fig.a4h}
  }
\end{figure}
\begin{figure}[H]
  \centering

  \subfigure[]{
  \includegraphics[width = 7cm,height = 4.2cm]{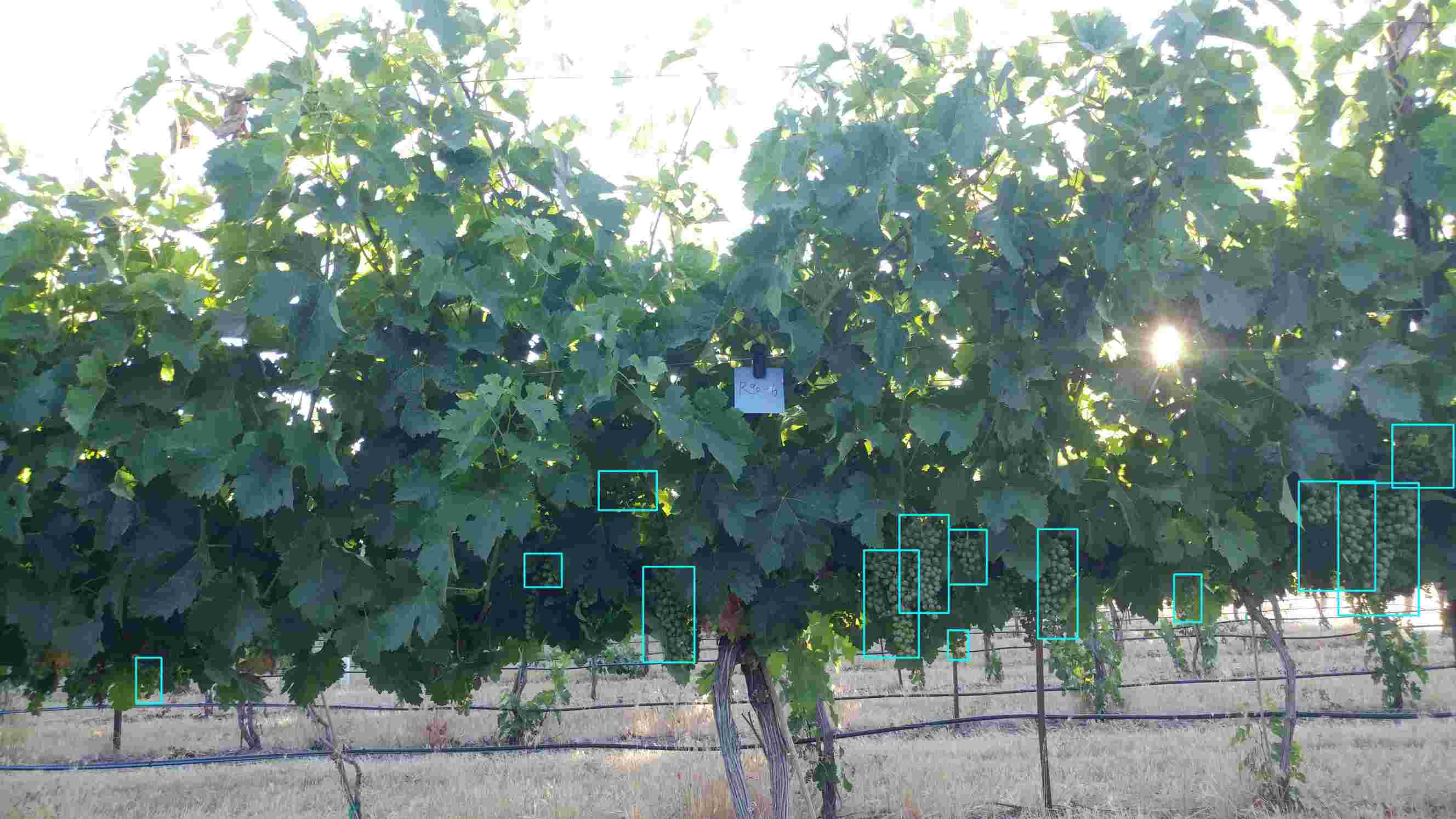} \label{fig.a4i}
  }
  \subfigure[]{
  \includegraphics[width = 7cm,height = 4.2cm]{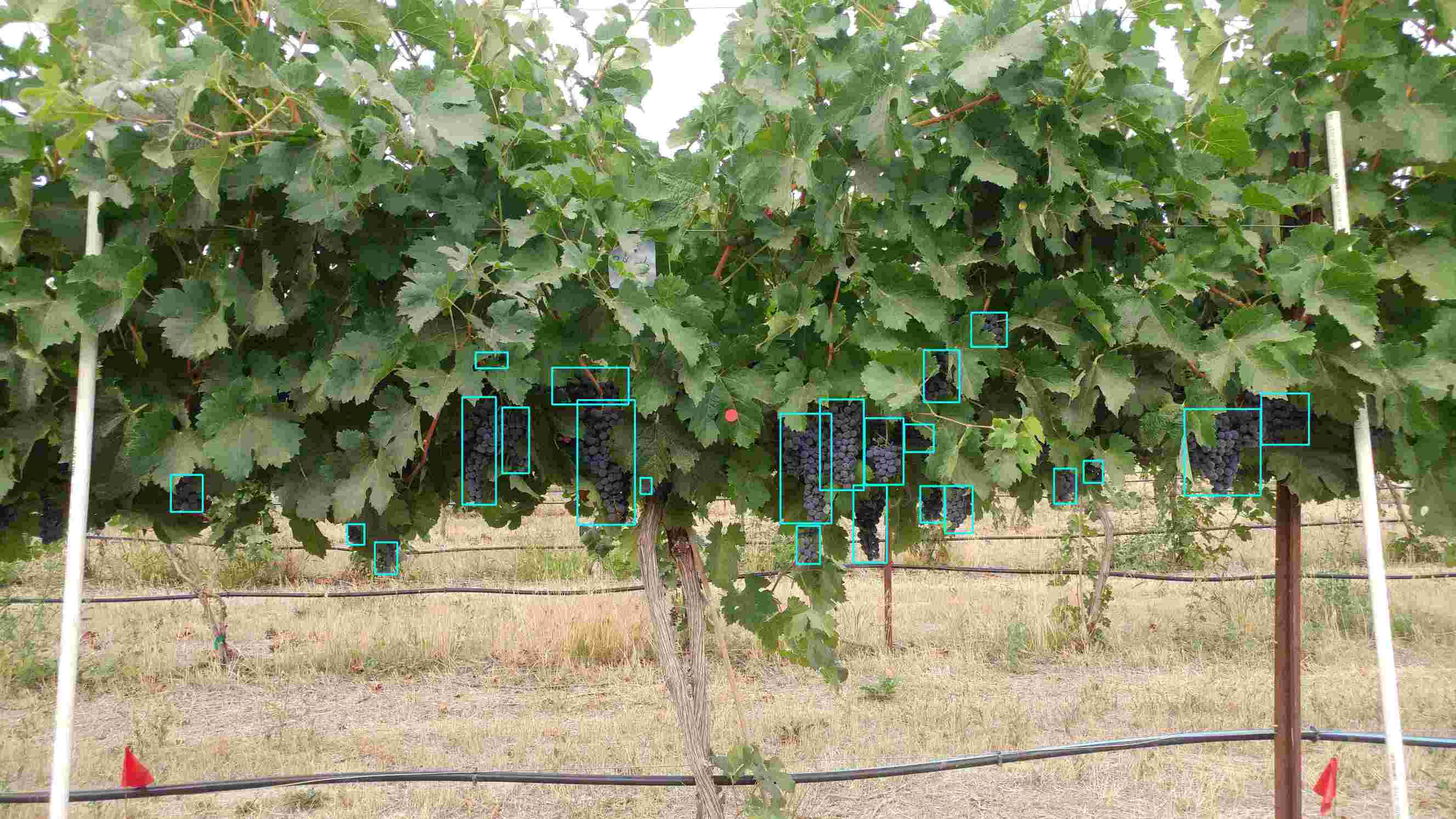} \label{fig.a4j}
  }
  \renewcommand*{\thefigure}{A.4}
  \caption{Demonstrations of detection results on the test set of Merlot (red variety) using (a-b) Faster R-CNN (bounding boxes in green color), (c-d) YOLOv3 (in blue color), (e-f) YOLOv4 (in red color), (g-h) YOLOv5 (in magenta color), and (i-j) Swin-transformer-YOLOv5 (in cyan color) at immature (left) and mature (right) stages.}
  \label{figa4}
\end{figure}
\newpage
\begin{figure}[!h]
  \centering
  \subfigure[]{
  \includegraphics[width = 5cm,height = 3cm]{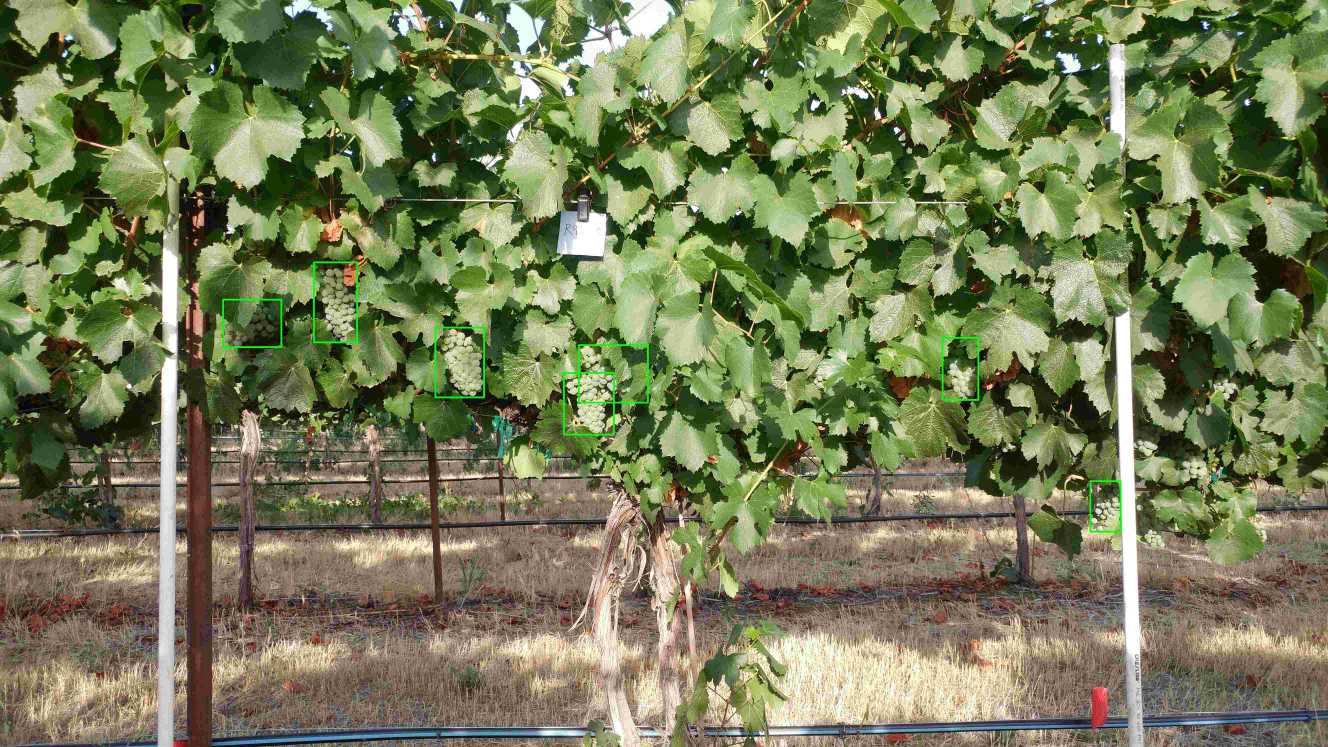} \label{fig.a5a}
  }
  \subfigure[]{
  \includegraphics[width = 5cm,height = 3cm]{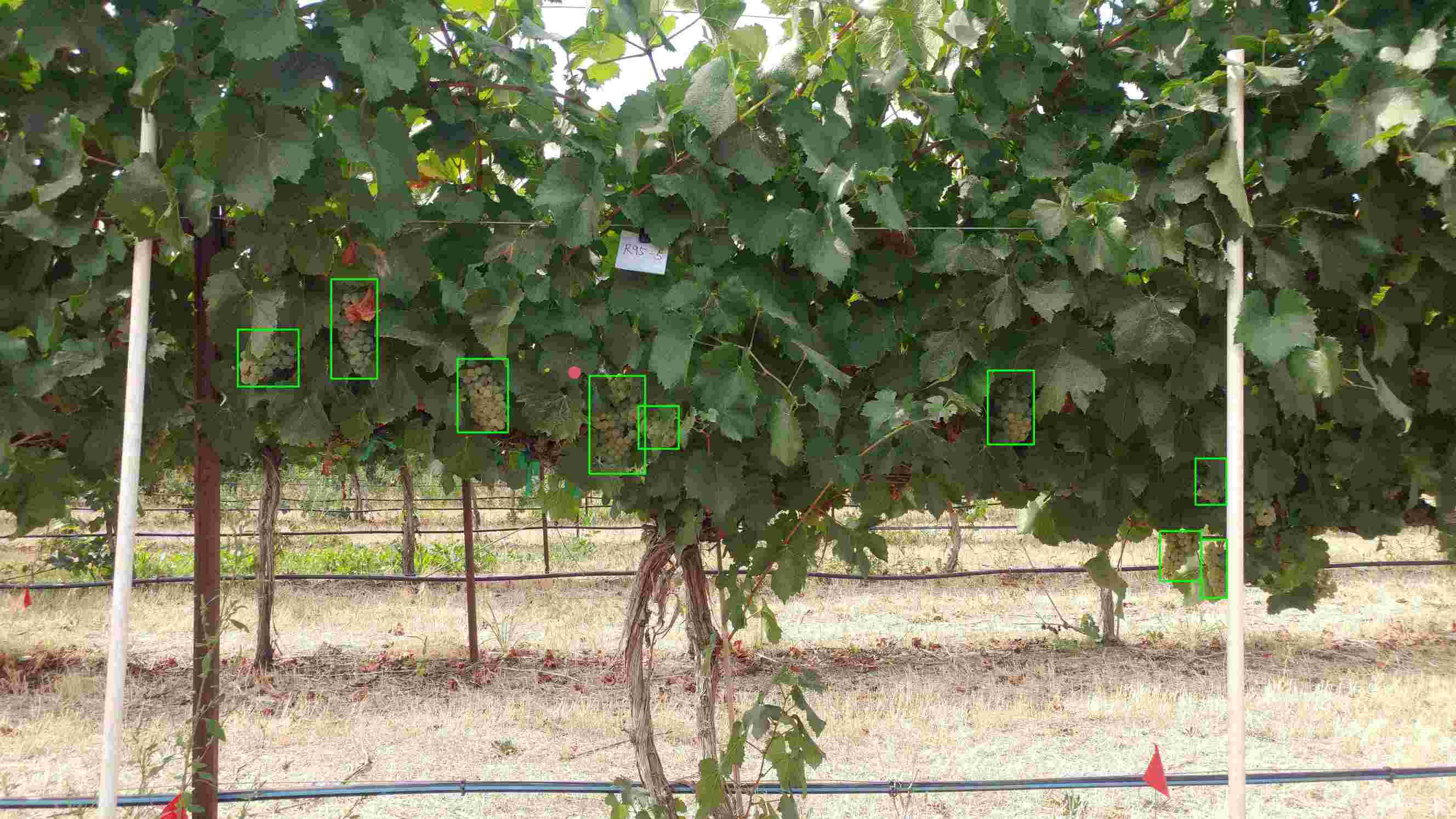} \label{fig.a5b}
  }
  \subfigure[]{
  \includegraphics[width = 5cm,height = 3cm]{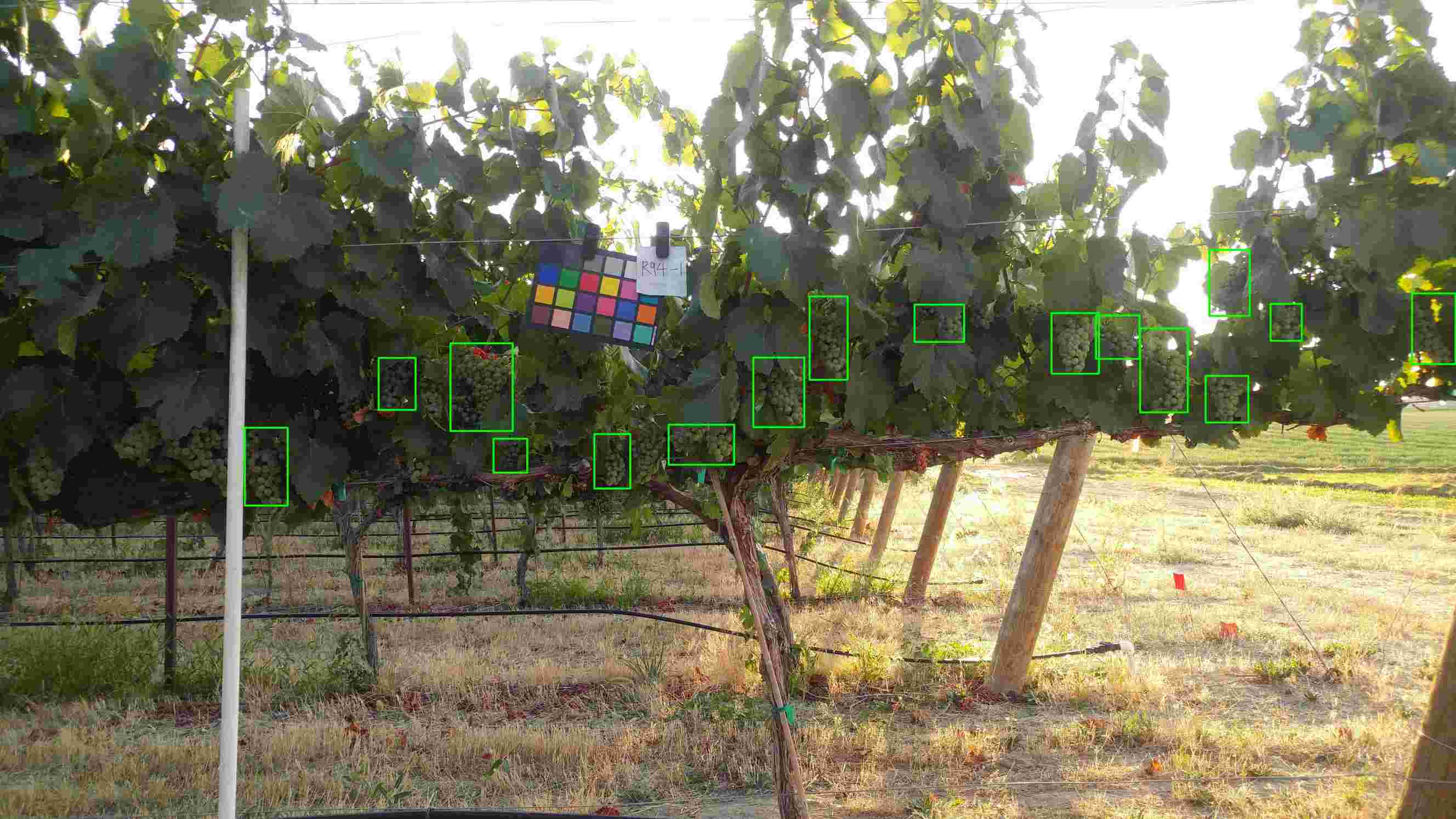} \label{fig.a5c}
  }
  \subfigure[]{
  \includegraphics[width = 5cm,height = 3cm]{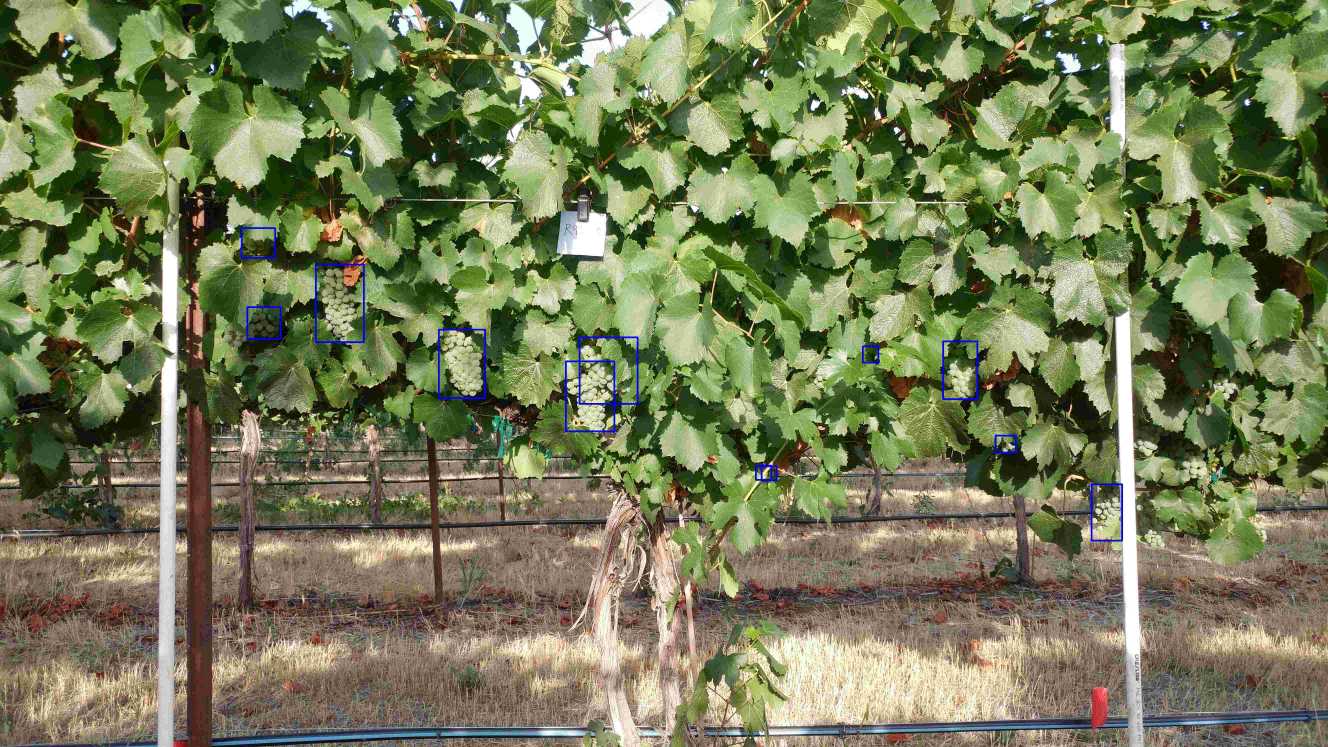} \label{fig.a5d}
  }
  \subfigure[]{
  \includegraphics[width = 5cm,height = 3cm]{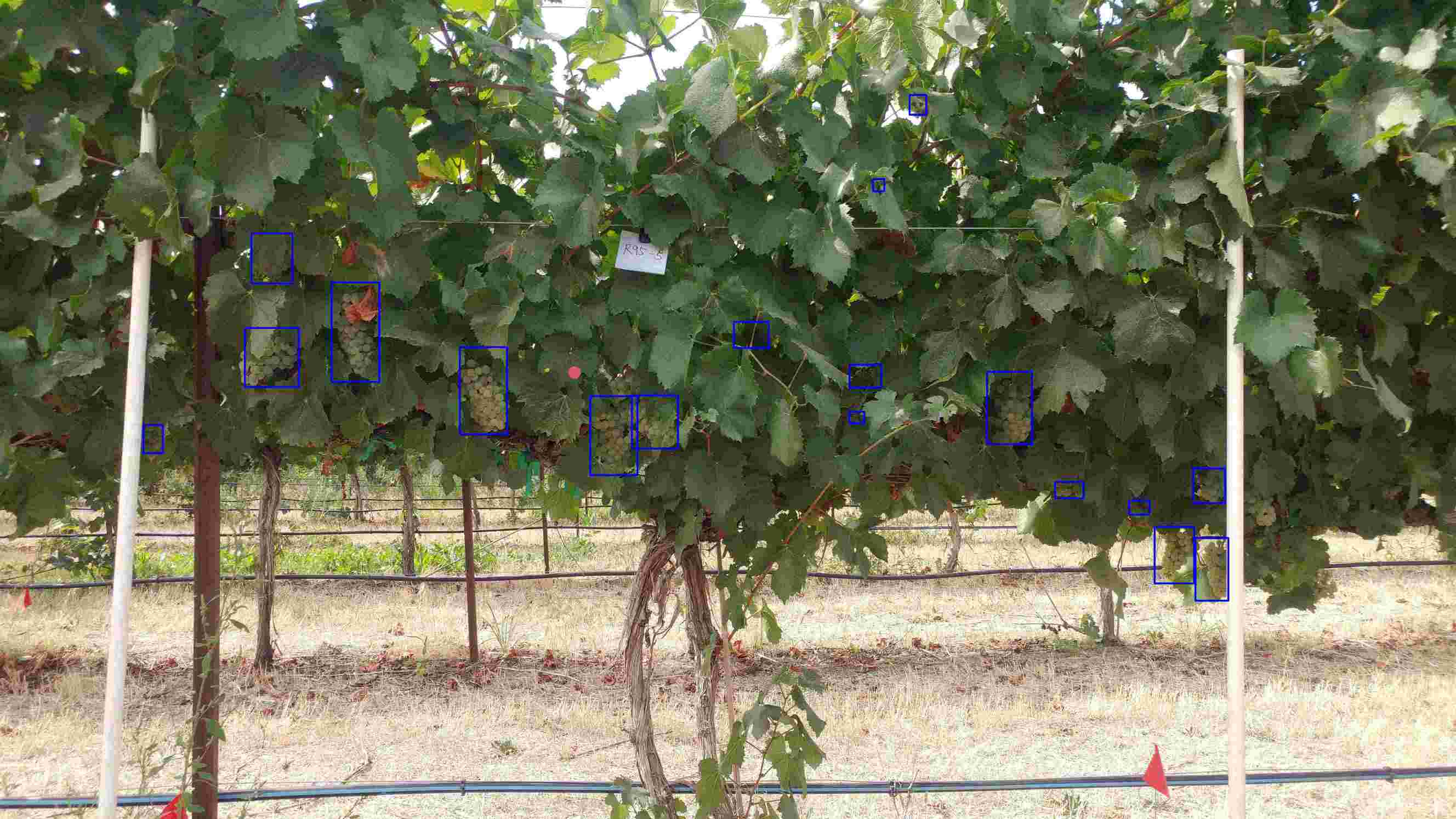} \label{fig.a5e}
  }
  \subfigure[]{
  \includegraphics[width = 5cm,height = 3cm]{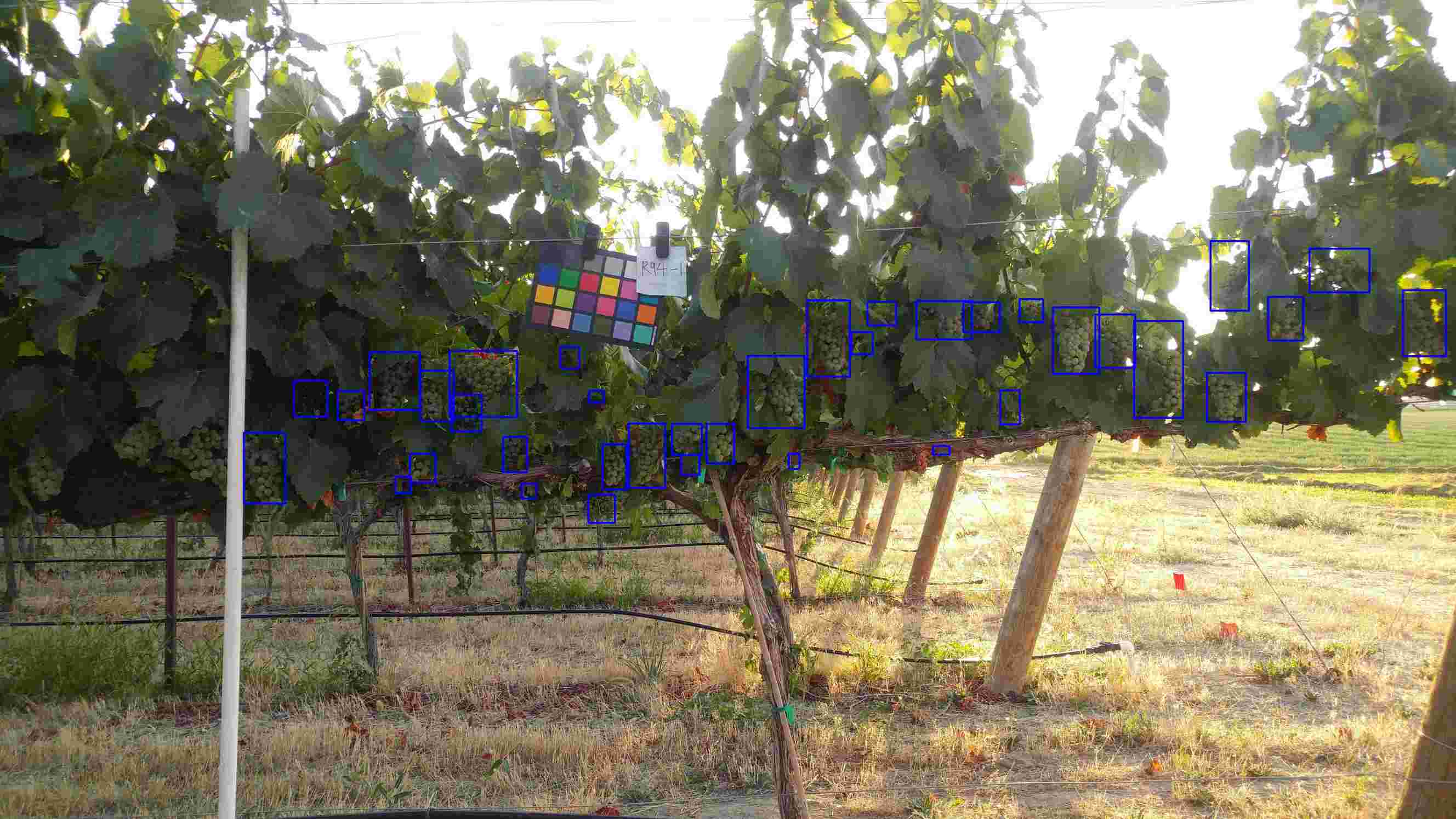} \label{fig.a5f}
  }
  \subfigure[]{
  \includegraphics[width =5cm,height = 3cm]{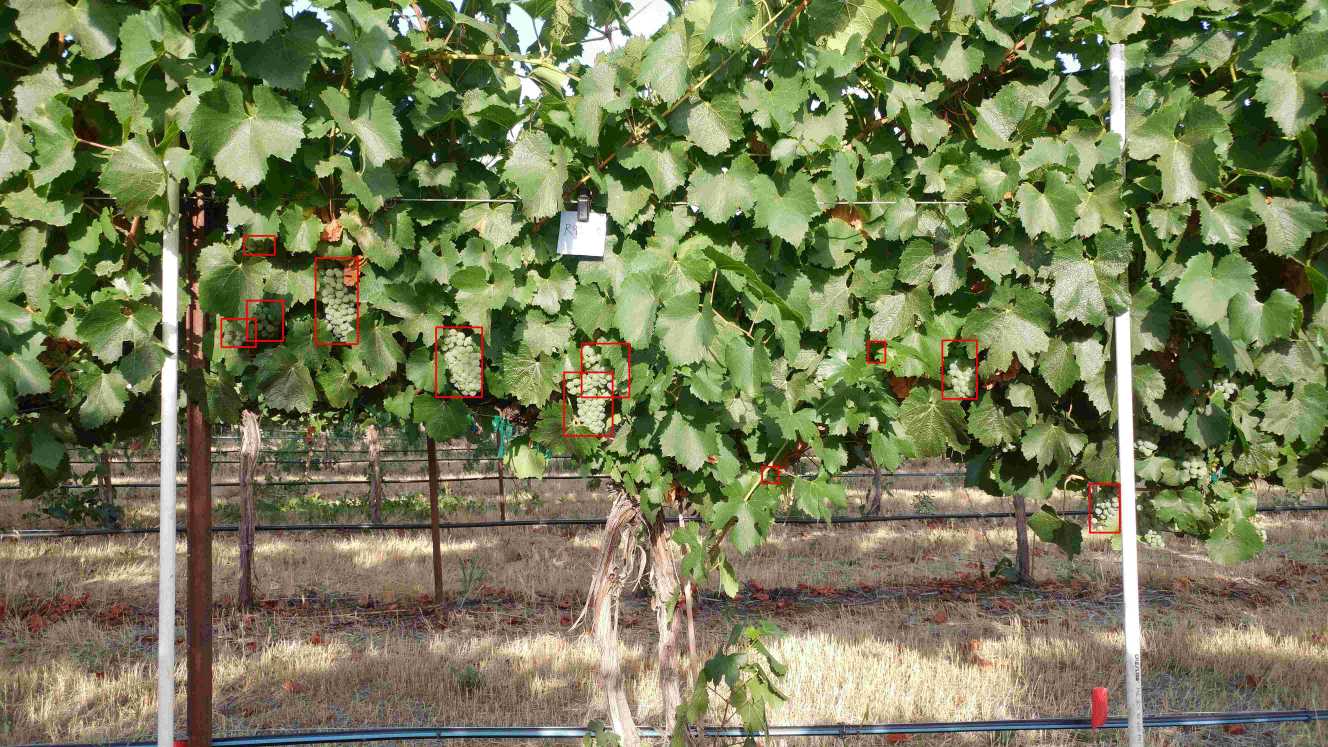} \label{fig.a5g}
  }
  \subfigure[]{
  \includegraphics[width = 5cm,height = 3cm]{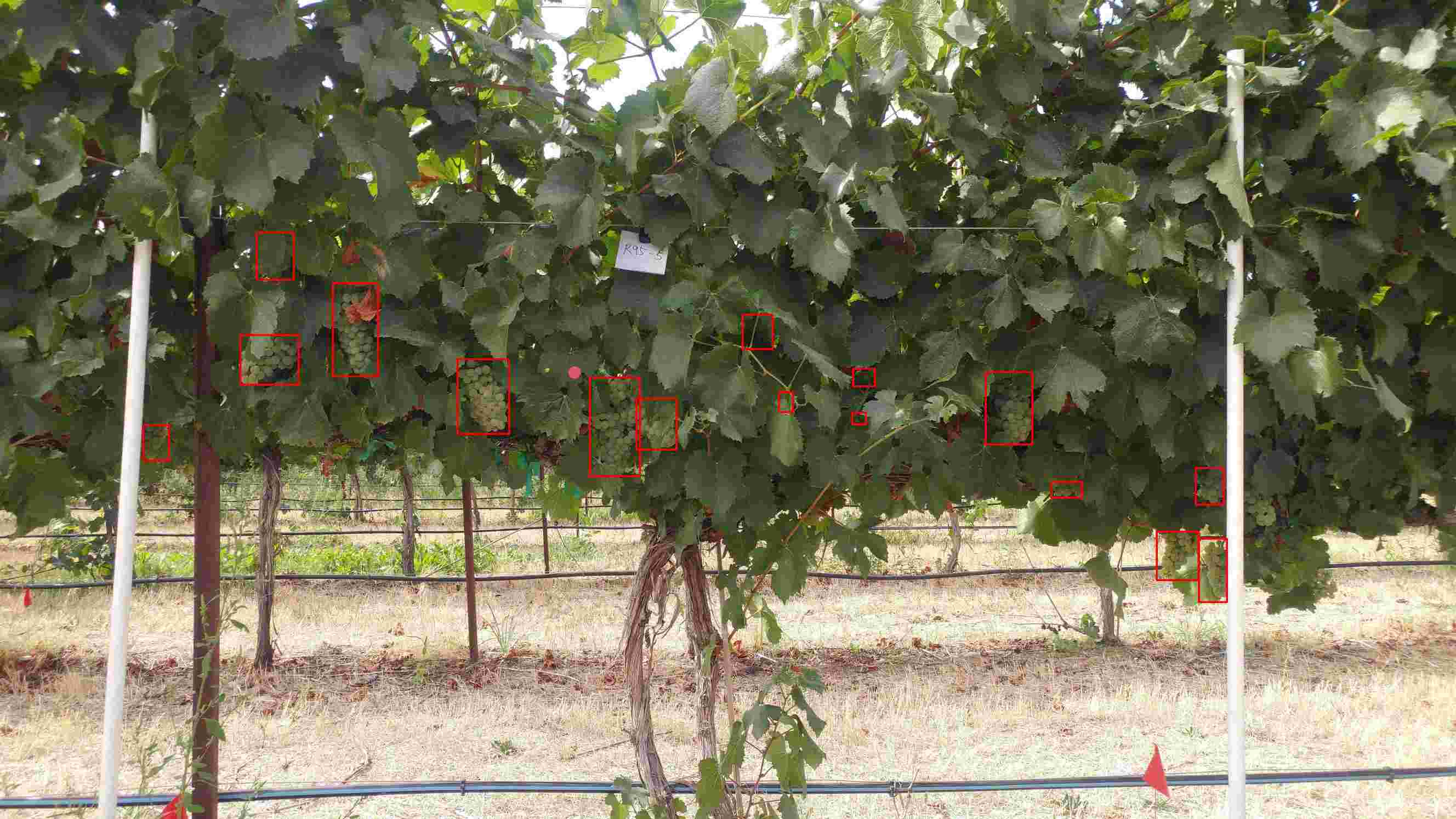} \label{fig.a5h}
  }
  \subfigure[]{
  \includegraphics[width = 5cm,height = 3cm]{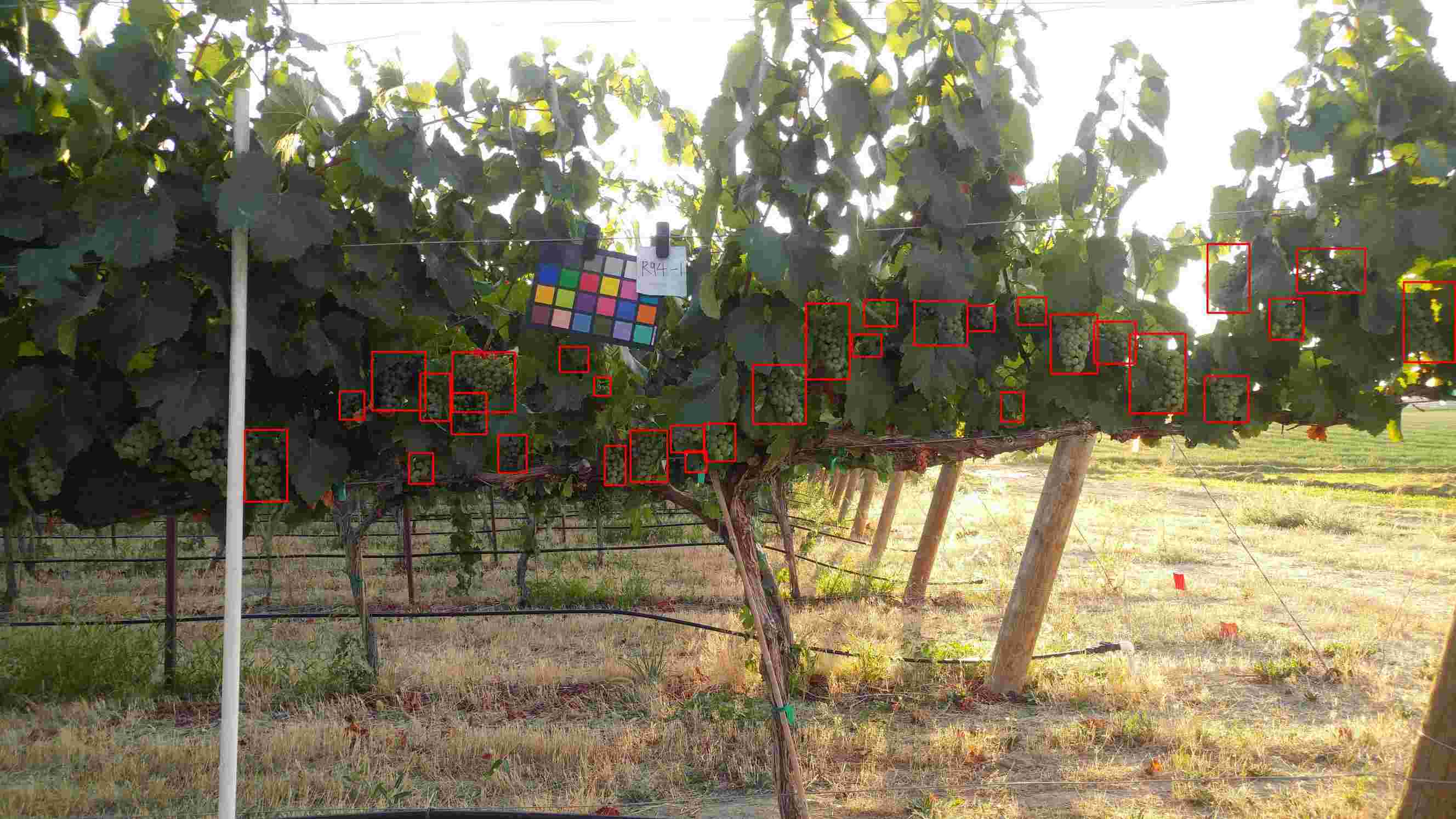} \label{fig.a5i}
  }
  \subfigure[]{
  \includegraphics[width =5cm,height = 3cm]{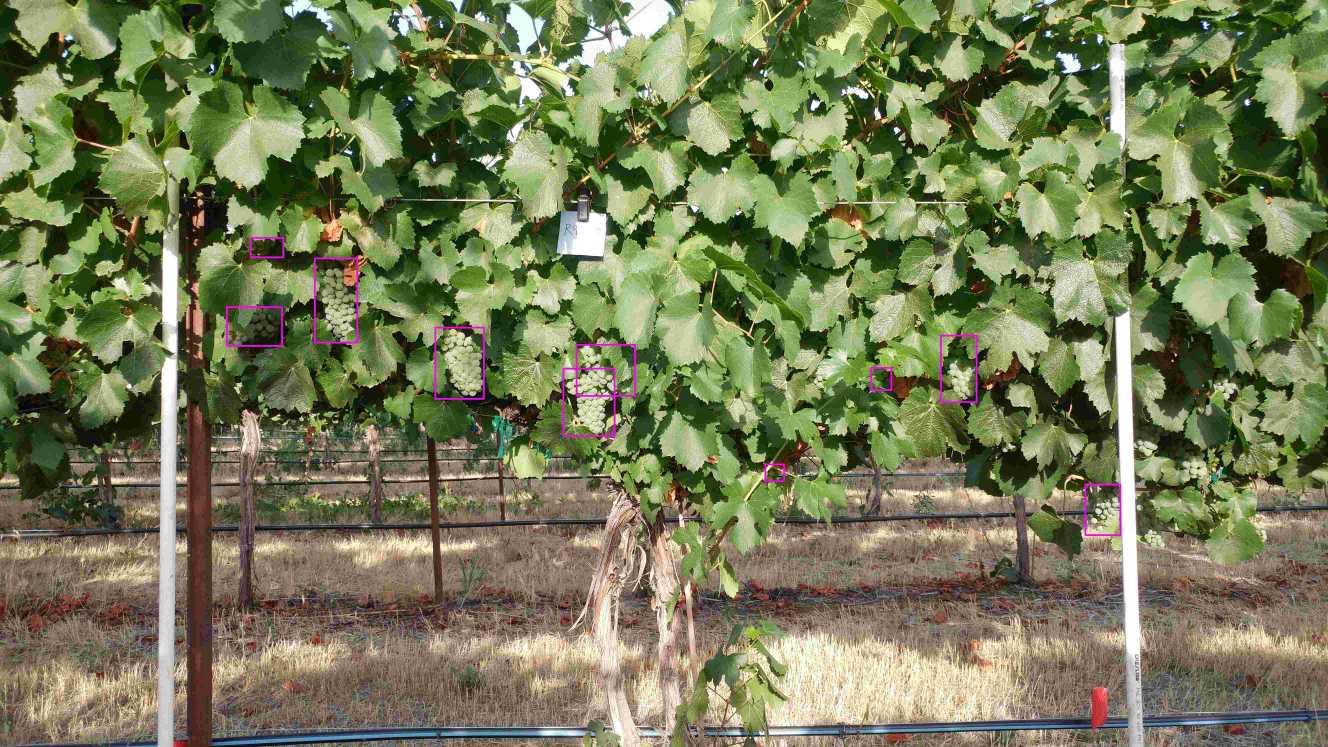} \label{fig.a5j}
  }
    \subfigure[]{
  \includegraphics[width = 5cm,height = 3cm]{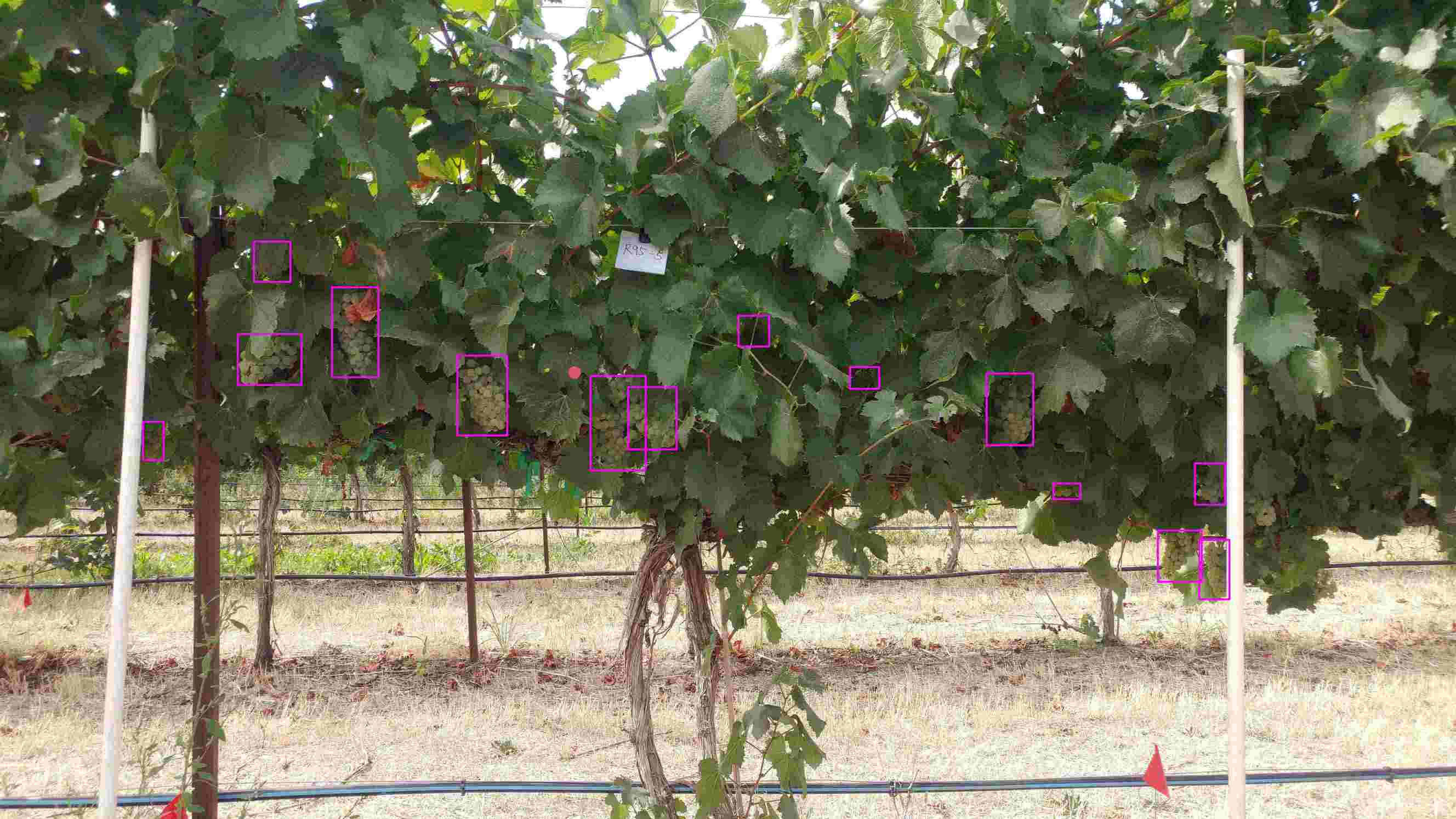} \label{fig.a5k}
  }
    \subfigure[]{
  \includegraphics[width = 5cm,height = 3cm]{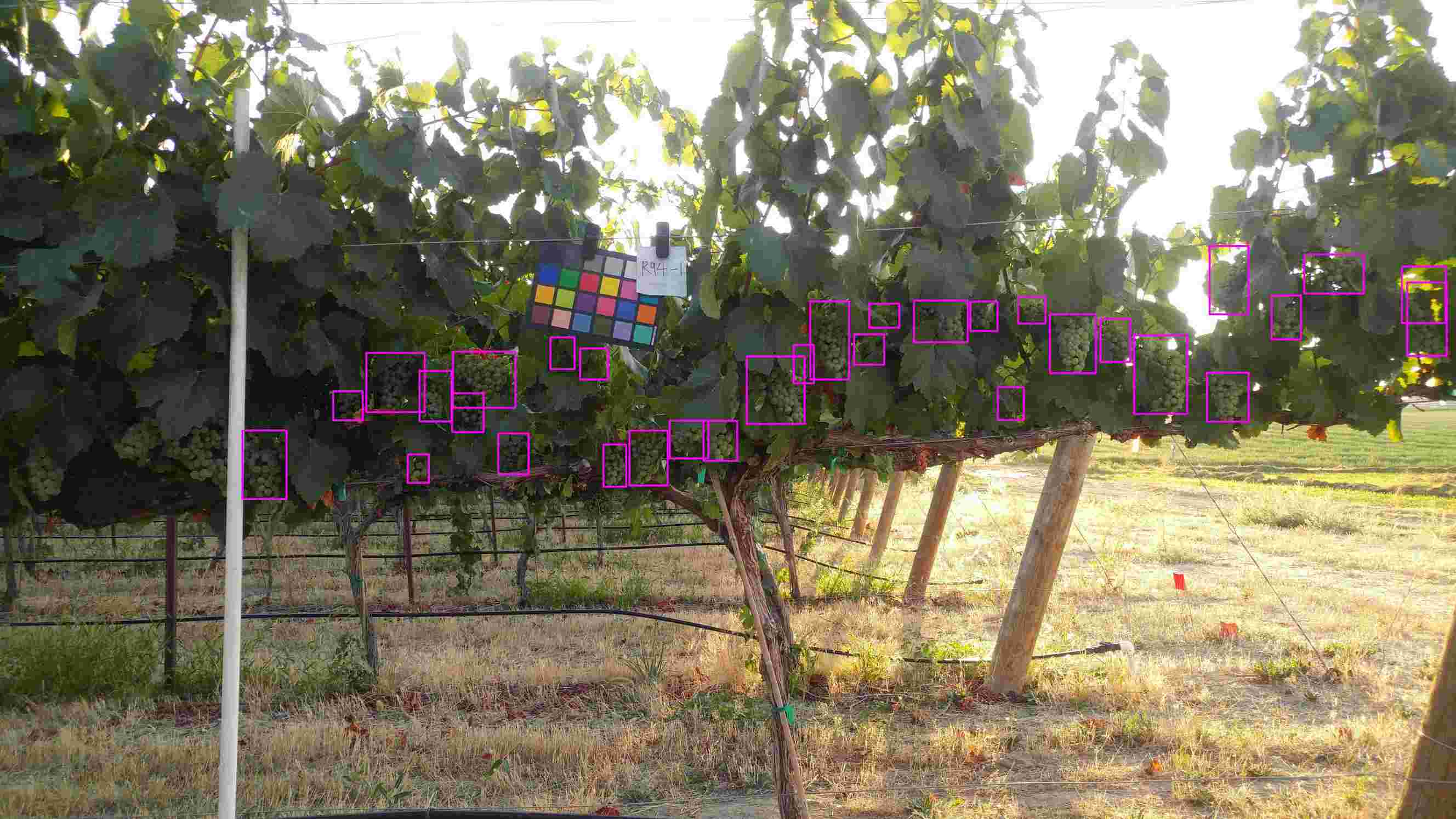} \label{fig.a5l}
  }
    \subfigure[]{
  \includegraphics[width = 5cm,height = 3cm]{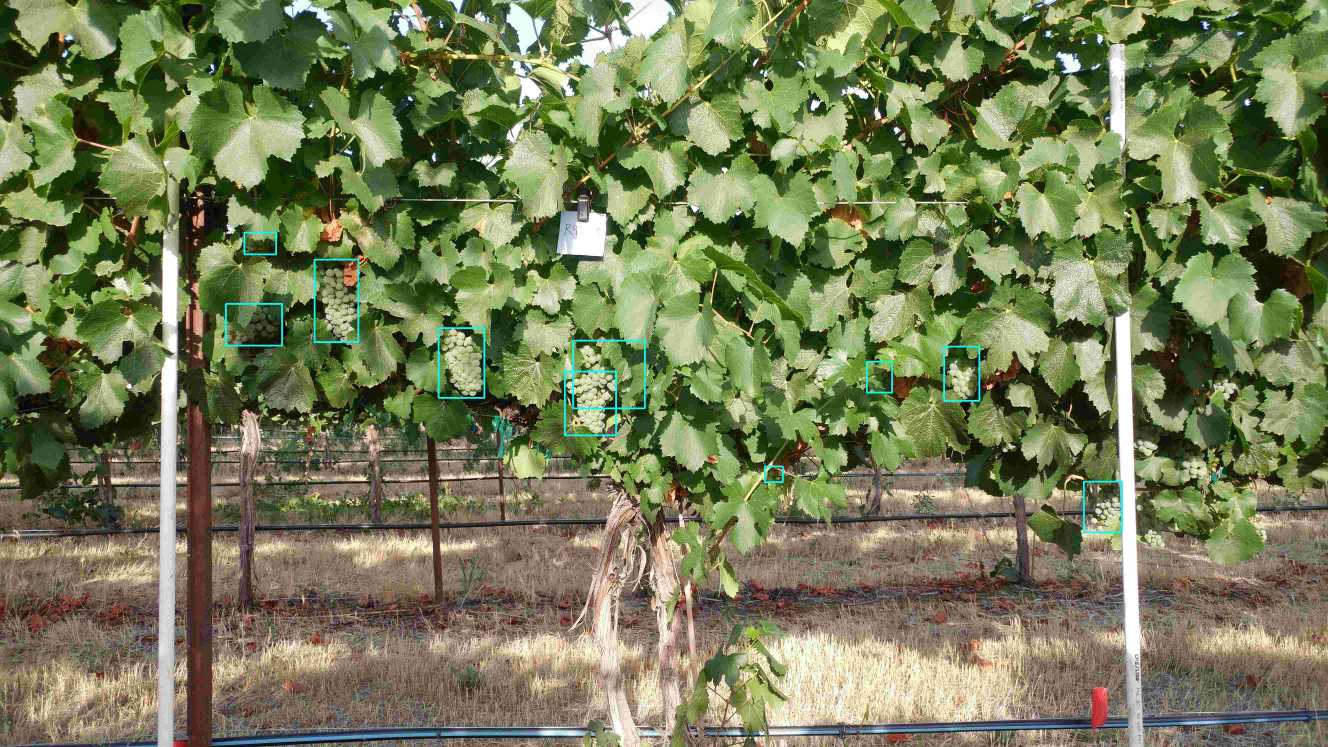} \label{fig.a5m}
  }
    \subfigure[]{
  \includegraphics[width =5cm,height = 3cm]{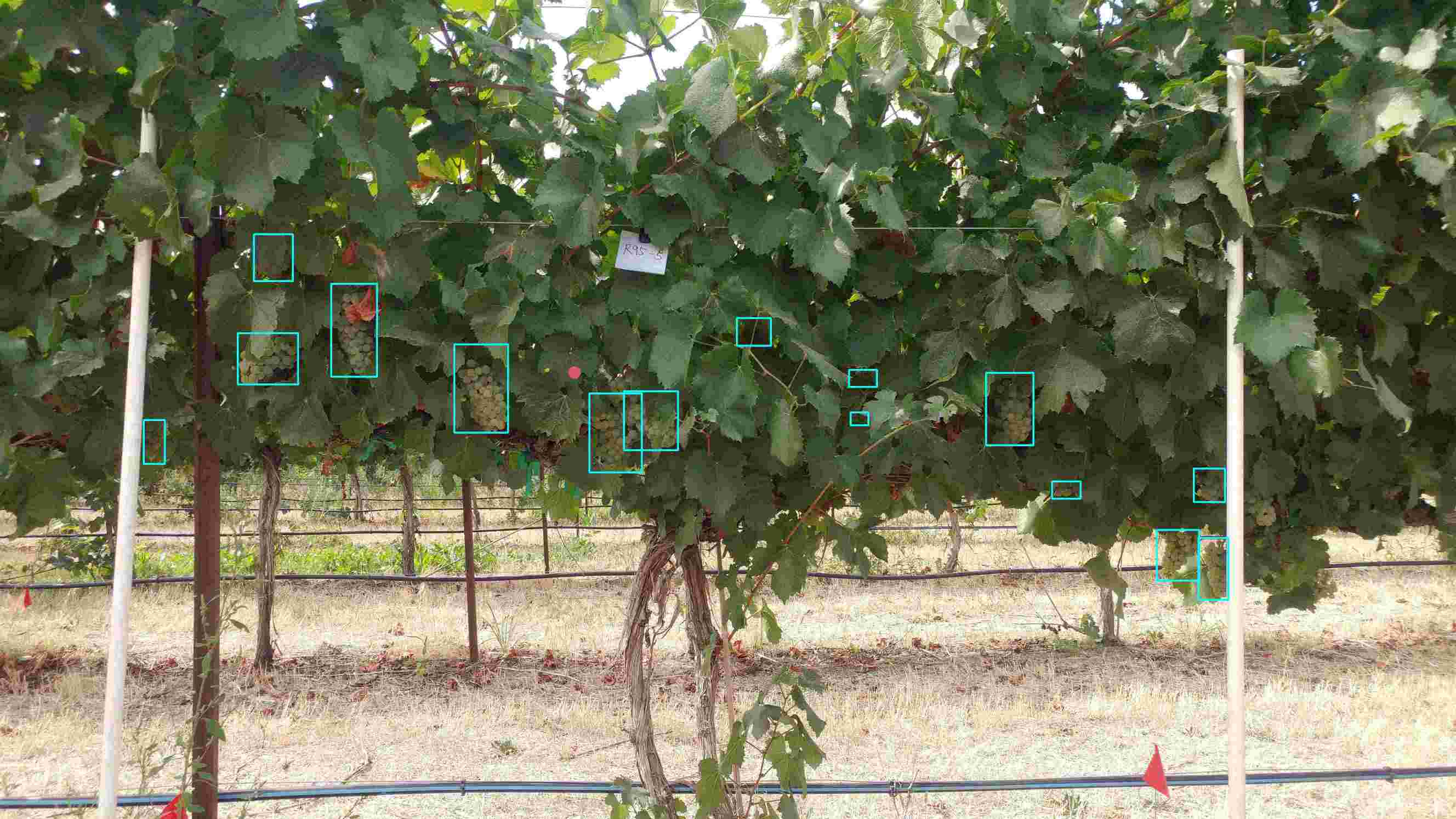} \label{fig.a5n}
  }
    \subfigure[]{
  \includegraphics[width = 5cm,height = 3cm]{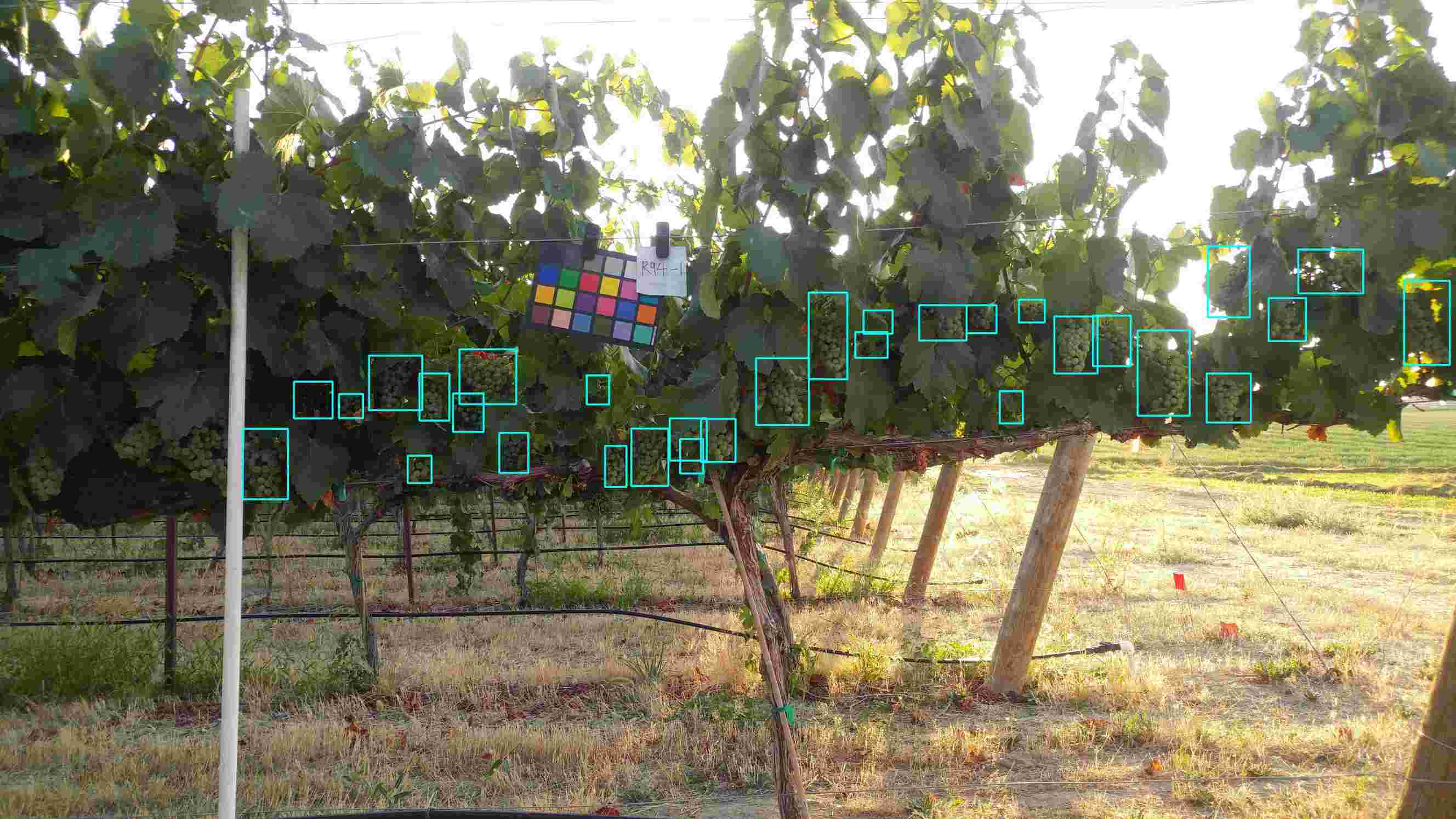} \label{fig.a5o}
  }
  
  \renewcommand*{\thefigure}{A.5}
  \caption{Demonstrations of detection results on the test set of Chardonnay (white variety) using (a-c) Faster R-CNN (bounding boxes in green color), (d-f) YOLOv3 (in blue color), (g-i) YOLOv4 (in red color), (j-l) YOLOv5 (in magenta color), and (m-o) Swin-transformer-YOLOv5 (in cyan color) under morning (left), noon (middle), and afternoon (right) sunlight directions/intensities.}
  \label{figa5}
\end{figure}
\newpage
\begin{figure}[!h]
  \centering
  \subfigure[]{
  \includegraphics[width = 5cm,height = 3cm]{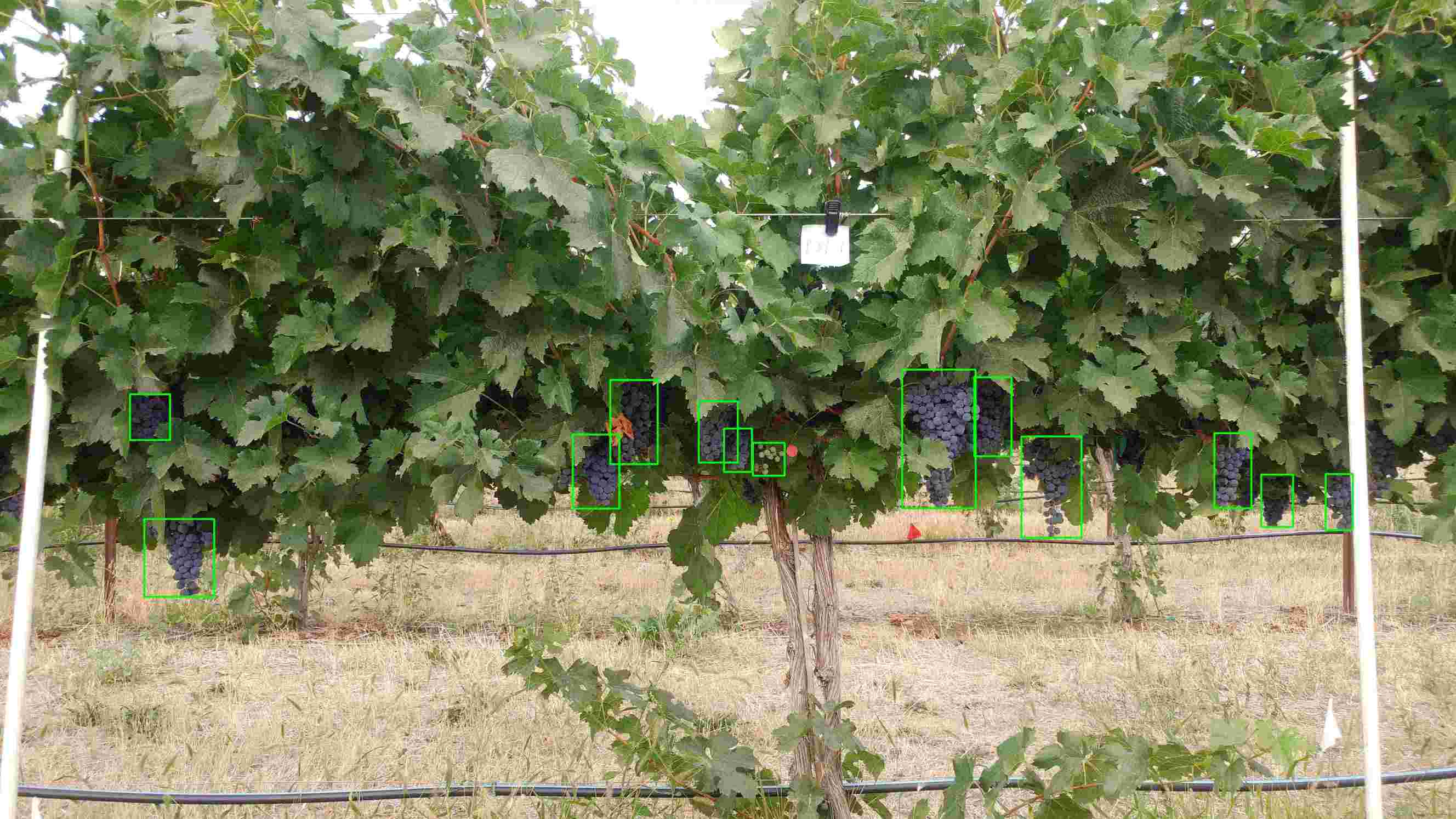} \label{fig.a6a}
  }
  \subfigure[]{
  \includegraphics[width = 5cm,height = 3cm]{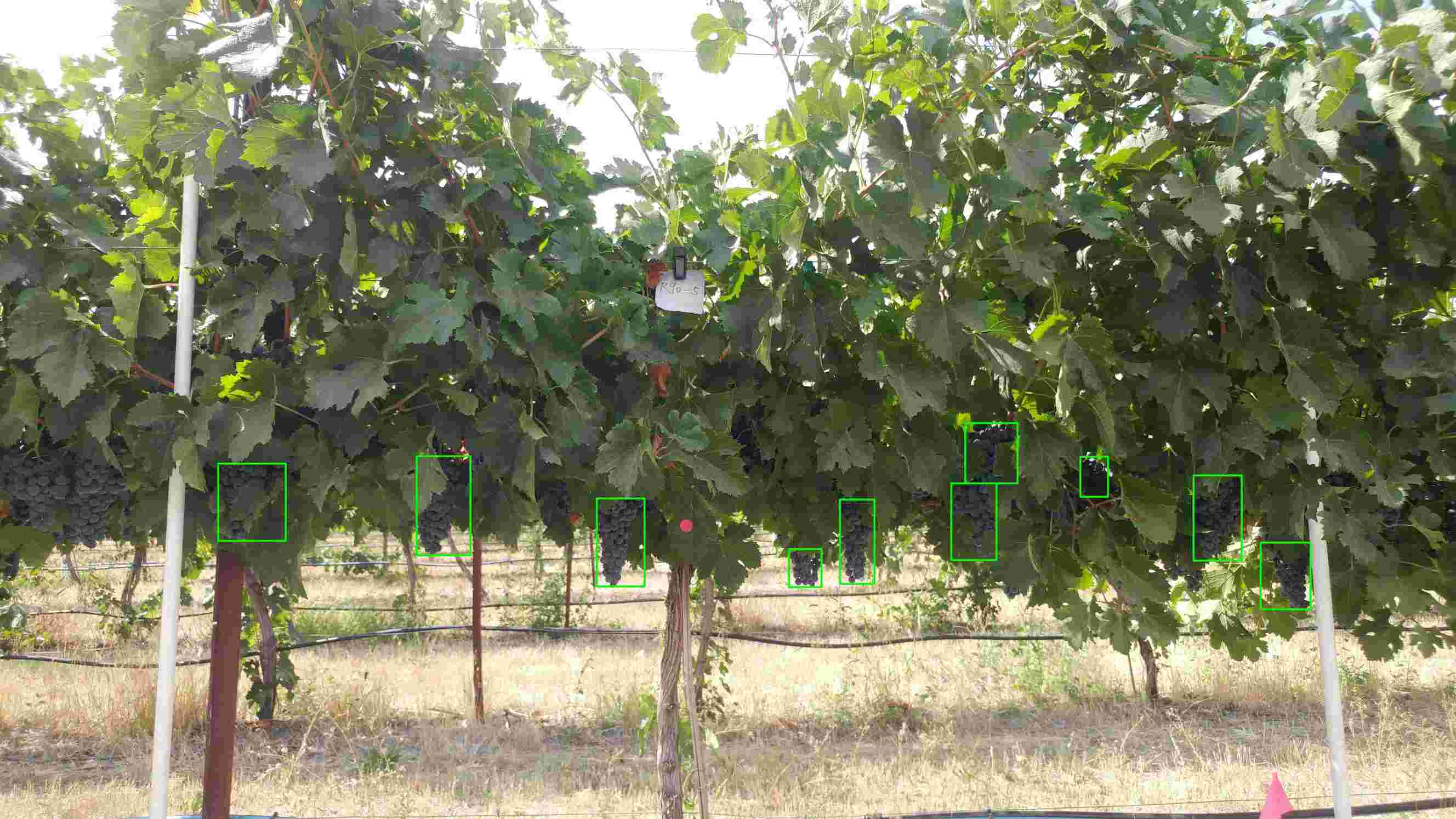} \label{fig.a6b}
  }
  \subfigure[]{
  \includegraphics[width = 5cm,height = 3cm]{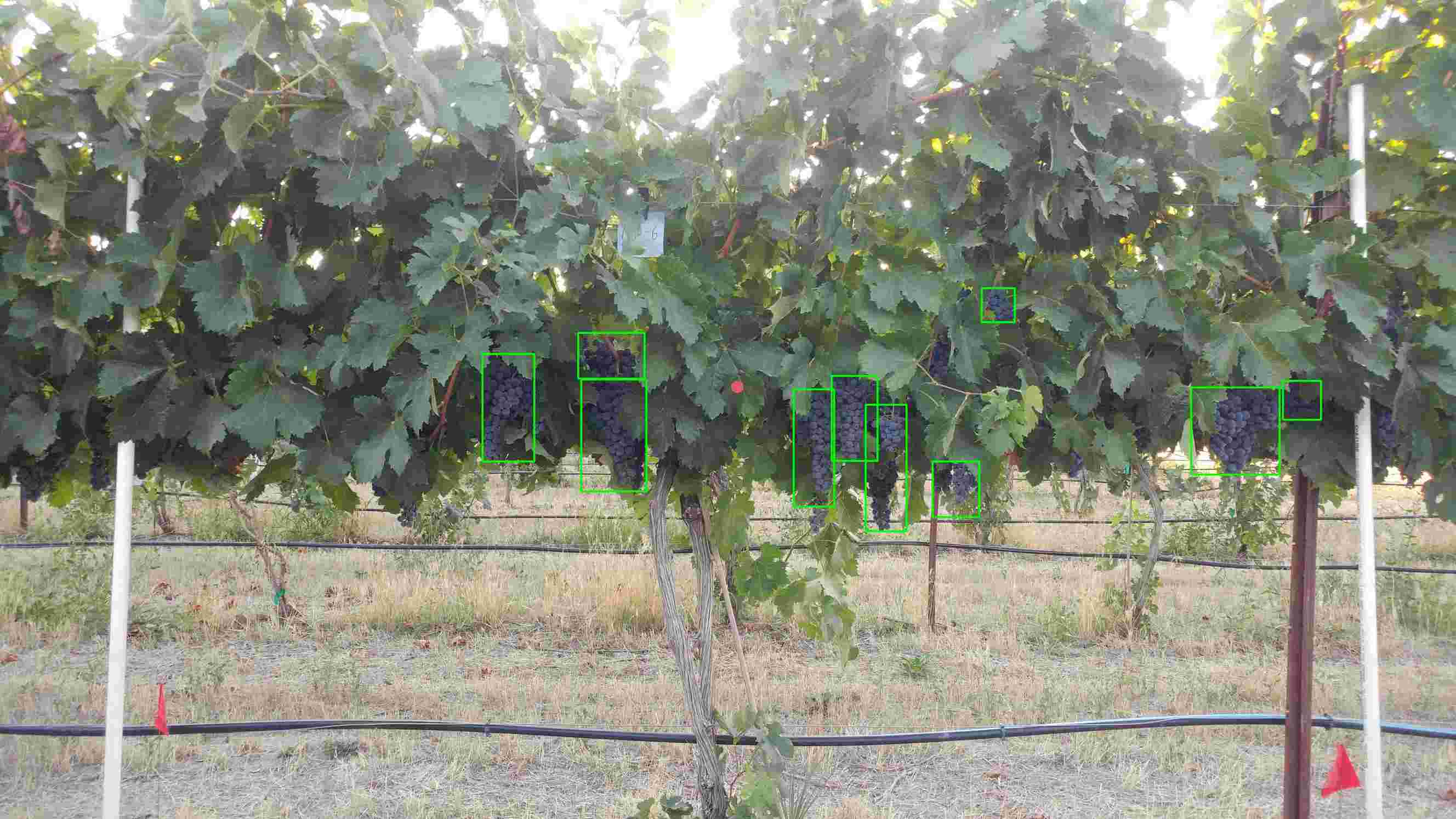} \label{fig.a6c}
  }
  \subfigure[]{
  \includegraphics[width = 5cm,height = 3cm]{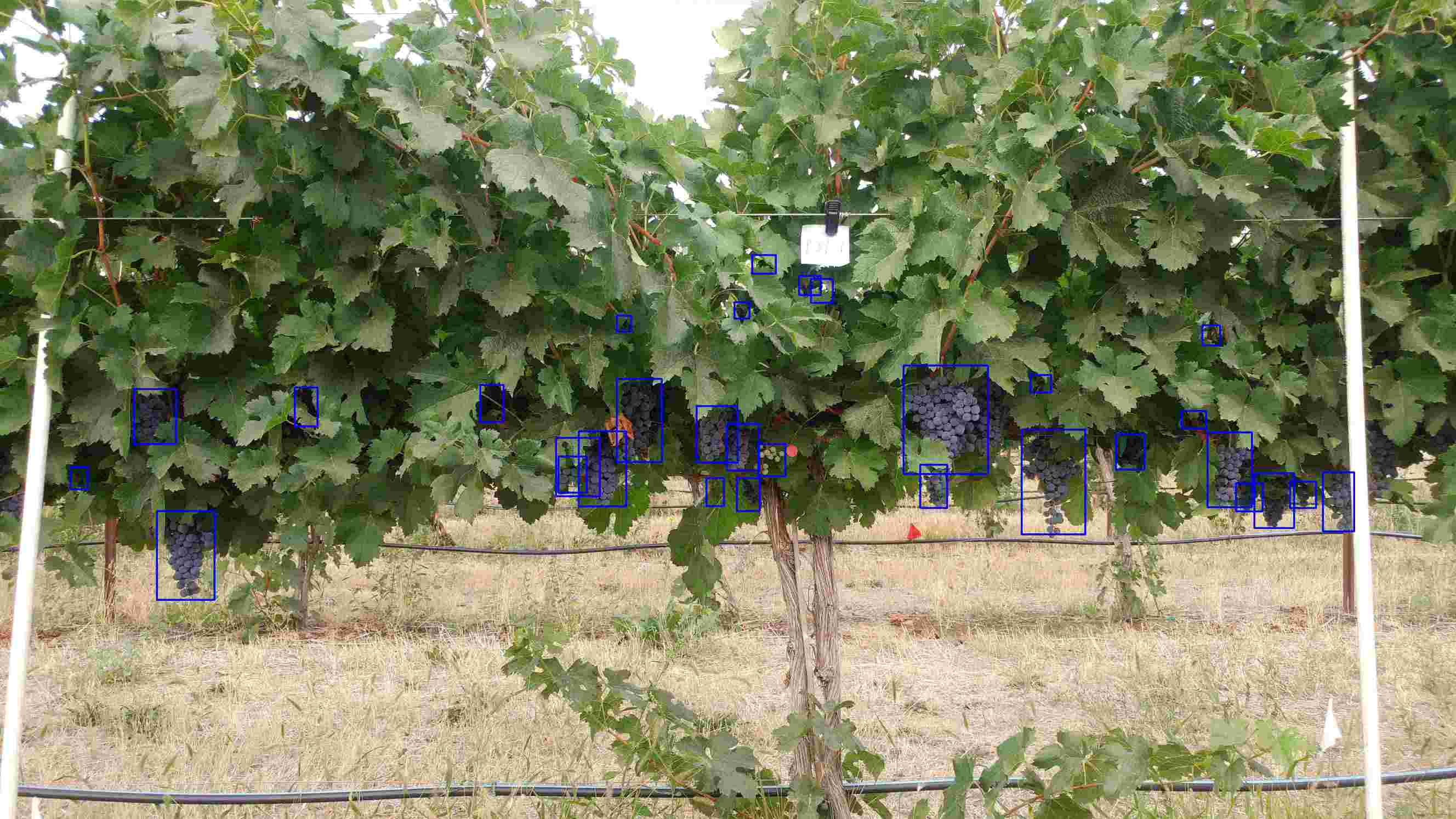} \label{fig.a6d}
  }
  \subfigure[]{
  \includegraphics[width = 5cm,height = 3cm]{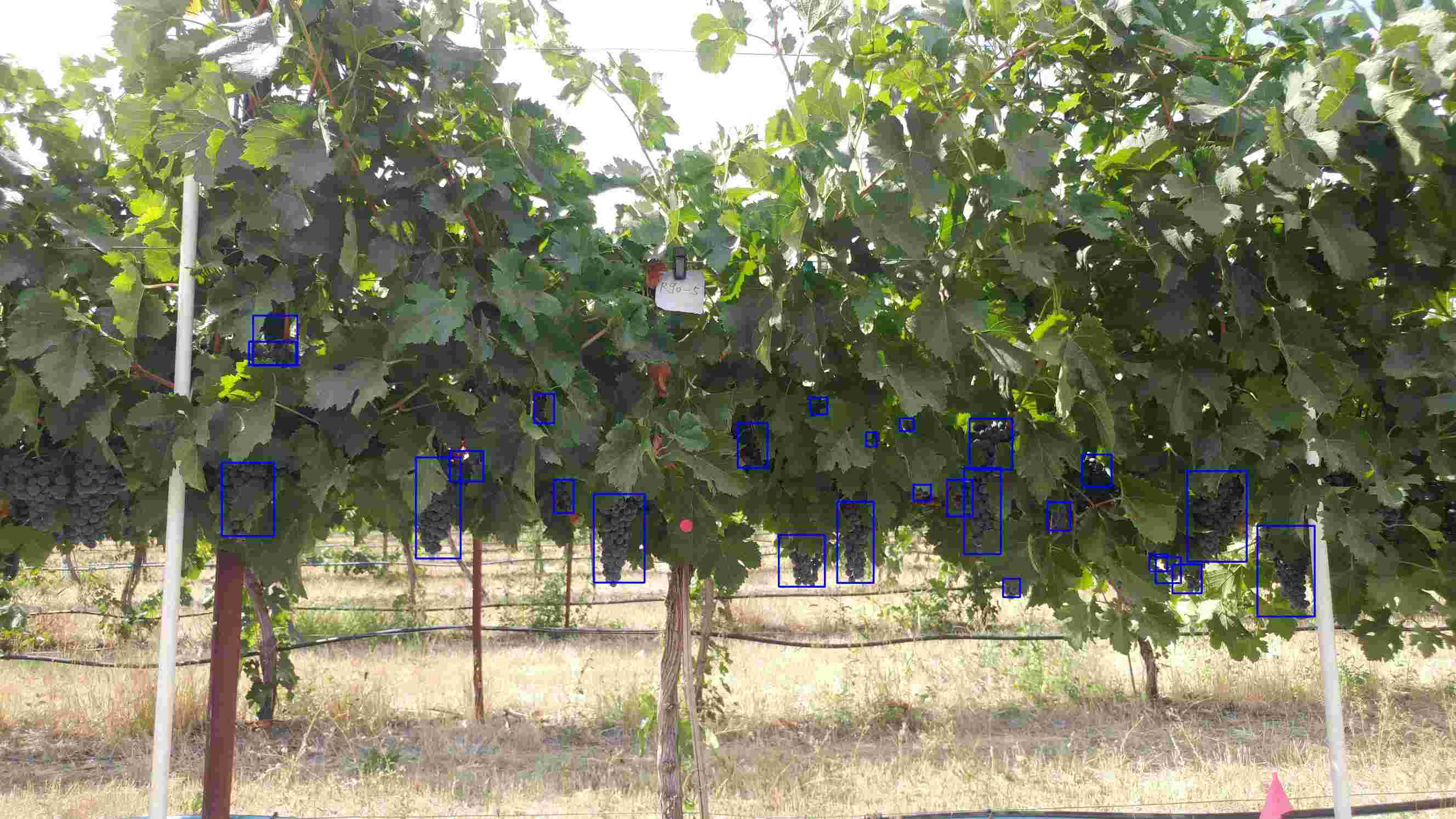} \label{fig.a6e}
  }
  \subfigure[]{
  \includegraphics[width = 5cm,height = 3cm]{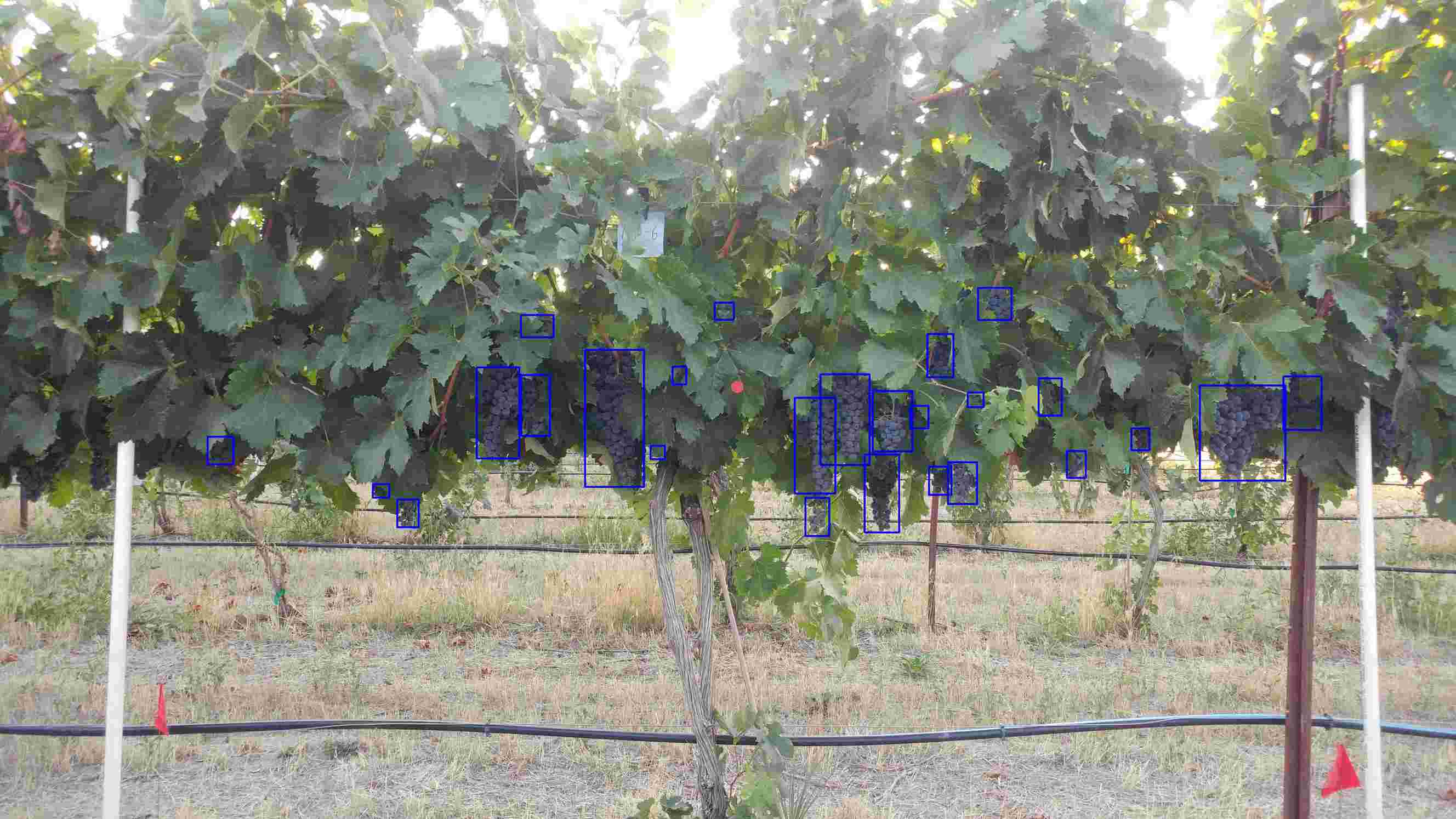} \label{fig.a6f}
  }
  \subfigure[]{
  \includegraphics[width =5cm,height = 3cm]{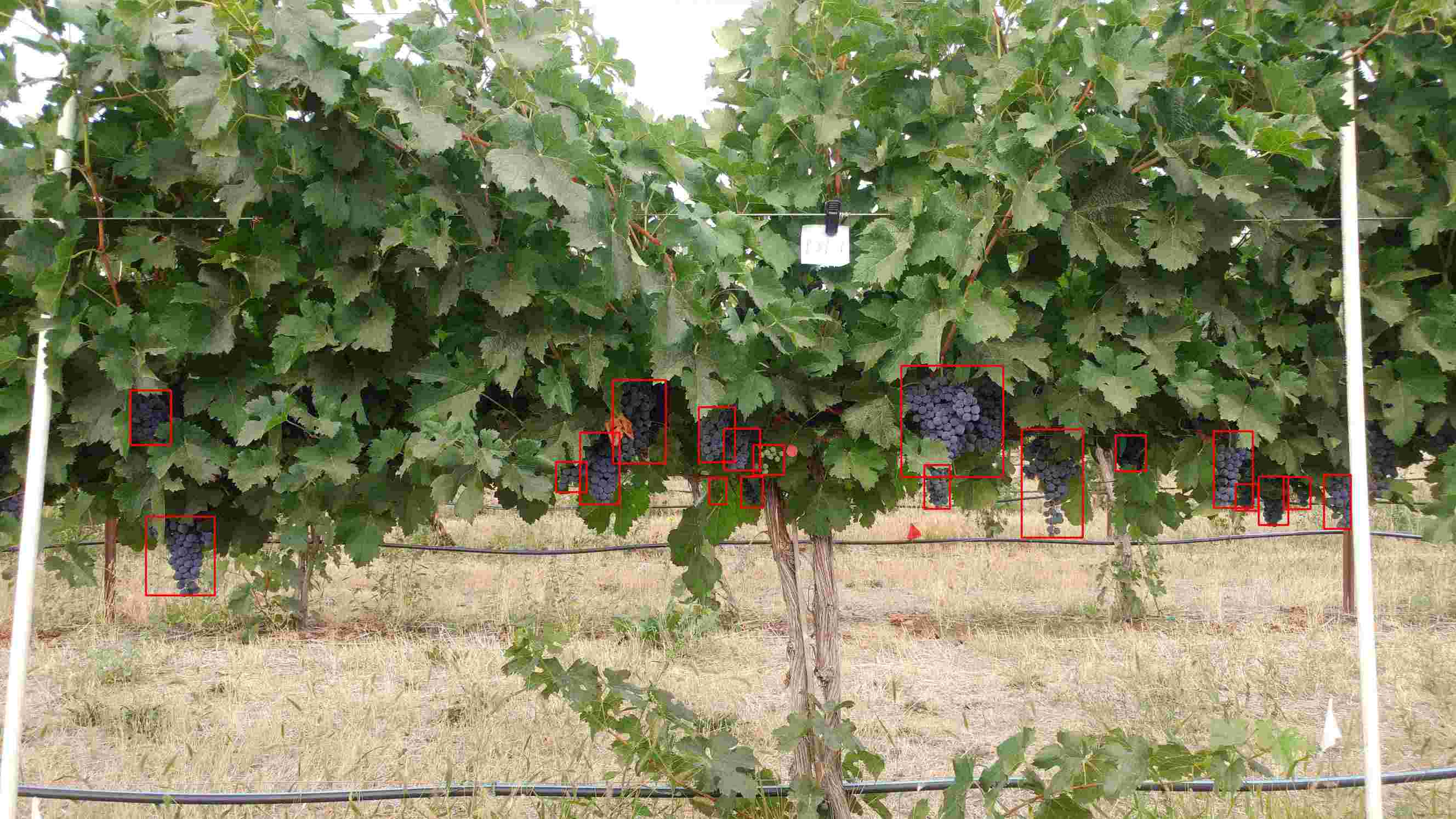} \label{fig.a6g}
  }
  \subfigure[]{
  \includegraphics[width = 5cm,height = 3cm]{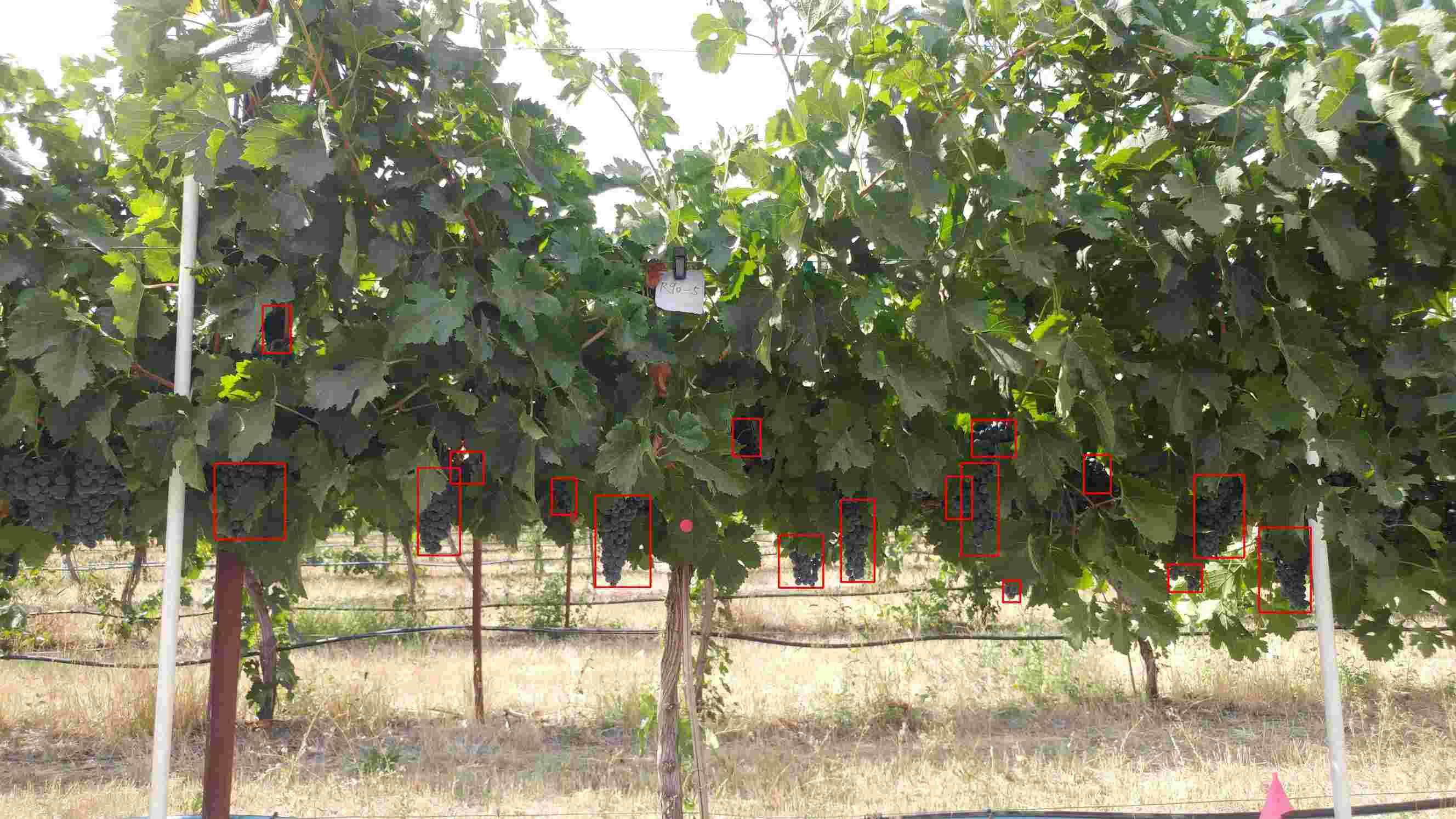} \label{fig.a6h}
  }
  \subfigure[]{
  \includegraphics[width = 5cm,height = 3cm]{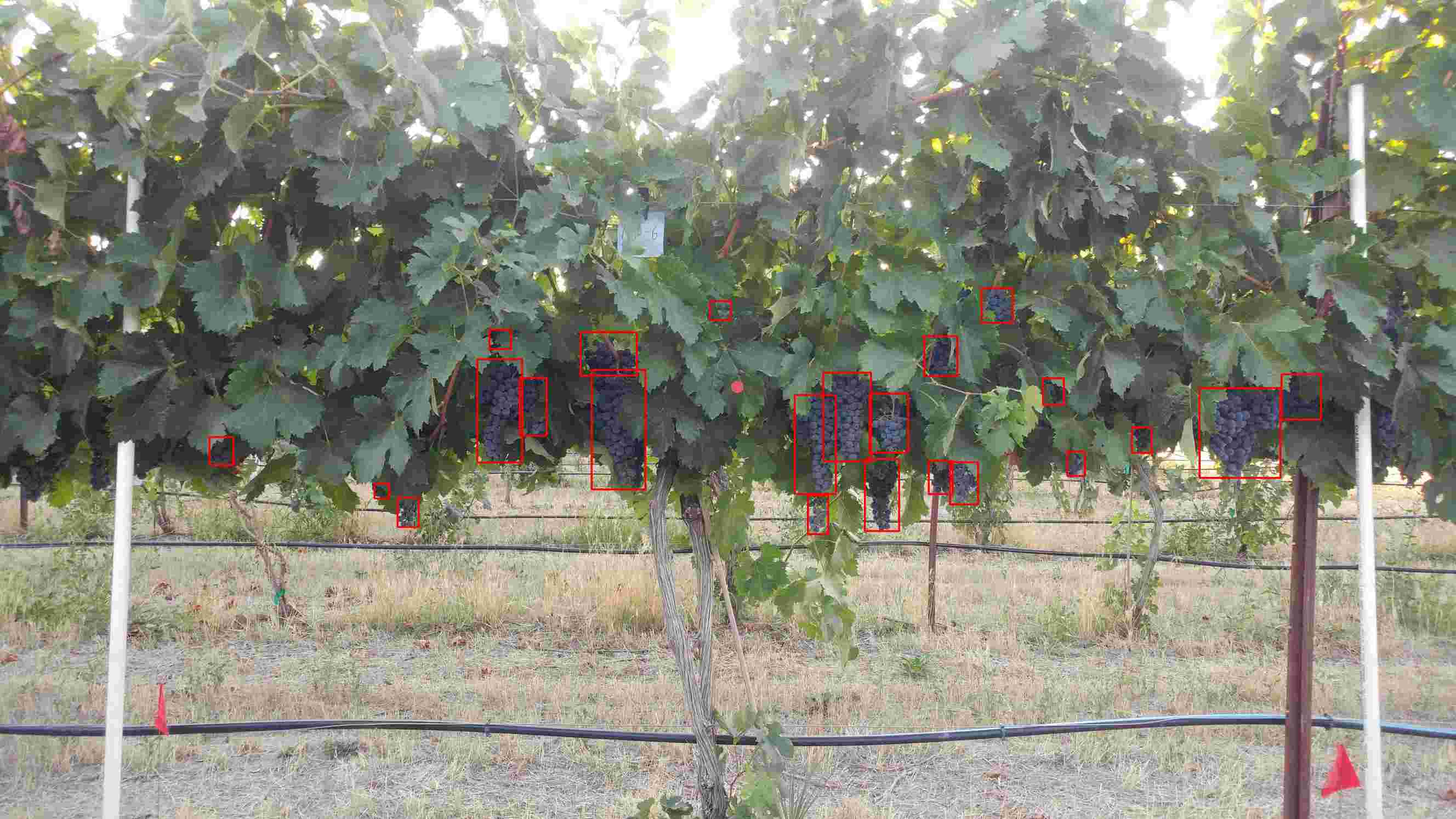} \label{fig.a6i}
  }
  \subfigure[]{
  \includegraphics[width =5cm,height = 3cm]{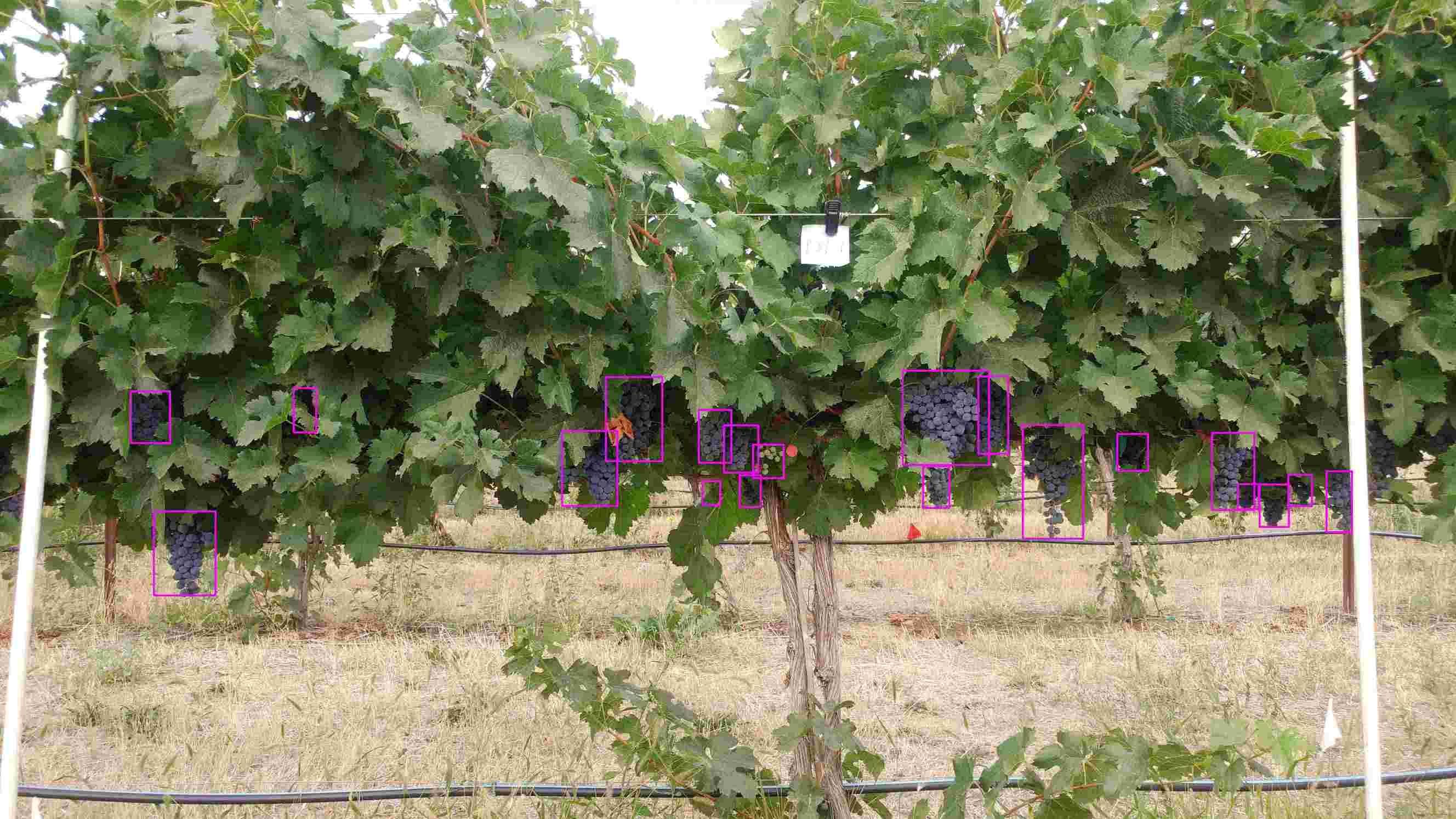} \label{fig.a6j}
  }
    \subfigure[]{
  \includegraphics[width = 5cm,height = 3cm]{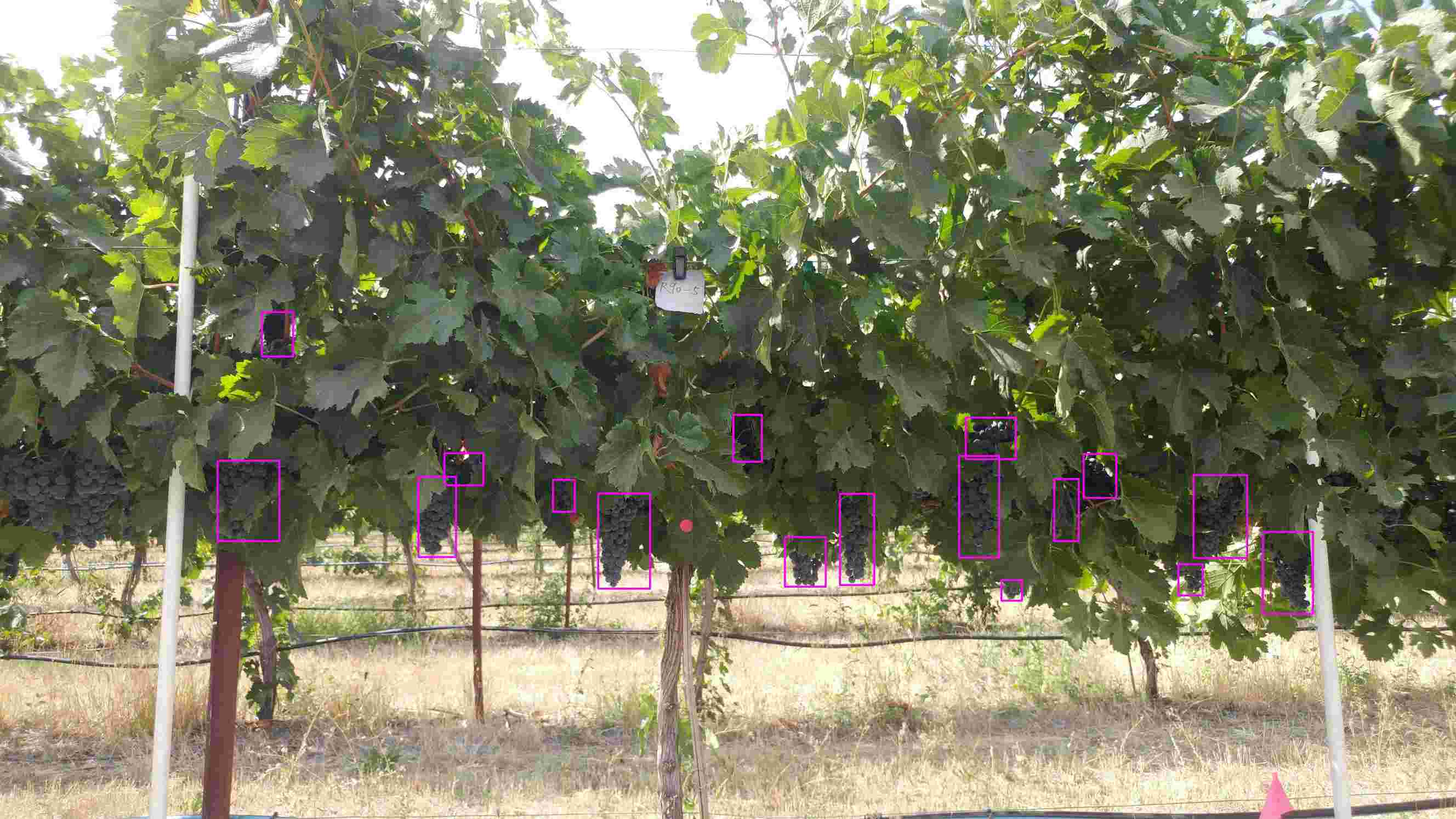} \label{fig.a6k}
  }
    \subfigure[]{
  \includegraphics[width = 5cm,height = 3cm]{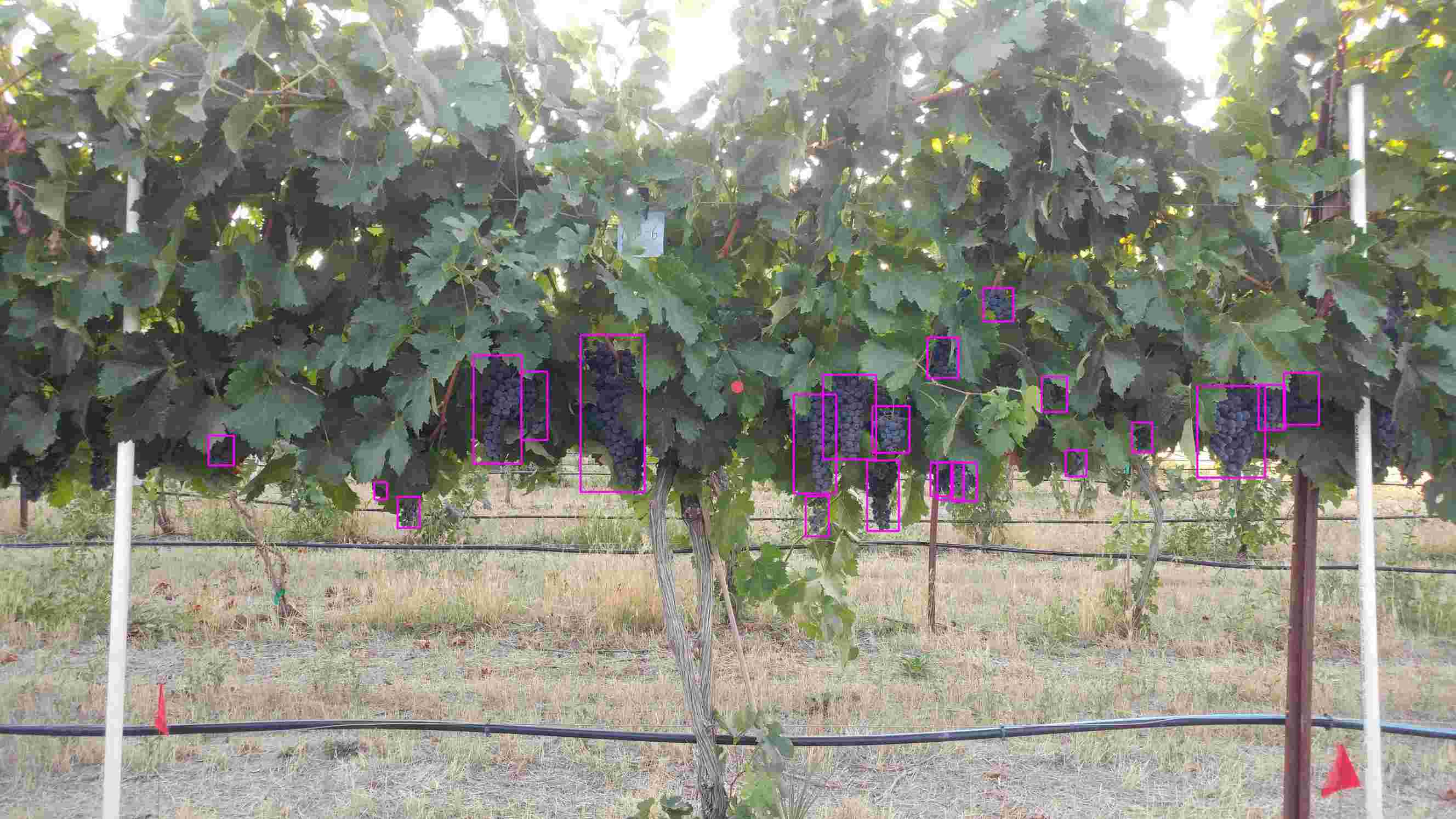} \label{fig.a6l}
  }
    \subfigure[]{
  \includegraphics[width = 5cm,height = 3cm]{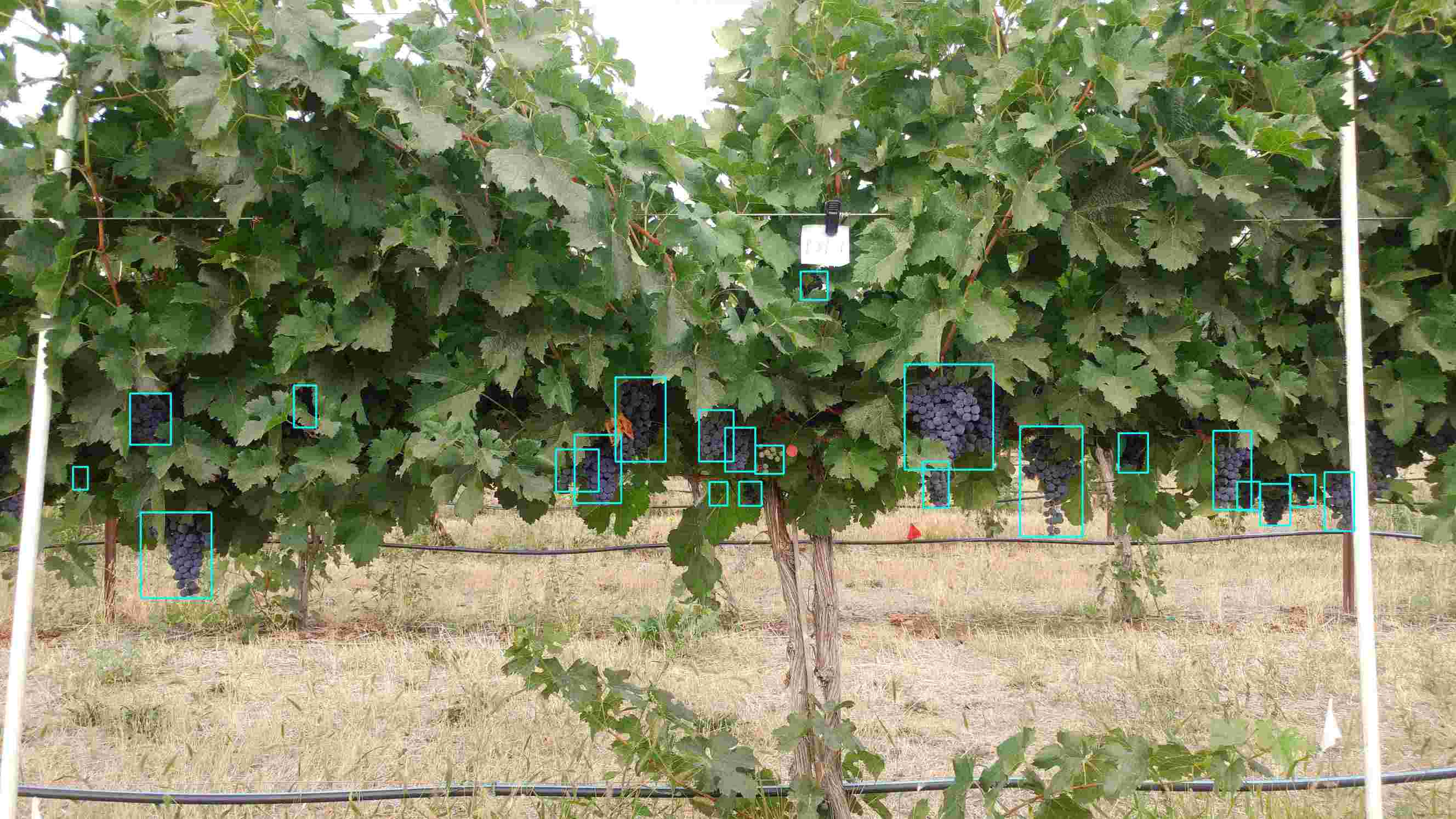} \label{fig.a6m}
  }
    \subfigure[]{
  \includegraphics[width =5cm,height = 3cm]{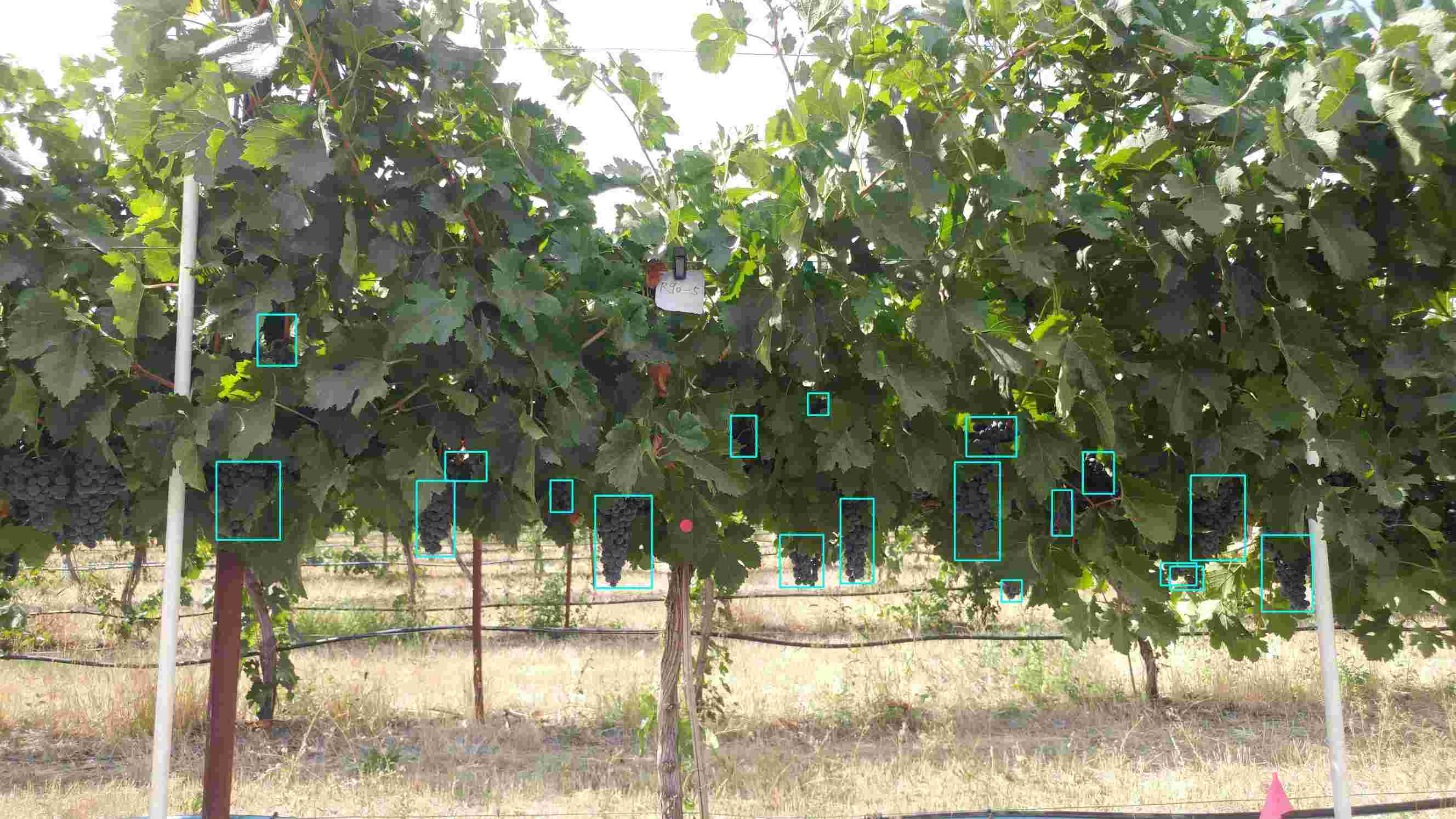} \label{fig.a6n}
  }
    \subfigure[]{
  \includegraphics[width = 5cm,height = 3cm]{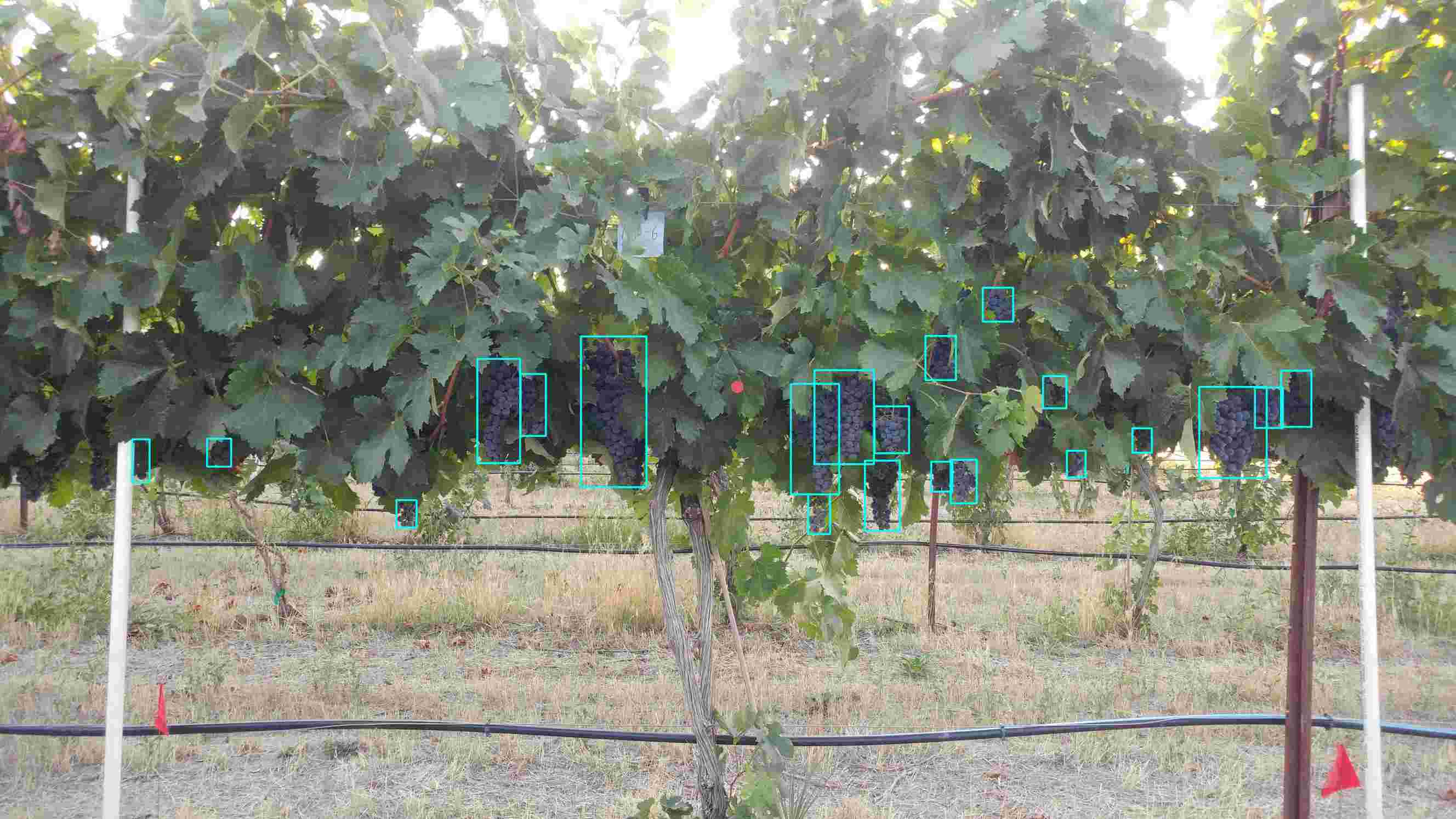} \label{fig.a6o}
  }
  
  \renewcommand*{\thefigure}{A.6}
  \caption{Demonstrations of detection results on the test set of Merlot (red variety) using (a-c) Faster R-CNN (bounding boxes in green color), (d-f) YOLOv3 (in blue color), (g-i) YOLOv4 (in red color), (j-l) YOLOv5 (in magenta color), and (m-o) Swin-transformer-YOLOv5 (in cyan color) under morning (left), noon (middle), and afternoon (right) sunlight directions/intensities.}
  \label{figa6}
\end{figure}

\end{document}